\documentclass{article} 
\usepackage{iclr2026_arxiv,times}

\usepackage{amsmath,amsfonts,bm}

\def\eqref#1{equation~\ref{#1}}

\def\1{\bm{1}}

\DeclareMathAlphabet{\mathsfit}{\encodingdefault}{\sfdefault}{m}{sl}
\SetMathAlphabet{\mathsfit}{bold}{\encodingdefault}{\sfdefault}{bx}{n}

\DeclareMathOperator*{\argmin}{arg\,min}

\usepackage[utf8]{inputenc} 
\usepackage[T1]{fontenc}    
\usepackage{hyperref}       
\usepackage{url}            
\usepackage{booktabs}       
\usepackage{amsmath}        
\usepackage{amsfonts}       
\usepackage{nicefrac}       
\usepackage{microtype}      
\usepackage{xcolor}         
\usepackage{colortbl}       
\usepackage{multirow}       
\usepackage{tablefootnote}  
\usepackage{graphicx}       
\usepackage{algorithm}      
\usepackage{algpseudocode}  
\usepackage{comment}        
\usepackage{caption}        

\usepackage[inline]{enumitem}

\usepackage[dvipsnames]{xcolor}
\usepackage{todonotes}
\usepackage{subcaption}

\usepackage{cancel}

\newtheorem{theorem}{Theorem}
\newtheorem{lemma}[theorem]{Lemma}
\numberwithin{theorem}{section}

\usepackage[nolist]{acronym}

\begin{acronym}
    \acro{HQ}{high-quality}
    \acro{LQ}{low-quality}
    \acro{HR}{high-resolution}
    \acro{LR}{low-resolution}
    \acro{LDCT}{Low-Dose Computed Tomography}
    \acro{RDCT}{Routine-Dose Computed Tomography}
    \acro{LF-MRI}{Low-Field Magnetic Resonance Imaging}
    \acro{HF-MRI}{High-Field Magnetic Resonance Imaging}
    \acro{BPAE}{bovine pulmonary artery endothelial}
    \acro{SNR}{signal-to-noise ratio}
    \acro{SR}{super-resolution}
    \acro{DL}{deep learning}
    \acro{NN}{neural network}
    \acrodefplural{NNs}{neural networks}
    \acro{CNN}{convolutional neural network}
    \acrodefplural{CNNs}{convolutional neural networks}
    \acro{GAN}{generative adversarial network}
    \acrodefplural{GANs}{generative adversarial networks}
    \acro{DDPM}{denoising diffusion probabilistic model}
    \acrodefplural{DDPMs}{denoising diffusion probabilistic models}
    \acro{DDIM}{denoising diffusion implicit model}
    \acrodefplural{DDIMs}{denoising diffusion implicit models}
    \acro{DDRM}{denoising diffusion restoration model}
    \acrodefplural{DDRMs}{denoising diffusion restoration models}
    \acro{RDIM}{residual diffusion implicit model}
    \acrodefplural{RDIMs}{residual diffusion implicit models}
    \acro{MSE}{mean squared error}
    \acro{PSNR}{peak signal-to-noise ratio}
    \acro{SSIM}{structural similarity index measure}
    \acro{LPIPS}{learned perceptual image patch similarity}
    \acro{SDE}{stochastic differential equation}
    \acro{ODE}{ordinary differential equation}
\end{acronym}

\usepackage{color,soul}
\renewcommand{\hl}{\relax}

\title{Residual Diffusion Implicit Models}

\author{Jo\~ao Guerreiro\textsuperscript{1,3,*}, Pedro Tom\'as\textsuperscript{1}, Helena Aidos\textsuperscript{2} \& Jacinto C. Nascimento\textsuperscript{3} \\
\textsuperscript{1} INESC-ID, Instituto Superior T\'ecnico, Universidade de Lisboa, Lisboa, Portugal \\
\textsuperscript{2} LASIGE, Faculdade de Ci\^encias, Universidade de Lisboa, Lisboa, Portugal \\
\textsuperscript{3} ISR-Lisboa, Instituto Superior T\'ecnico, Universidade de Lisboa, Lisboa, Portugal \\
\textsuperscript{*} Corresponding author: 
\texttt{joao.l.carrilho.guerreiro@tecnico.ulisboa.pt}
}

\iclrfinalcopy
\begin{document}

\maketitle

\begin{abstract}
Diffusion models achieve state-of-the-art results across multiple tasks. However, in inverse problems, standard initialization from pure Gaussian noise misaligns the generative process with real-world degradations. More recent methods such as diffusion bridges impose strict endpoint constraints and often require long reverse processes that are prone to hallucinations. Alternative consistency models provide noise-invariant, one-step mappings but lack inherent variance modeling and can degrade under severe corruption. Hence, residual diffusion implicit models (RDIMs) are proposed, constituting a generalized framework that explicitly models the residuals between high-quality (HQ) and low-quality (LQ) images, aligning the forward process with the actual degradation. A non-Markovian implicit reverse sampler is derived, which can skip intermediate timesteps, enabling accurate few-step or even single-step reconstruction, while mitigating the hallucinations inherent to long diffusion chains. RDIM also introduces a controllable variance mechanism that interpolates between deterministic and stochastic sampling, balancing fidelity and diversity. Furthermore, it enables the straightforward use of perceptual losses, when needed. Experiments on denoising and super-resolution benchmarks demonstrate that RDIMs consistently outperforms the state of the art, including bridge and consistency models, in terms of PSNR, SSIM, and LPIPS, reducing hallucinations while requiring only a few sampling steps (often just one). 
The results position RDIMs as an efficient solution for a broad range of image restoration tasks.
\end{abstract}

\section{Introduction}
\label{sec:introduction}
Image reconstruction is a fundamental problem in computer vision and signal processing, aiming to recover \ac{HQ} images from corrupted observations. Tasks such as image denoising and \ac{SR} are crucial for numerous real-world applications, including medical and biological imaging, satellite imagery, and consumer photo enhancement \citep{sagheer2020review, wang2022comprehensive, delbracio2021mobile}.

\Acp{DDPM} \citep{ho2020denoising} have emerged as a powerful class of models \hl{for image synthesis and have been successfully adapted for image reconstruction}. 
Their probabilistic formulation and iterative refinement enable them to handle challenging tasks by progressively improving predictions through small corrective updates \citep{saharia2022image}. Moreover, their stochasticity enables the exploration of multiple plausible paths, promoting output diversity and often leading to better solutions \citep{lugmayr2022repaint,whang2022deblurring}. These properties make diffusion models well-suited to 
deal with severe noise and information loss \citep{chung2022diffusion}. 


However, these strengths also introduce practical challenges. Although stochasticity is beneficial for capturing diversity and avoiding poor generalization \citep{lugmayr2022repaint, whang2022deblurring, dhariwal2021diffusion}, excessive and uncontrolled variability can hinder convergence in inverse problems, destabilizing the reconstruction process and leading to inconsistent outputs. Therefore, balancing stochasticity is crucial \citep{chung2022come}. 
More critically, the standard \ac{DDPM} formulation initializes the reverse process from pure noise, which is misaligned with reconstruction tasks where a degraded input already provides valuable information \citep{chung2022come, yue2023resshift,wu2024one}.
Additionally, the recursive formulation of diffusion models leads to an inefficient reverse process requiring to traverse all diffusion timesteps, often hundreds \citep{shih2023parallel,liu2024residual}, making them computationally expensive and impractical in latency-sensitive settings.
\hl{Notably, techniques based on denoising diffusion bridge models (DDBMs) \mbox{\citep{zhou2024denoising}} alleviate the noise–data mismatch by explicitly conditioning on degradation endpoints, but they also typically require iterating through all timesteps. While recent works tried to address this issue \mbox{\citep{pan2025unidb++}}, we aim to further improve the reconstruction quality using a minimal number of diffusion steps.}

\hl{To tackle these challenges, we revisit ResShift~\mbox{\citep{yue2023resshift}} and introduce a principled theoretical generalization, which we refer to as }\ac{RDIM}.\hl{ Our framework can be interpreted as a bridge-like approach as it constructs a process connecting a starting-point (}\ac{LQ}\hl{ image) to an ending-point (}\ac{HQ}\hl{ image) while preserving sample-level correspondences between the two domains, an essential property for \mbox{\ac{SR}} and denoising tasks. A key feature of RDIM is its ability to introduce controlled stochasticity into the transportation between the two domains by relaxing the terminal constraint. This can be achieved by controllable variance mechanism that interpolates between deterministic and stochastic reconstructions. This provides greater modeling flexibility compared to the fixed terminal states typically imposed in diffusion bridge methods, allowing to obtain state-of-the-art results in SR and denoising applications. Moreover, an implicit sampling mechanism} in the style of \acp{DDIM} \hl{\mbox{\citep{song2020denoising}} is introduced to allow skipping intermediate steps and improving the efficiency of the reconstruction process through few-step or even single-step HQ reconstructions.} In summary, the main contributions of this work are:
\begin{itemize}
  \item A novel diffusion framework for inverse problems that generalizes ResShift\hl{, offering a bridge-like alignment,} and provides an implicit formulation with efficient sampling, enabling reconstructions in a few or even on a single step.
  \item A controllable variance mechanism that interpolates between deterministic and stochastic reconstructions, balancing fidelity and diversity depending on degradation severity.
  \item \hl{Evidence that reducing the number of reverse steps accelerates inference and yields more faithful reconstructions by limiting hallucinations that arise in long diffusion chains.}
  \item State-of-the-art results on denoising and \ac{SR} benchmarks, showing that \acp{RDIM} outperforms existing methods while reducing the number of inference steps by up to 100$\times$.
\end{itemize}
The implementation is available at \texttt{https://github.com/joaolcguerreiro/RDIM}.

\begin{figure}[t!!!]
    \centering
    \includegraphics[width=\textwidth]{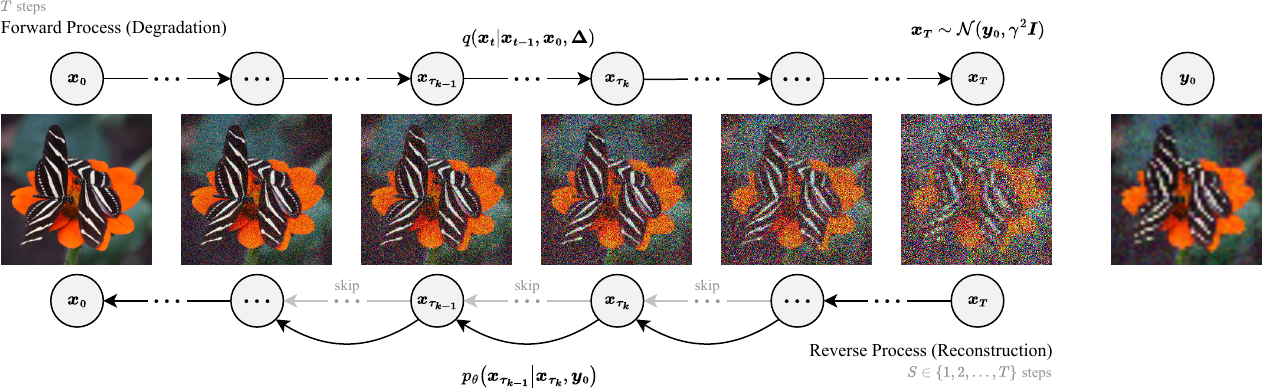}
    \caption{Overview of \ac{RDIM}, a diffusion framework tailored for inverse problems, such as image reconstruction. The reverse process can accurately reconstruct back the data in $S \le T$ steps.}
    \label{fig:rdim}
\end{figure}

\section{Methodology}
\label{sec:method}

\ac{RDIM} is a diffusion framework tailored for inverse problems (herein focused on image reconstruction) where the forward process gradually degrades the original data into an informed corrupted version. The reverse process is efficient, allowing for a minimal number of steps (see Figure \ref{fig:rdim}).

\subsection{Problem Definition}
\label{sec:problem_definition}
Inverse problems are concerned with the recovery of a signal, $\boldsymbol{x_{0}} \in \mathcal{X}$, from a corrupted observation, $\boldsymbol{y_{0}} \in \mathcal{Y}$. Particularly, the forward model that degrades the original signal can be expressed as:
\begin{equation}
\label{eq:degradation_process}
\boldsymbol{y_{0}} = \mathcal{F}\left(\boldsymbol{x_{0}}\right),
\end{equation}
where $\mathcal{F}: \mathcal{X} \to \mathcal{Y}$ is a known or unknown forward operator that often entails information loss (e.g., blurring, downsampling, masking, or noise). Accordingly, such problems are often ill-posed. 

Meanwhile, \ac{DL} techniques can be leveraged to learn a parametric reconstruction model $\mathcal{R}: \mathcal{Y} \to \mathcal{X}$, with trainable parameters $\Theta$, that invert the forward model:
\begin{equation}
\label{eq:dl_reconstruction}
\boldsymbol{x_{0}} \approx \mathcal{R}\left(\boldsymbol{y_{0}}; \Theta\right).
\end{equation}

Traditional diffusion models 
reconstruct the signal $\boldsymbol{x_{0}}$ 
through a parameterized Markov chain with length $T$,
which starts from pure noise and progressively denoises latent variables, $\boldsymbol{x_{t}}$, at each step $t \in \{1, 2, \dots, T\}$. Hence, they first derive a diffusion process that transforms $\boldsymbol{x_{0}}$ into pure noise. Subsequently, they learn to reverse this process by training a parametric model, $p_{\theta}$, which can reconstruct $\boldsymbol{x_{0}}$ back from pure noise, $\boldsymbol{x_{T}} \sim \mathcal{N}(0, \boldsymbol{I})$, while conditioning on the corresponding degraded observation, $\boldsymbol{y_{0}}$. 
However, this diffusion process is fundamentally misaligned with the degradation model in Equation~\ref{eq:degradation_process}, since it maps $\boldsymbol{x_{0}}$ to pure noise rather than to the corrupted observation $\boldsymbol{y_{0}}$. In contrast, the \ac{RDIM} forward process is explicitly designed to align with the degradation mechanism by progressively removing the residuals between the clean and corrupted signals while optionally injecting a controllable amount of noise. This stochastic component introduces variability that improves generalization, enabling the model to balance fidelity and diversity during reconstruction and better capture the uncertainty inherent in inverse problems.

\subsection{Markovian Forward Process}
\label{sec:markovian_forward_process}
Considering that $\boldsymbol{x_{0}}$ and $\boldsymbol{y_{0}}$ denote the original data and its corrupted version\footnote{To match dimensionalities, $\boldsymbol{y_0}$ is upsampled for \ac{SR} tasks and its channels are replicated for colorization.}, respectively, the \ac{RDIM} forward process (degradation) intends to gradually remove fractions of the residual, $\boldsymbol{\Delta} = \boldsymbol{x_{0}} - \boldsymbol{y_{0}}$, from $\boldsymbol{x_{0}}$ over a series of timesteps $t \in \{1, 2, \dots, T\}$. For that purpose, a forward process fixed to a Markov chain is first defined, which converts the distribution of the original data, $q(\boldsymbol{x_0})$, into the last latent variable distribution. Following, the whole Markovian forward process is defined as:
\begin{equation}
\label{eq:forward_process_markovian}
q(\boldsymbol{x_{1:T}}|\boldsymbol{x_0}, \boldsymbol{\Delta}) = \prod_{t=1}^{T} q(\boldsymbol{x_t}|\boldsymbol{x_{t-1}}, \boldsymbol{\Delta}),
\end{equation}
where all latent variables $\boldsymbol{x_{1}},  \dots, \boldsymbol{x_{T}}$ have the same dimensionality as the original data, $\boldsymbol{x_{0}} \sim q(\boldsymbol{x_0})$.

The residual is removed from $\boldsymbol{x_{0}}$ according to a fixed weighting schedule $\lambda_1, \lambda_2, \dots, \lambda_T$, which is also used to parameterize the variance in each diffusion transition distribution, defined as a Gaussian. Consequently, at each timestep $t$, the latent variable $\boldsymbol{x_{t}}$ is expressed in terms of the latent variable at the previous timestep, $\boldsymbol{x_{t-1}}$, and the residual, $\boldsymbol{\Delta}$, as follows:
\begin{equation}
\label{eq:forward_process_transition_markovian}
q(\boldsymbol{x_t}|\boldsymbol{x_{t-1}}, \boldsymbol{\Delta}) = \mathcal{N}(\boldsymbol{x_t}|\boldsymbol{x_{t-1}} - \lambda_t\boldsymbol{\Delta}, \gamma^2\lambda_t\boldsymbol{I}),
\end{equation}
where $\gamma \in [0, \infty)$ is a constant hyperparameter introduced to control the strength of the variance, thus allowing interpolation between a deterministic (when $\gamma = 0$) and a stochastic ($\gamma > 0$) forward process. 
Moreover, each weight $\lambda_{t}$, used to control the amount of residual to be removed between each diffusion step, is computed in terms of small non-negative constant hyperparameters $\beta_{0}, \beta_{1}, \dots, \beta_{T}$ as $\lambda_t = \beta_t - \beta_{t-1}$ (see Section \ref{sec:residual_beta_schedule} for details \mbox{on the $\beta$-schedule).}

Furthermore, to avoid a computationally expensive diffusion process, the cumulative forward transitions, $q(\boldsymbol{x_t}|\boldsymbol{x_0}, \boldsymbol{\Delta}) = q(\boldsymbol{x_t}|\boldsymbol{y_0})$, are expressed in closed form by relying on the reparameterization trick (see Appendix \ref{app:forward_process_cumulative_transition_distribution}):
\begin{equation}
\label{eq:forward_process_cumulative_transition}
q(\boldsymbol{x_t}|\boldsymbol{x_0}, \boldsymbol{\Delta}) = \mathcal{N}(\boldsymbol{x_t} | \boldsymbol{x_0} - \beta_t\boldsymbol{\Delta}, \gamma^2\beta_t \boldsymbol{I}).
\end{equation}

Although this forward process matches ResShift~\citep{yue2023resshift}, the corresponding recursive formulation yields an inefficient reverse process that must iterate over many timesteps (particularly for \ac{HQ} inverse problems). Therefore, a \ac{DDIM}-inspired non-Markovian forward process is derived, which preserves the marginal in Eq.~(\ref{eq:forward_process_cumulative_transition}) while still allowing a Markovian reverse process.


\subsection{Non-Markovian Forward Process}
\label{sec:non_markovian_forward_process}
The forward process is implicitly constructed to ensure consistency with the marginal $q(\boldsymbol{x_t}|\boldsymbol{x_0}, \boldsymbol{\Delta})$ and the reverse process. As a result, each forward transition becomes additionally conditioned on $\boldsymbol{x_0}$ rather than just on the immediate previous timestep, $\boldsymbol{x_{t-1}}$, and the residual, $\boldsymbol{\Delta}$. This introduces explicit dependency on the initial data $\boldsymbol{x_0}$, decoupling the forward process from strict Markovian constraints. Moreover, the forward process is expressed in terms of the forward transition posterior, $q(\boldsymbol{x_{t-1}}|\boldsymbol{x_{t}}, \boldsymbol{x_0}, \boldsymbol{\Delta})$, further reflecting the non-Markovian behavior and preservation of $q(\boldsymbol{x_t}|\boldsymbol{x_0}, \boldsymbol{\Delta})$. Therefore, although the \ac{RDIM} forward process is still a distribution over trajectories that start at $\boldsymbol{x_0}$ and end at $\boldsymbol{x_T}$, it is defined as a joint distribution that is factored in reverse\footnote{The forward transition, $q(\boldsymbol{x_t}|\boldsymbol{x_{t-1}}, \boldsymbol{x_0}, \boldsymbol{\Delta})$, can be derived via Bayes' rule.}:
\begin{equation}
\label{eq:forward_process_non_markovian}
q(\boldsymbol{x_{1:T}}|\boldsymbol{x_0}, \boldsymbol{\Delta}) = q(\boldsymbol{x_{T}}|\boldsymbol{x_0}, \boldsymbol{\Delta})\prod_{t=2}^{T} q(\boldsymbol{x_{t-1}}|\boldsymbol{x_{t}}, \boldsymbol{x_0}, \boldsymbol{\Delta}).
\end{equation}

The non-Markovian nature of the forward process enables designing a reverse process that can be deterministic and simulated with a reduced number of transitions due to the conditioning on $\boldsymbol{x_0}$. 
In addition, since the ResShift training objective only depends on the marginal distribution, $q(\boldsymbol{x_t}|\boldsymbol{x_0}, \boldsymbol{\Delta})$, which is preserved, then \ac{RDIM} optimization (see Section \ref{sec:optimization}) will lead to the same training objective as ResShift. Consequently, already trained ResShift models can be leveraged for \ac{RDIM} sampling without requiring additional retraining.


\subsection{Reverse Process}
\label{sec:reverse_process}


The reverse process (reconstruction) intends to revert the forward process, thus sampling back the data, $\boldsymbol{x_0}$. This is achieved by starting from $\boldsymbol{x_T} \sim \mathcal{N}(\boldsymbol{y_0}, \gamma^2\boldsymbol{I})$ and iteratively refining the latent variables $\boldsymbol{x_t}$ 
until $\boldsymbol{x_0}$ is reached. Accordingly, the reverse process involves computing the forward transition posterior $q(\boldsymbol{x_{t-1}}|\boldsymbol{x_{t}}, \boldsymbol{x_0}, \boldsymbol{\Delta})$ (reverse transition), defined as a Gaussian distribution:
\begin{equation}
\label{eq:reverse_process_transition}
q(\boldsymbol{x_{t-1}}|\boldsymbol{x_{t}}, \boldsymbol{x_0}, \boldsymbol{\Delta}) = \mathcal{N}(\boldsymbol{x_{t-1}}|\boldsymbol{\tilde{\mu}_t}, \tilde{\sigma}^2_t\boldsymbol{I}),
\end{equation}
where $\boldsymbol{\tilde{\mu}_t}$ is the mean of the Gaussian distribution and $\tilde{\sigma}^2_t\boldsymbol{I} = \boldsymbol{\tilde{\Sigma}_t}$ is the isotropic covariance matrix. 
Particularly, the reverse transition is designed to preserve the marginal $q(\boldsymbol{x_t}|\boldsymbol{x_0}, \boldsymbol{\Delta})$ (see Appendix \ref{app:reverse_process_transition_distribution}). Considering 
$\tilde{\sigma}^2_t$ matches the ResShift variance, $\tilde{\lambda}_t = \gamma^2\frac{\beta_{t-1}}{\beta_t}\lambda_t$,
the mean, $\boldsymbol{\tilde{\mu}_t}$, is given as:
\begin{equation}
\label{eq:reverse_process_transition_closed_form_mean}
\boldsymbol{\tilde{\mu}_t} =
\begin{cases}
    \boldsymbol{x_0} - \beta_{t-1}\boldsymbol{\Delta}, & \text{if } \gamma = 0, \\
    \boldsymbol{x_0} - \beta_{t-1}\boldsymbol{\Delta} + \sqrt{\gamma^2\beta_{t-1} - \tilde{\sigma}^2_t}\left(\frac{\boldsymbol{x_t} - \boldsymbol{x_0} + \beta_t\boldsymbol{\Delta}}{{\sqrt{\gamma^2\beta_t}}}\right), & \text{if } \gamma \ne 0,
\end{cases}
\end{equation}
where, for $\gamma=0$, the reverse process essentially becomes a linear interpolation between the corrupted and original data, 
which underscores that the \ac{RDIM} forward process 
is aligned with a forward model (degradation process) that converts $\boldsymbol{x_0}$ into $\boldsymbol{y_0}$.

Furthermore, fixing $\tilde{\sigma}^2_t$ to the ResShift variance, $\tilde{\lambda}_t$, also results in $\boldsymbol{\tilde{\mu}_t}$ matching the mean of the ResShift reverse transition (see Appendix \ref{app:reverse_transition_with_resshift_variance}). Hence, \ac{RDIM} becomes ResShift for this specific variance, revealing that ResShift is a particular case of \ac{RDIM}. Subsequently, a constant hyperparameter, $\eta \in [0, 1]$, can be introduced to interpolate between a deterministic ($\eta{=}0$) and a stochastic ($\eta{>}0$) reverse process when $\gamma \neq 0$, allowing control over the variability in the \ac{RDIM} reverse trajectory:
\begin{equation}
\label{eq:reverse_process_transition_closed_form_mean_non_zero_gamma_eta}
\boldsymbol{\tilde{\mu}}_{\boldsymbol{t}|\gamma{\ne}0} = \boldsymbol{x_0} - \beta_{t-1}\boldsymbol{\Delta} + \sqrt{\gamma^2\beta_{t-1} - \eta^2\tilde{\lambda}_t}\left(\frac{\boldsymbol{x_t} - \boldsymbol{x_0} + \beta_t\boldsymbol{\Delta}}{{\sqrt{\gamma^2\beta_t}}}\right), \quad \tilde{\sigma}^2_{t|\gamma{\ne}0} = \eta^2\tilde{\lambda}_t.
\end{equation}
where $\eta=1$ makes the \ac{RDIM} reverse process identical to ResShift. 
Meanwhile, setting $\gamma=0$ converts \ac{RDIM} into a strictly deterministic model ($\gamma{=}0 \Rightarrow \tilde{\lambda}_t{=}0$), 
avoiding sampling random noise. 

However, during inference, $\boldsymbol{x_0}$ and $\boldsymbol{\Delta}$ are unknown, thus sampling from the true reverse transition distribution is not possible. Therefore, a learnable parametric model, $p_{\theta}(\boldsymbol{x_{t-1}}|\boldsymbol{x_t}, \boldsymbol{y_0})$, defined as a Gaussian distribution, is introduced to approximate the true reverse transition $q(\boldsymbol{x_{t-1}}|\boldsymbol{x_t}, \boldsymbol{x_0}, \boldsymbol{\Delta})$:
\begin{equation}
\label{eq:reverse_process_transition_approximation_general}
p_{\theta}(\boldsymbol{x_{t-1}}|\boldsymbol{x_t}, \boldsymbol{y_0}) = \mathcal{N}\left(\boldsymbol{x_{t-1}}| \boldsymbol{\mu_{\theta}}\left(\boldsymbol{x_t}, \boldsymbol{y_0}, t\right), \sigma^2_{\theta}\left(\boldsymbol{x_t}, \boldsymbol{y_0}, t\right)\boldsymbol{I}\right),
\end{equation}
where $\boldsymbol{\mu_{\theta}}\left(\boldsymbol{x_t}, \boldsymbol{y_0}, t\right)$ is the mean of the Gaussian distribution and $\sigma^2_{\theta}\left(\boldsymbol{x_t}, \boldsymbol{y_0}, t\right)\boldsymbol{I} = \boldsymbol{\Sigma_{\theta}}\left(\boldsymbol{x_t}, \boldsymbol{y_0}, t\right)$ is the isotropic covariance matrix. In particular, the variance of the true reverse transition, $\tilde{\sigma}^2_t$, does not have any learnable parameters because it is defined in terms of constant hyperparameters, which are known. Therefore, the variance of $p_{\theta}(\boldsymbol{x_{t-1}}|\boldsymbol{x_t}, \boldsymbol{y_0})$ can be fixed to equal exactly the variance of $q(\boldsymbol{x_{t-1}}|\boldsymbol{x_t}, \boldsymbol{x_0}, \boldsymbol{\Delta})$:
\begin{equation}
\label{eq:reverse_process_transition_approximation_variance}
\sigma^2_{\theta}\left(\boldsymbol{x_t}, \boldsymbol{y_0}, t\right) = \tilde{\sigma}^2_t.
\end{equation}

Meanwhile, $\boldsymbol{\mu_{\theta}}\left(\boldsymbol{x_t}, \boldsymbol{y_0}, t\right)$ approximates the mean of the true reverse transition, $\boldsymbol{\tilde{\mu}_t}$. Considering that $\boldsymbol{x_0}$ and $\boldsymbol{\Delta}$ are the only unknown terms and $\boldsymbol{\Delta}$ can be estimated from $\boldsymbol{x_0}$ and $\boldsymbol{y_0}$, then the model solely needs to predict $\boldsymbol{x_0}$ (see Appendix \ref{app:training objective}). Accordingly, the mean $\boldsymbol{\mu_{\theta}}\left(\boldsymbol{x_t}, \boldsymbol{y_0}, t\right)$ is defined as:
\begin{equation}
\label{eq:reverse_process_transition_approximation_mean}
\boldsymbol{\mu_{\theta}}\left(\boldsymbol{x_t}, \boldsymbol{y_0}, t\right) =
\begin{cases}
    \boldsymbol{\hat{x}_{0}} - \beta_{t-1}\boldsymbol{\hat{\Delta}}, & \text{if } \gamma = 0, \\
    \boldsymbol{\hat{x}_{0}} - \beta_{t-1}\boldsymbol{\hat{\Delta}} + \sqrt{\gamma^2\beta_{t-1} - \eta^2\tilde{\lambda}_t}\left(\frac{\boldsymbol{x_t} - \boldsymbol{\hat{x}_{0}} + \beta_t\boldsymbol{\hat{\Delta}}}{{\sqrt{\gamma^2\beta_t}}}\right), & \text{if } \gamma \ne 0,
\end{cases}
\end{equation}
where $\boldsymbol{\hat{x}_{0}} = f_{\theta}(\boldsymbol{x_t}, \boldsymbol{y_0}, t)$ denotes the $\boldsymbol{x_0}$ prediction from a neural network given $\boldsymbol{x_t}$, $\boldsymbol{y_0}$, and timestep $t$. The neural network is parameterized by weights $\theta$ and $\boldsymbol{\hat{\Delta}} = \boldsymbol{\hat{x}_0} - \boldsymbol{y_0}$ represents the $\boldsymbol{\Delta}$ estimation. Hence, the whole approximate reverse process is expressed by the following joint distribution:
\begin{equation}
\label{eq:reverse_process_approximation}
p_{\theta}(\boldsymbol{x_{0:T}}|\boldsymbol{y_0}) = p(\boldsymbol{x_{T}} | \boldsymbol{y_0})\prod_{t=1}^{T} p_{\theta}(\boldsymbol{x_{t-1}}|\boldsymbol{x_t}, \boldsymbol{y_0}).
\end{equation}



\subsection{Long-Range Reverse Transition}
\label{sec:long_range_reverse_transition}
Particularly, the derived reverse transition structurally matches the reparameterized form of the marginal $q(\boldsymbol{x_{t-1}}|\boldsymbol{x_0}, \boldsymbol{\Delta})$ (see Appendix \ref{app:reverse_process_transition_distribution}), which models the cumulative transitions from $\boldsymbol{x_0}$ to $\boldsymbol{x_{t-1}}$ in the forward process. Therefore, the reverse transition formulation aligns with the concept of cumulative transitions, allowing the reverse process to efficiently sample any state at an arbitrary timestep $\tau_{k-1} \in \{0, 1, \dots, T - 1\}$ by skipping intermediate latent variables in the reverse trajectory. Accordingly, the reverse process can be simulated with fewer timesteps, thereby accelerating sampling. Using the reparameterization trick, $\boldsymbol{x_{\tau_{k-1}}} \sim p_{\theta}\left(\boldsymbol{x_{\tau_{k-1}}} | \boldsymbol{x_{\tau_{k}}}, \boldsymbol{y_0}\right)$ can be sampled as follows:
\begin{equation}
\label{eq:reverse_process_long_range_transition_reparameterization}
\boldsymbol{x_{\tau_{k-1}}} =
\begin{cases}
    \boldsymbol{\hat{x}_{0}} - \beta_{\tau_{k-1}}\boldsymbol{\hat{\Delta}}, & \text{if } \gamma = 0, \\
    \boldsymbol{\hat{x}_{0}} - \beta_{\tau_{k-1}}\boldsymbol{\hat{\Delta}} + \sqrt{\gamma^2\beta_{\tau_{k-1}} - \eta^2\tilde{\lambda}_{\tau_{k}}}\boldsymbol{\hat{\epsilon}} + \sqrt{\eta^2\tilde{\lambda}_{\tau_{k}}}\boldsymbol{z}, & \text{if } \gamma \ne 0,
\end{cases}
\end{equation}
where $(\tau_{k-1}, \tau_k) \in \left\{(t', t) \in \mathbb{N}_{0}^{2} \mid t' + 1 \leq t \leq T \right\}$, $\boldsymbol{z} \sim \mathcal{N}\left(0, \boldsymbol{I}\right)$, and $\boldsymbol{\hat{\epsilon}}$ is expressed by the following relationship when $\gamma \ne 0$ (see Equation (\ref{eq:forward_process_cumulative_transition_reparameterization}) in Appendix \ref{app:reverse_process_transition_distribution}):
\begin{equation}
\label{eq:reverse_process_transition_closed_form_mean_non_zero_gamma_epsilon_reparameterization}
\boldsymbol{\hat{\epsilon}} = \frac{\boldsymbol{x_{\tau_{k}}} - \boldsymbol{\hat{x}_{0}} + \beta_{\tau_{k}}\boldsymbol{\hat{\Delta}}}{{\sqrt{\gamma^2\beta_{\tau_{k}}}}}.
\end{equation}
Essentially, each iteration of the reverse process involves predicting the original data sample, $\boldsymbol{x_0}$. This estimate is then used to compute the residual $\boldsymbol{\Delta}$ and the noise component $\boldsymbol{\epsilon}$, which together guide the update to the next less-degraded state, $\boldsymbol{x_{\tau_{k-1}}}$. As the reverse process progresses, the model gradually refines its prediction of $\boldsymbol{x_0}$ at each step, leveraging the increasingly accurate intermediate states. This iterative refinement culminates in an accurate prediction of $\boldsymbol{x_0}$. Moreover, the ability of the reverse process to skip intermediate steps not only enables few-step generation but also allows one-step predictions, thus demonstrating the efficiency and flexibility of the \ac{RDIM} sampling procedure. Here, the number of sampling timesteps along the reverse trajectory, $S \in \{1, 2, \dots, T\}$, is set arbitrarily. For each case, a uniform schedule is used, as detailed in Appendix \ref{app:uniform_sampling_timestep_scheduler}.

\subsection{\texorpdfstring{Residual $\boldsymbol{\beta}$-Schedule}{Residual Beta-Schedule}}
\label{sec:residual_beta_schedule}
\begin{figure*}[t]
    \centering
    \includegraphics[width=\textwidth]{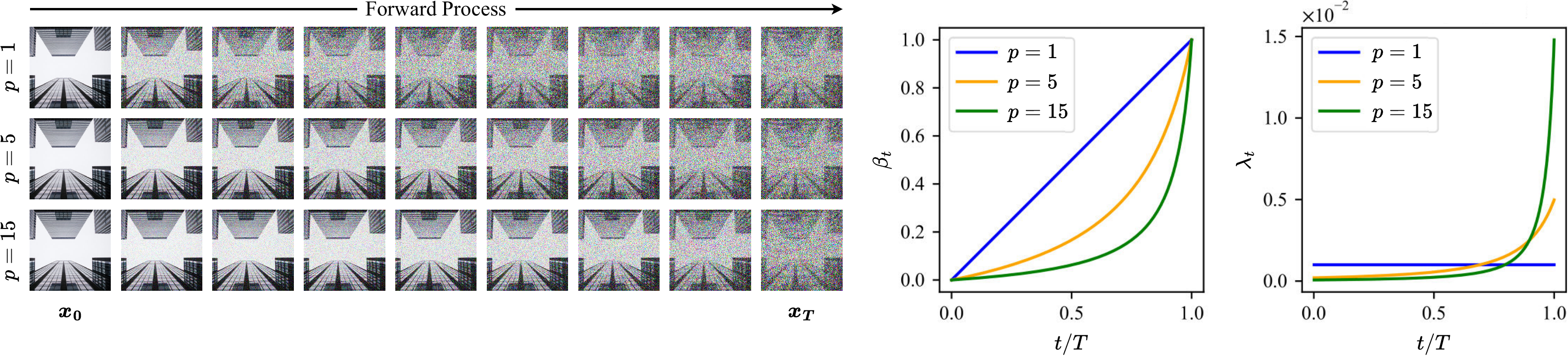}
    \caption{Progression of weights $\beta_t$ and $\lambda_t$ across timesteps and impact of $p$ on the diffusion process.}
    \label{fig:beta_schedule}
\end{figure*}

The residual $\beta$-schedule employed is defined by a circular curve (similar to the fourth quadrant $p$-norm shape), ensuring a smooth and adjustable transition between $\boldsymbol{x_0}$ and $\boldsymbol{x_T}$:
\begin{equation}
\label{eq:beta_schedule}
\beta_t = \frac{t}{T + \left(p - 1\right)\left(T - t\right)},
\end{equation}
where $p \in (0, \infty)$ is a parameter that allows controlling the steepness of the curve. As it increases the $\beta$-schedule exhibits a slower initial progression, followed by a rapid increase to larger and more pronounced updates. This design allows for a gentle removal of the residual and injection of noise in the early timesteps of the forward process, which become progressively more aggressive throughout the diffusion trajectory. Figure \ref{fig:beta_schedule} illustrates the impact of the parameter $p$ on the diffusion process. 


Furthermore, this choice for the $\beta$-schedule ensures that $\beta_{0} = 0$ and $\beta_{T} = 1$, such that the residual, $\boldsymbol{\Delta}$, is fully removed from $\boldsymbol{x_{0}}$ after exactly $T$ timesteps. As a result, the last latent variable, $\boldsymbol{x_T}$, converges to a noisy sample centered at the corrupted data, $\boldsymbol{y_0}$. 
Additionally, since $\beta_{0} = 0$, it follows that when $\gamma \ne 0$ the variance of any reverse transition distribution from $\boldsymbol{x_{\tau_{k}}}$ to $\boldsymbol{x_{0}}$ is $\eta^2\tilde{\lambda}_{\tau_{k}} = \eta^2\gamma^2\frac{\beta_{0}}{\beta_{\tau_{k}}}\lambda_{\tau_{k}} = 0$. Therefore, $p_{\theta}(\boldsymbol{x_{0}}|\boldsymbol{x_{\tau_{k}}}, \boldsymbol{y_0})_{\gamma \ne 0}$ degenerates into a $\delta$-distribution centered at $\boldsymbol{\hat{x}_{0}}$. Logically, under these conditions, $\eta$ does not have any impact on the last transition of the reverse process.

\subsection{Optimization}
\label{sec:optimization}
At each step of the sampling process, the neural network parameterized by weights $\theta$ yields an estimate of $\boldsymbol{x_0}$. During training, these parameters are learned to assure that the model marginal $p_{\theta}(\boldsymbol{x_{0}}|\boldsymbol{y_0})$ fits the true posterior distribution $q(\boldsymbol{x_0}|\boldsymbol{y_0})$ via:
\begin{equation}
\label{eq:target_data_distribution}
q(\boldsymbol{x_0}|\boldsymbol{y_0}) \approx p_{\theta}(\boldsymbol{x_{0}}|\boldsymbol{y_0}) = \int p(\boldsymbol{x_{T}} | \boldsymbol{y_0})\prod_{t=1}^{T} p_{\theta}(\boldsymbol{x_{t-1}}|\boldsymbol{x_t}, \boldsymbol{y_0}) \, d\boldsymbol{x_{1:T}},
\end{equation}
which ensures, during inference, that the data, $\boldsymbol{x_{0}}$, can be sampled back accurately given $\boldsymbol{y_0}$. Accordingly, $p_{\theta}(\boldsymbol{x_{t-1}}|\boldsymbol{x_t}, \boldsymbol{y_0})$ is required to closely approximate the true forward transition posterior, $q(\boldsymbol{x_{t-1}}|\boldsymbol{x_{t}}, \boldsymbol{x_0}, \boldsymbol{\Delta})$. This is achieved by minimizing the Kullback–Leibler (KL) divergence between both distributions, while accounting for all timesteps. In fact, this objective can be reduced for simplicity to (see Appendix \ref{app:training objective}):
\begin{equation}
\label{eq:training_objective_function_simplified}
\mathcal{L}_{\text{simple}}(\theta) = \mathbb{E}_{\boldsymbol{x_0}, \boldsymbol{\Delta}, t}\left[\|\boldsymbol{x_0} - \boldsymbol{\hat{x}_{0}}\|^2\right].
\end{equation}

Notably, ResShift shares the same training objective as \ac{RDIM}, further highlighting \mbox{that ResShift is} a particular case of \ac{RDIM} and that its trained models can be used for \ac{RDIM} sampling without retraining. The \ac{RDIM} training and sampling procedures are described in Algorithms \ref{alg:rdim_training} and \ref{alg:rdim_sampling}, respectively.

\noindent
\begin{minipage}[t]{0.48\textwidth}
\begin{algorithm}[H]
\caption{Training}\label{alg:rdim_training}
\small
\begin{algorithmic}[1]
\Repeat
    \State $\boldsymbol{x_0}, \boldsymbol{y_0} \sim q(\boldsymbol{x_0}, \boldsymbol{y_0}) = q(\boldsymbol{x_0})q(\boldsymbol{y_0}|\boldsymbol{x_0})$
    \State $\boldsymbol{\Delta} = \boldsymbol{x_0} - \boldsymbol{y_0}$
    \State $t \sim \mathcal{U}(1, T)$
    \State $\boldsymbol{\epsilon} \sim \mathcal{N}(0, \boldsymbol{I})$
    \State $\boldsymbol{x_t} \sim q(\boldsymbol{x_t}|\boldsymbol{x_0}, \boldsymbol{\Delta})$
    \State $\boldsymbol{\hat{x}_{0}} = f_{\theta}(\boldsymbol{x_t}, \boldsymbol{y_0}, t)$
    \State $\mathcal{L} = \|\boldsymbol{x_0} - \boldsymbol{\hat{x}_{0}}\|^2$
    \State Take gradient descent step on $\nabla_{\theta} \mathcal{L}$
\Until convergence
\State \Return $f_{\theta}$
\end{algorithmic}
\end{algorithm}
\end{minipage}
\hfill
\begin{minipage}[t]{0.48\textwidth}
\begingroup
\renewcommand{\baselinestretch}{1.164}\normalsize
\begin{algorithm}[H]
\caption{Sampling}\label{alg:rdim_sampling}
\small
\begin{algorithmic}[1]
\State $\Upsilon = \left\{\tau_{S} = T, \tau_{S - 1}, \dots, \tau_{1}, \tau_{0} = 0\right\}$
\State $\boldsymbol{x_T} \sim \mathcal{N}(\boldsymbol{y_0}, \gamma^2\boldsymbol{I})$
\For{$k = S, S-1, \dots, 1$}
    \State $\boldsymbol{\hat{x}_{0}} = f_{\theta}(\boldsymbol{x_{\tau_{k}}}, \boldsymbol{y_0}, \tau_{k})$
    \State $\boldsymbol{\hat{\Delta}} = \boldsymbol{\hat{x}_{0}} - \boldsymbol{y_0}$
    \State \textbf{if} $\gamma \ne 0$ \textbf{then} $\boldsymbol{\hat{\epsilon}} = \frac{\boldsymbol{x_{\tau_{k}}} - \boldsymbol{\hat{x}_{0}} + \beta_{\tau_{k}}\boldsymbol{\hat{\Delta}}}{{\sqrt{\gamma^2\beta_{\tau_{k}}}}}$
    \State $\boldsymbol{x_{\tau_{k-1}}} \sim p_{\theta}\left(\boldsymbol{x_{\tau_{k-1}}} | \boldsymbol{x_{\tau_{k}}}, \boldsymbol{y_0}\right)$
\EndFor
\State \Return $\boldsymbol{x_{0}}$
\end{algorithmic}
\end{algorithm}
\endgroup
\end{minipage}

\section{Experiments}
\label{sec:experiments}
\ac{RDIM} is evaluated on image denoising and single image \ac{SR} using the FMD \citep{zhang2019poisson}, SIDD \citep{abdelhamed2018high, abdelhamed2019ntire}, and DIV2K \citep{agustsson2017ntire, timofte2017ntire} datasets. \hl{Several} \ac{RDIM} variants with $\gamma\,{=}\,3.0$, $\eta\,{=}\,1.0$, and $p\,{=}\,5.0$ are considered, differing only in the number of sampling timesteps and loss targets. RDIM-PQ stands for RDIM trained with a perceptual quality (PQ) objective. RDIM-$S$ and RDIM-PQ-$S$ denote sampling with $S$ steps. Particularly, \ac{RDIM}-$1$ and RDIM-PQ-$1$ correspond to single-step deterministic inferences ($S\,{=}\,1$). Their deterministic nature results from the final reverse transition degenerating into a $\delta$-distribution when $\gamma \ne 0$ and $\beta_0 = 0$ (see Sections \ref{sec:long_range_reverse_transition} and \ref{sec:residual_beta_schedule}). Moreover, \ac{RDIM} is compared against \hl{multiple state-of-the-art methods, including} ResShift with $S\,{=}\,T\,{=}\,100$. Although ResShift is often employed with $S\,{=}\,T\,{=}\,10$, there is a significant performance improvement when using longer diffusion chains. This effect is evident in the experiments shown in Appendix \ref{app:comparative_analysis_of_multiple_rdim_configurations_in_image_denoising}, where ResShift improves \ac{PSNR} from $39.363$ dB for $T\,{=}\,10$ to $43.599$ dB for $T\,{=}\,100$. Additional qualitative results on image inpainting, colorization, and deblurring are provided on FFHQ \citep{karras2019style}. Experimental details are in Appendix \ref{app:experimental_details_and_additional_results}, including RDIM \mbox{assessment when varying $S$ (Figure~\ref{fig:results_fmd_bpae_sampling_timesteps}).}

\begin{table}[b!]
  \caption{Comparative analysis of RDIM against relevant state-of-the-art techniques for (a) denoising and (b) SR. \textcolor{ForestGreen}{Green color} highlights the best score overall and \textcolor{RoyalBlue}{Blue color} the second best.}
  \centering
  \begin{subtable}{0.55\linewidth}
  \caption{Denoising on images from the FMD (BPAE and zebrafish confocal fluorescence microscopy images) and SIDD datasets.}
  \label{tab:results_fmd_sidd}
  \scalebox{0.67}{%
  \begin{tabular}{l@{\quad}cc@{\quad}cc@{\quad}cc@{\quad}c}
    \toprule
    \multirow{2.3}{*}{\minipage{1cm} \bf Denoising\\ \bf Method\endminipage} & \multirow{2.6}{*}{S$\downarrow$} & \multicolumn{2}{c}{FMD-BPAE} & \multicolumn{2}{c}{FMD-Zebrafish} & \multicolumn{2}{c}{SIDD-Medium} \\
    \cmidrule(lr){3-4}
    \cmidrule(lr){5-6}
    \cmidrule(lr){7-8}
    & & PSNR$\uparrow$ & SSIM$\uparrow$ &  PSNR$\uparrow$ & SSIM$\uparrow$ & PSNR$\uparrow$ & SSIM$\uparrow$ \\
    \midrule
    Noisy & -- & 31.596 & 0.812 & 26.732 & 0.603 & 27.797 & 0.515 \\
    BM3D & -- & 35.862 & 0.933 & 35.289 & 0.918 & 35.880 & 0.906 \\
    DnCNN & -- & 37.609 & 0.950 &  37.169 & 0.941 & 39.838 & 0.957 \\
    DDPM & 100 & 41.775 & 0.981 & 43.214 & 0.974 & 39.329 & 0.945 \\
    DDIM-25 & 25 & 35.168  & 0.953 & 39.060  & 0.960 & 28.627  & 0.855 \\
    DDIM-50 & 50 & 38.608  & 0.972 & 41.211  & 0.969 & 34.665  & 0.912 \\
    ResShift & 100 & 43.599 & 0.984 & \textcolor{ForestGreen}{\textbf{45.167}} & \textcolor{RoyalBlue}{0.976} & 39.663 & 0.949 \\
    \midrule
    RDIM-$1$ & \textcolor{ForestGreen}{\textbf{1}} & \textcolor{RoyalBlue}{43.987} & \textcolor{RoyalBlue}{0.985} & 44.229 & \textcolor{RoyalBlue}{0.976} & \textcolor{ForestGreen}{\textbf{40.335}} & \textcolor{ForestGreen}{\textbf{0.962}} \\
    RDIM-$10$ & \textcolor{RoyalBlue}{10} & \textcolor{ForestGreen}{\textbf{44.147}} & \textcolor{ForestGreen}{\textbf{0.986}} &  \textcolor{RoyalBlue}{45.027} & \textcolor{ForestGreen}{\textbf{0.978}} & \textcolor{RoyalBlue}{39.979} & \textcolor{RoyalBlue}{0.958} \\
    \bottomrule
  \end{tabular}}
  \end{subtable}
  \hfill
  \begin{subtable}{0.42\linewidth}
  \caption{$\times2$ and $\times4$ \ac{SR} on images from the DIV2K dataset\hl{ under unknown degradations}.}
  \label{tab:results_div2k}
  \scalebox{0.67}{%
  \begin{tabular}{l@{\quad}cc@{\quad}cc@{\quad}c}
    \toprule
    \multirow{2.3}{*}{\minipage{1cm} \bf SR\\ \bf Method\endminipage} & \multirow{2.6}{*}{S$\downarrow$} & \multicolumn{2}{c}{DIV2K-$\times2$} & \multicolumn{2}{c}{DIV2K-$\times4$} \\
    \cmidrule(lr){3-4}
    \cmidrule(lr){5-6}
    & & PSNR$\uparrow$ & SSIM$\uparrow$ &  PSNR$\uparrow$ & SSIM$\uparrow$ \\
    \midrule
    LR (Bicubic) & -- &  25.112 & 0.704 & 21.742 & 0.574 \\
    ESRGAN & -- & 30.017 & 0.857 & 24.957 & 0.690 \\
    DDPM & 100 & 31.949 & 0.893 & 26.446 & 0.739 \\
    DDIM-$25$ & 25 & 29.003 & 0.839 & 20.894 & 0.504 \\
    DDIM-$50$ & 50 & 31.150 & 0.879 & 24.949 & 0.687 \\
    ResShift & 100 & 32.368 & 0.903 & 26.627 & 0.750 \\
    \midrule
    RDIM-$1$ & \textcolor{ForestGreen}{\textbf{1}} & \textcolor{ForestGreen}{\textbf{33.887}} & \textcolor{ForestGreen}{\textbf{0.924}} & \textcolor{ForestGreen}{\textbf{28.280}} & \textcolor{ForestGreen}{\textbf{0.798}} \\
    RDIM-$10$ & \textcolor{RoyalBlue}{10} & \textcolor{RoyalBlue}{33.019} & \textcolor{RoyalBlue}{0.914} & \textcolor{RoyalBlue}{27.266} & \textcolor{RoyalBlue}{0.770} \\
    \bottomrule
  \end{tabular}}
  \end{subtable}
\end{table}

\paragraph{Image Denoising.} \ac{RDIM} is compared against BM3D \citep{dabov2007image}, DnCNN \citep{zhang2017beyond}, \ac{DDPM} \citep{ho2020denoising}, \hl{DDIM} \citep{song2020denoising}, and ResShift \citep{yue2023resshift}. 
For fairness, all diffusion models use the same network architecture (see Appendix \ref{app:network_architecture}) and diffusion timesteps ($T=100$). 
Results are listed in Table \ref{tab:results_fmd_sidd}. On FMD-BPAE, \ac{RDIM}-$10$ achieves the best results in terms of \ac{PSNR} and \ac{SSIM}, followed by \ac{RDIM}-$1$. On FMD-Zebrafish, ResShift attains the best \ac{PSNR} score, but is $10\times$ slower than \ac{RDIM}-$10$, which obtains comparable \ac{PSNR} performance and the best \ac{SSIM} score. On SIDD-Medium, \ac{RDIM}-$1$ yields superior results. Diffusion models, which inherently capture richer structures than DnCNN, have their gains diminished on SIDD-Medium due to the use of a small $64{\times}64$ patch size (which is kept the same across all experiments for consistency). \hl{Meanwhile, DnCNN performs full-image processing at inference, giving it a slight unfair advantage.} Figure \ref{fig:results_sidd}, Appendix~\ref{app:additional_qualitative_results}, presents a qualitative comparison.

\paragraph{Super-Resolution.} A comparative analysis with $\times2$ and $\times4$ downsampling factors evaluates \ac{RDIM} against ESRGAN \citep{wang2018esrgan}, \ac{DDPM}, \hl{DDIM,} and ResShift. As before, diffusion models were trained under the same conditions, including architecture and diffusion timesteps ($T=100$). Results are shown in Table \ref{tab:results_div2k}. On both DIV2K-Unknown-$\times2$ and DIV2K-Unknown-$\times4$, \ac{RDIM}-$1$ performs the best, followed by \ac{RDIM}-$10$, highlighting that \ac{RDIM} consistently surpasses ResShift and \ac{DDPM}. Figure \ref{fig:results_div2k}, in Appendix~\ref{app:additional_qualitative_results} showcases qualitative results. 

\begin{figure}[h]
    \centering
    \begin{minipage}[t]{0.59\textwidth}
        \vspace{0pt}
        \centering
        \includegraphics[width=\linewidth]{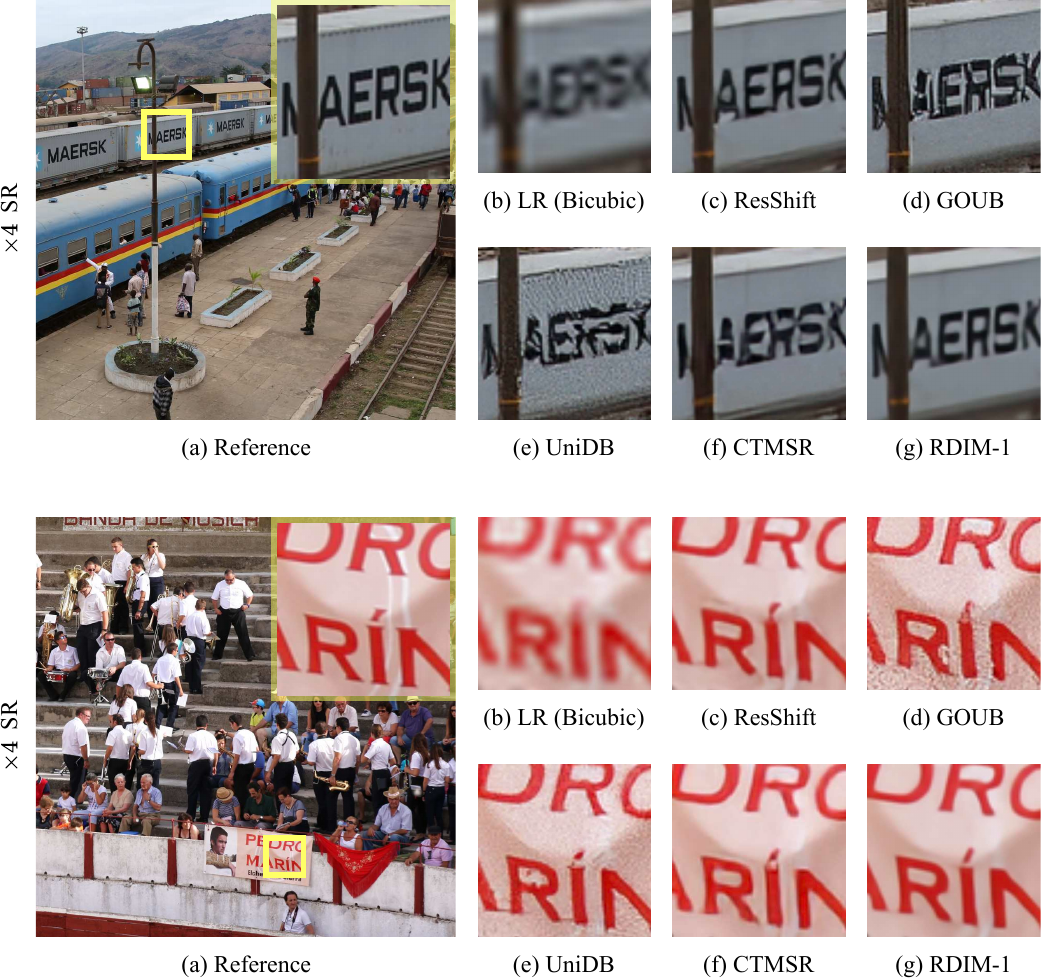}
        \caption{Qualitative $\times4$ \ac{SR} results on DIV2K-Bicubic-$\times4$.}
        \label{fig:results_div2k_x4_bicubic}
    \end{minipage}
    \hfill
    \begin{minipage}[t]{0.3685\linewidth}
        \vspace{0pt}
        \centering
        \captionof{table}{$\times4$ \ac{SR} on bicubic downsampled images from the DIV2K dataset.}
        \label{tab:results_div2k_x4}
        \resizebox{\linewidth}{!}{
            \begin{tabular}{l@{\quad}cc@{\quad}c@{\quad}c}
            \toprule
            \multirow{2.3}{*}{\minipage{1cm} \bf SR\\ \bf Method\endminipage} & \multirow{2.6}{*}{S$\downarrow$} & \multicolumn{3}{c}{DIV2K-$\times4$} \\
            \cmidrule(lr){3-5}
            & & PSNR$\uparrow$ & SSIM$\uparrow$ & LPIPS$\downarrow$ \\
            \midrule
            DDRM$^\dagger$ & 100 & 24.350 & 0.592 & 0.364 \\
            ResShift & 100 & 27.455 & 0.780 & 0.153 \\
            IR-SDE$^\dagger$ & 100 & 25.900 & 0.657 & 0.231 \\
            DDBM$^\ddagger$ & 100 &  24.210 & 0.581 & 0.384 \\
            GOUB-SDE$^\dagger$ & 100 & 26.890 & 0.748 & 0.220 \\
            GOUB-ODE$^\dagger$ & 100 & 28.500 & 0.807 & 0.328 \\
            UniDB-SDE$^\dagger$ & 100 & 25.460 & 0.686 & 0.179 \\
            UniDB-ODE$^\dagger$ & 100 & 28.640 & 0.807 & 0.323 \\
            UniDB++-$50$$^\ddagger$ & 50 & 26.610 & 0.754 & 0.159 \\
            UniDB++-$20$$^\ddagger$ & 20 & 27.380 & 0.777 & 0.179 \\
            UniDB++-$5$$^\ddagger$ & 5 & 28.400 & 0.805 & 0.235 \\
            MaRS-$5$$^\ddagger$ & 5 & 27.730 & 0.783 & 0.286 \\
            DBIM-$5$$^\ddagger$ & 5 & 28.050 & 0.795 & 0.260 \\
            CTMSR-$1$ & \textbf{1} & 27.087 & 0.759 & 0.130 \\
            \midrule
            RDIM-$1$ & \textbf{1} & \textbf{29.180} & \textbf{0.824} & 0.257 \\
            RDIM-$5$ & 5 & 28.408 & 0.806 & 0.197 \\
            RDIM-$10$ & 10 & 27.963 & 0.795 & 0.178 \\
            RDIM-$20$ & 20 & 27.636 & 0.786 & 0.166 \\
            RDIM-$50$ & 50 & 27.427 & 0.779 & 0.154 \\
            \midrule
            RDIM-PQ-$1$ & \textbf{1} & 29.004 & 0.817 & \textbf{0.114} \\
            \bottomrule \\[-0.85em]
            \multicolumn{5}{l}{$^\dagger$~Retrieved from \citet{zhu2025unidb}.}\\
            \multicolumn{5}{l}{$^\ddagger$~Retrieved from \citet{pan2025unidb++}.}
        \end{tabular}}
    \end{minipage}
\end{figure}

An additional analysis is conducted on $\times4$ bicubic downsampled images from the DIV2K dataset, comparing RDIM against DDRM \citep{kawar2022denoising}, ResShift, IR-SDE \citep{luo2023image}, DDBM \citep{zhou2024denoising}, GOUB \citep{yue2024image}, UniDB \citep{zhu2025unidb}, CTMSR \citep{you2025consistency}, MaRS \citep{li2025mars}, DBIM \citep{zheng2024diffusion}, and UniDB++ \citep{pan2025unidb++}. 
Results are presented in Table \ref{tab:results_div2k_x4}. RDIM-$1$ achieves the highest PSNR and SSIM among all methods, with RDIM-PQ-$1$ following closely. This suggests that RDIM offers an advantage for applications where distortion fidelity is critical (e.g., medical imaging). Notably, RDIM-PQ-$1$ explicitly optimizes for LPIPS and attains the lowest score on this metric while attaining high PSNR and SSIM. Overall, a clear perception–distortion trade-off emerges, as further illustrated in Appendix \ref{appendix:perception-distortion trade-off}, where RDIM demonstrates a more favorable balance than state-of-the-art alternatives.




A qualitative comparison in Figure \ref{fig:results_div2k_x4_bicubic} further demonstrates that RDIM yields sharper and more faithful reconstructions, particularly in areas rich in fine textures and structural detail. Other methods introduce noticeable artifacts and deformations. More results are shown in Figures \ref{fig:results_div2k_x4_bicubic_additional_1} \mbox{and \ref{fig:results_div2k_x4_bicubic_additional_2} (Appendix \ref{app:additional_qualitative_results}).}

\begin{figure*}[h]
    \centering
    \includegraphics[width=\textwidth]{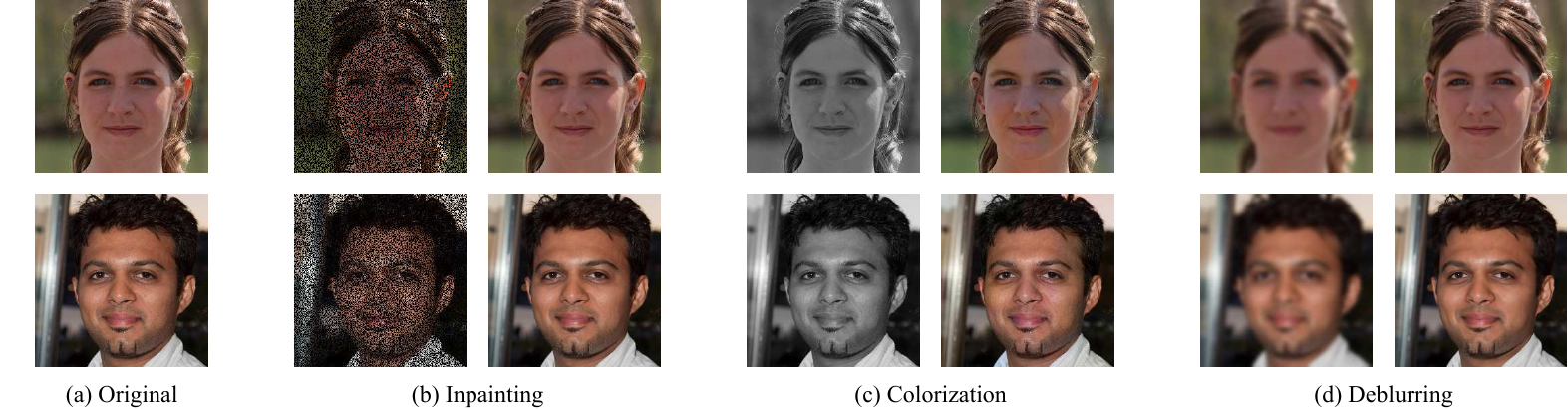}
    \caption{\ac{RDIM}-$10$ results in image inpainting, colorization, and deblurring on the FFHQ dataset. In (b), (c) and (d), the left side represents the input image and the right side the output.}
    \label{fig:results_ffhq}
\end{figure*}

\paragraph{Additional Image Restoration Tasks.} Further evaluation on image inpainting, colorization, and deblurring tasks demonstrates the generalization capabilities of \ac{RDIM}. Figure \ref{fig:results_ffhq} presents qualitative results obtained with \ac{RDIM}-$10$. Additional \hl{details and results} are provided in Appendix \ref{app:experimental_details_and_additional_results}.

\paragraph{Discussion of results.} \ac{RDIM} and ResShift consistently outperform \ac{DDPM}, emphasizing that their diffusion process is  more closely aligned with these inverse problems. Moreover, \ac{RDIM} demonstrates performance comparable to ResShift, often surpassing it, while requiring significantly fewer sampling timesteps. \hl{This stands in contrast with the DDIM behavior when applied to a pretrained DDPM. DDIM incurs noticeable degradation when reducing the sampling count from $S = 100$ to $S = 50$, and does not support reliable single-step inference. This suggests that the residual modeling dynamics are intrinsically well aligned with implicit sampling, enabling acceleration factors (up to $\times 100$) that are not achievable by applying DDIM to DDPM.} Moreover, since \ac{DDPM} and ResShift require a reverse process with the same number of timesteps as their forward diffusion process, reducing their diffusion steps to match the \ac{RDIM} sampling time would result in a degradation in performance \citep{shih2023parallel}. This effect is evident in the experiments conducted in Appendix \ref{app:comparative_analysis_of_multiple_rdim_configurations_in_image_denoising}, where ResShift with a reduced number of diffusion \hl{steps} underperforms compared to its \mbox{higher-timestep configurations.}

Furthermore, FMD-Zebrafish contains noisier images than FMD-BPAE. As shown in Table \ref{tab:results_fmd_sidd}, \ac{RDIM}-$1$ outperforms ResShift on FMD-BPAE, whereas ResShift performs better on FMD-Zebrafish. This suggests that in the presence of stronger degradations a more stochastic approach is advantageous, as variability promotes output diversity. Conversely, when degradations are mild, a more deterministic method ensures consistent and accurate restoration. Therefore, balancing stochasticity is crucial to adapt the method effectively to varying noise levels and degradation strengths. Notably, \ac{RDIM}-$10$ achieves comparable results to ResShift in FMD-Zebrafish while requiring only $10$ sampling steps instead of $100$, rendering inference $10\times$ faster. Further demonstrating its efficiency, \ac{RDIM} accelerates sampling up to $100\times$ compared to ResShift and \ac{DDPM} on FMD-BPAE. Additionally, experiments on SIDD highlight that \ac{RDIM} effectively supports \ac{HR} image reconstruction even when operating on relatively small patches (e.g., $64 \times 64$) compared to the full image size, which here reach resolutions of up to $\approx 5300 \times 3000$ pixels. Naturally, increasing the patch size will improve performance and could enable restoration of images at even higher resolutions.

In \ac{SR}\hl{ under unknown degradations}, standard diffusion models and ResShift, often exhibit a tendency to hallucinate details that deviate from the ground truth, particularly when employing long diffusion chains. As illustrated in Figure \ref{fig:results_div2k}, while iterative refinement encourages the generation of natural-looking textures, it frequently trades off fidelity for perceptual quality, leading to reconstructions that drift away from the original structure (see Appendix \ref{app:comparative_analysis_of_multiple_rdim_configurations_in_image_denoising} for further evidence). Furthermore, the deterministic \mbox{\ac{RDIM}-$1$} outperforms all methods, suggesting a more deterministic approach to \ac{SR} is beneficial, as too much stochasticity can introduce unwanted variability in the output and the iterative refinement of long-chain diffusion can become detrimental. \hl{A similar trend is observed on the DIV2K-Bicubic-$\times4$ benchmark. As shown in Table \mbox{\ref{tab:results_div2k_x4}}, RDIM again achieves the highest PSNR and SSIM scores while operating with far fewer sampling steps. It maintains sharper and more faithful textures, as illustrated in Figure \mbox{\ref{fig:results_div2k_x4_bicubic}}. These results confirm that the advantages of residual-based implicit sampling carry over to classical SR settings.}

\section{Related Work}
\label{sec:related_work}

\noindent\textbf{Diffusion Models} (DMs)
~\citep{sohl2015deep,song2019generative,song2020score} generate images by iteratively denoising latent variables sampled from a Gaussian prior. For image-to-image tasks, conditioning mechanisms such as classifier guidance~\citep{dhariwal2021diffusion} or classifier-free guidance~\citep{ho2022classifier} enable the generation of target images given source observations~\citep{saharia2022palette,sasaki2021unit,zhao2022egsde}.
However, because DMs start from pure noise, they remain misaligned with inverse problems where the input already contains meaningful structure. Hence, several diffusion-based reconstruction approaches adapt the generative process to low-quality inputs. SR3~\citep{saharia2022image} and SRDiff~\citep{li2022srdiff} condition DDPMs on low-resolution inputs, while~\citet{whang2022deblurring} use residual-based refinements to improve deblurring. DDRM~\citep{kawar2022denoising} addresses general linear inverse problems via posterior sampling with a pretrained DM, and ResShift~\citep{yue2023resshift} leverages residual modeling between high-resolution and low-resolution images. Despite their effectiveness, these methods still require traversing all diffusion steps sequentially. RDIM generalizes residual modeling while enabling DDIM-style long-range sampling and controllable stochasticity, significantly reducing the number of steps needed for high-quality reconstruction.

\noindent\textbf{Accelerating DM Sampling}
have become an attractive research area, usually focusing on reducing the number of steps to a dozen or fewer. Within the body of work, training-based distillation approaches~\citep{salimans2022progressive,luhman2021knowledge,song2023consistency,meng2023distillation,li2023snapfusion,luo2023latent,kim2023consistency} compress long trajectories into few-step solvers, while training-free methods leverage \acp{ODE}, e.g. DDIM~\citep{song2020denoising}, DPM-Solver~\citep{popov2021diffusion,bao2022analytic,lu2025dpm,zheng2023dpm}. These methods enable fast sampling but remain primarily designed for unconditional synthesis and do not explicitly align the forward dynamics with the degradation process. In contrast, RDIM targets paired inverse problems by explicitly modeling residuals and allowing few-step implicit updates.

%
\noindent\textbf{Flow Models} (FMs)
~\citep{albergo2023stochastic,do2024variational,lipman2022flow,liu2022flow} learn deterministic \ac{ODE} flows between arbitrary distributions using the flow matching objective~\citep{lipman2022flow}, which is closely related to DM's score matching~\citep{song2020score}. They can be understood as zero-variance limits of diffusion bridges, producing transport maps for unpaired or cross-domain translation. However, the deterministic nature of these flows restricts their capacity to capture uncertainty, an important property for restoration tasks involving strong degradations. RDIM differs by maintaining stochastic residual modeling with controllable variance, which empirically improves robustness and generalization.

\noindent\textbf{Bridge Models} (BMs)
can be categorized into Schrödinger bridges (SB) and diffusion bridges (DB). The former constructs a stochastic process connecting two arbitrary marginal distributions \citep{de2021diffusion,chen2021likelihood,liu20232}, while the later conditions a \ac{SDE} on fixed endpoints~\citep{heng2111simulating,li2023bbdm,zhou2024denoising}. Both methods have been exploited for image-to-image translation. In particular, DDBM \citep{zhou2024denoising} learns to simulate the time-reversal of a DB based on score matching. \citet{li2023bbdm} (BBDM) instead focus on constructing a Brownian bridge. SBALIGN~\citep{somnath2023aligned} and $\Omega$-Bridge~\citep{liu2023learning} use Doob’s $h$-transform to guide trajectories toward prescribed terminal states, and GOUB~\citep{yue2024image} incorporates a mean-reverting Ornstein–Uhlenbeck (OU) bridge to improve stability.

Although BM are powerful, SBs operate in unpaired settings, not enforcing or learning correspondences between samples. DBs impose strict boundary conditioning on endpoints, which can bias trajectories toward smoother transitions, blurring high-frequency details~\citep{kieu2025bidirectional}. In contrast,
RDIM is bridge-like in that its forward process defines a stochastic interpolation between $x_0$ and $y_0$, but it differs fundamentally from SB/DB frameworks: RDIM operates in paired settings, relaxes endpoint constraints, and uses an implicit DDIM-style sampler that supports step skipping.

%
\noindent\textbf{Stochastic optimal control} (SOC) has recently been adopted to steer diffusion trajectories. DIS~\citep{berner2023optimal} formalized the connection between SOC and diffusion, while RB-Modulation~\citep{rout2024rb} applied SOC principles for training-free style transfer. UniDB~\citep{zhu2025unidb} integrates SOC with diffusion bridges using penalty terms to guide forward trajectories toward terminal states, improving perceptual quality. RDIM differs by requiring neither fixed endpoints nor SOC penalties, instead relying on a flexible residual-based forward process that remains aligned with practical degradations.

\textbf{Alternative methods} 
to standard diffusion processes have also been explored. Inversion-by-direct-iteration (IDI)~\citep{delbracioinversion} replaces the stochastic denoising trajectory with a fixed-point iterative scheme, offering competitive restoration without a diffusion process. DiracDiffusion~\citep{fabian2024diracdiffusion} proposes a deterministic, data-consistent update rule that incrementally reconstructs images using Dirac-like propagation rather than probabilistic diffusion. Residual Denoising Diffusion Models (RDDM)~\citep{liu2024residual} incorporate residual learning into the diffusion process to accelerate convergence and reduce the dependency on long sampling chains. Iterative $\alpha$-(de)blending~\citep{heitz2023iterative} introduces a minimalist deterministic diffusion variant based on recursive blending operations, enabling efficient incremental reconstruction. These approaches share with RDIM the motivation of improving reconstruction fidelity and reducing sampling cost, but differ fundamentally in that RDIM preserves the generative diffusion structure while aligning the forward process with the degradation model and enabling DDIM-style long-range sampling.


\section{Conclusion}
\label{sec:conclusion}
\acp{RDIM} constitute a diffusion framework tailored for inverse problems that explicitly models the residuals between \ac{HQ} and \ac{LQ} images. Aligning the forward process with the actual degradation and leveraging implicit sampling enables \acp{RDIM} to produce accurate reconstructions with significantly fewer steps than conventional \acp{DDPM}. Furthermore, \ac{RDIM} achieves superior results compared to \ac{DDPM}, reducing hallucinations while maintaining fidelity, highlighting that starting the reverse process closer to the \ac{LQ} images offers a more informed and effective initialization. Experiments on denoising and \ac{SR} demonstrate consistent improvements over \acp{DDPM} and performance comparable to or exceeding ResShift, achieving \ac{HQ} results with single or few step inference. These results establish \acp{RDIM} as an efficient and versatile approach for a wide range of image reconstruction tasks.

\section*{Acknowledgments}
This work was supported by national funds through Fundação para a Ciência e a Tecnologia, I.P. (FCT), under the research units INESC-ID (\href{https://doi.org/10.54499/UID/50021/2025}{UID/50021/2025}), LARSyS (\href{https://doi.org/10.54499/UID/50009/2025}{UID/50009/2025}), and LASIGE (\href{https://doi.org/10.54499/UID/00408/2025}{UID/00408/2025}), as well as Advanced Computing Project 2025.7643.CPCA.A1 and PhD research grant (\href{https://doi.org/10.54499/2023.02827.BD}{2023.02827.BD}).

\bibliography{iclr2026_conference}

@inproceedings{sohl2015deep,
  title={Deep unsupervised learning using nonequilibrium thermodynamics},
  author={Sohl-Dickstein, Jascha and Weiss, Eric and Maheswaranathan, Niru and Ganguli, Surya},
  booktitle={International conference on machine learning (ICML)},
  pages={2256--2265},
  year={2015},
}

@inproceedings{ho2022classifier,
  title={Classifier-Free Diffusion Guidance},
  author={Ho, Jonathan and Salimans, Tim},
  booktitle={NeurIPS 2021 Workshop on Deep Generative Models and Downstream Applications}
}

@inproceedings{saharia2022palette,
  title={Palette: Image-to-image diffusion models},
  author={Saharia, Chitwan and Chan, William and Chang, Huiwen and Lee, Chris and Ho, Jonathan and Salimans, Tim and Fleet, David and Norouzi, Mohammad},
  booktitle={ACM SIGGRAPH 2022 conference proceedings},
  pages={1--10},
  year={2022}
}

@article{sasaki2021unit,
  title={{UNIT-DDPM}: Unpaired image translation with denoising diffusion probabilistic models},
  author={Sasaki, Hiroshi and Willcocks, Chris G and Breckon, Toby P},
  journal={arXiv preprint arXiv:2104.05358},
  year={2021}
}

@article{zhao2022egsde,
  title={{EGSDE}: Unpaired image-to-image translation via energy-guided stochastic differential equations},
  author={Zhao, Min and Bao, Fan and Li, Chongxuan and Zhu, Jun},
  journal={Advances in Neural Information Processing Systems (NeurIPS)},
  volume={35},
  pages={3609--3623},
  year={2022}
}

@article{albergo2023stochastic,
  title={Stochastic interpolants: A unifying framework for flows and diffusions, 2023},
  author={Albergo, Michael S and Boffi, Nicholas M and Vanden-Eijnden, Eric},
  journal={URL https://arxiv. org/abs/2303.08797},
  volume={3},
  year={2023}
}

@article{do2024variational,
  title={Variational Flow Models: Flowing in Your Style},
  author={Do, Kien and Kieu, Duc and Nguyen, Toan and Nguyen, Dang and Le, Hung and Nguyen, Dung and Nguyen, Thin},
  journal={arXiv preprint arXiv:2402.02977},
  year={2024}
}

@inproceedings{lipman2022flow,
  title={Flow Matching for Generative Modeling},
  author={Lipman, Yaron and Chen, Ricky TQ and Ben-Hamu, Heli and Nickel, Maximilian and Le, Matthew},
  booktitle={International Conference on Learning Representations (ICLR)},
  year=2023
}

@inproceedings{liu2022flow,
  title={Flow Straight and Fast: Learning to Generate and Transfer Data with Rectified Flow},
  author={Liu, Xingchao and Gong, Chengyue and others},
  booktitle={International Conference on Learning Representations (ICLR)},
  year={2023}
}

@inproceedings{liu20232,
  title={{I$^2$SB: Image-to-Image Schr\"odinger Bridge}},
  author={Liu, Guan-Horng and Vahdat, Arash and Huang, De-An and Theodorou, Evangelos and Nie, Weili and Anandkumar, Anima},
  booktitle={International Conference on Machine Learning (ICML)},
  pages={22042--22062},
  year={2023},
  organization={PMLR}
}

@article{de2021diffusion,
  title={Diffusion {Schr{\"o}dinger} bridge with applications to score-based generative modeling},
  author={De Bortoli, Valentin and Thornton, James and Heng, Jeremy and Doucet, Arnaud},
  journal={Advances in neural information processing systems (NeurIPS)},
  volume={34},
  pages={17695--17709},
  year={2021}
}

@inproceedings{chen2021likelihood,
  title={Likelihood training of {Schr\"odinger} bridge using forward-backward {SDE}s theory},
  author={Chen, Tianrong and Liu, Guan-Horng and Theodorou, Evangelos A},
  booktitle={International Conference on Learning Representations (ICLR)},
  year={2022}
}

@inproceedings{li2023bbdm,
  title={{BBDM}: Image-to-image translation with brownian bridge diffusion models},
  author={Li, Bo and Xue, Kaitao and Liu, Bin and Lai, Yu-Kun},
  booktitle={IEEE/CVF conference on computer vision and pattern Recognition (CVPR)},
  pages={1952--1961},
  year={2023}
}

@inproceedings{somnath2023aligned,
  title={Aligned diffusion schr{\"o}dinger bridges},
  author={Somnath, Vignesh Ram and Pariset, Matteo and Hsieh, Ya-Ping and Martinez, Maria Rodriguez and Krause, Andreas and Bunne, Charlotte},
  booktitle={Uncertainty in Artificial Intelligence},
  pages={1985--1995},
  year={2023},
  organization={PMLR}
}

@inproceedings{liu2023learning,
  title={Learning diffusion bridges on constrained domains},
  author={Liu, Xingchao and Wu, Lemeng and Ye, Mao and others},
  booktitle={International Conference on Learning Representations (ICLR)},
  year={2023}
}

@inproceedings{kieu2025bidirectional,
  title={Bidirectional diffusion bridge models},
  author={Kieu, Duc and Do, Kien and Nguyen, Toan and Nguyen, Dang and Nguyen, Thin},
  booktitle={Proceedings of the 31st ACM SIGKDD Conference on Knowledge Discovery and Data Mining V. 2},
  pages={1139--1148},
  year={2025}
}

@article{berner2023optimal,
  title={An optimal control perspective on diffusion-based generative modeling. preprint},
  author={Berner, J and Richter, L and Ullrich, K},
  journal={arXiv},
  volume={2211},
  year={2023}
}

@article{rout2024rb,
  title={{RB-Modulation}: Training-free personalization of diffusion models using stochastic optimal control},
  author={Rout, Litu and Chen, Yujia and Ruiz, Nataniel and Kumar, Abhishek and Caramanis, Constantine and Shakkottai, Sanjay and Chu, Wen-Sheng},
  journal={arXiv preprint arXiv:2405.17401},
  year={2024}
}

@inproceedings{salimans2022progressive,
  title={Progressive Distillation for Fast Sampling of Diffusion Models},
  author={Salimans, Tim and Ho, Jonathan},
  booktitle={International Conference on Learning Representations (ICLR)},
  year={2022}
}

@article{heng2111simulating,
  title={Simulating diffusion bridges with score matching},
  author={Heng, Jeremy and De Bortoli, Valentin and Doucet, Arnaud and Thornton, James},
  journal={Biometrika},
  volume={112},
  number={4},
  year={2025},
}

@article{luhman2021knowledge,
  title={Knowledge distillation in iterative generative models for improved sampling speed},
  author={Luhman, Eric and Luhman, Troy},
  journal={arXiv preprint arXiv:2101.02388},
  year={2021}
}

@article{li2023snapfusion,
  title={{SnapFusion: Text-to-image diffusion model on mobile devices within two seconds}},
  author={Li, Yanyu and Wang, Huan and Jin, Qing and Hu, Ju and Chemerys, Pavlo and Fu, Yun and Wang, Yanzhi and Tulyakov, Sergey and Ren, Jian},
  journal={Advances in Neural Information Processing Systems (NeurIPS)},
  volume={36},
  pages={20662--20678},
  year={2023}
}

@inproceedings{meng2023distillation,
  title={On distillation of guided diffusion models},
  author={Meng, Chenlin and Rombach, Robin and Gao, Ruiqi and Kingma, Diederik and Ermon, Stefano and Ho, Jonathan and Salimans, Tim},
  booktitle={Proceedings of the IEEE/CVF conference on computer vision and pattern recognition (CVPR)},
  pages={14297--14306},
  year={2023}
}

@inproceedings{song2023consistency,
author = {Song, Yang and Dhariwal, Prafulla and Chen, Mark and Sutskever, Ilya},
title = {Consistency models},
year = {2023},
booktitle = {International Conference on Machine Learning (ICML)},
articleno = {1335},
numpages = {42},
}

@article{luo2023latent,
  title={Latent consistency models: Synthesizing high-resolution images with few-step inference},
  author={Luo, Simian and Tan, Yiqin and Huang, Longbo and Li, Jian and Zhao, Hang},
  journal={arXiv preprint arXiv:2310.04378},
  year={2023}
}

@inproceedings{kim2023consistency,
  title={Consistency Trajectory Models: Learning Probability Flow ODE Trajectory of Diffusion},
  author={Kim, Dongjun and Lai, Chieh-Hsin and Liao, Wei-Hsiang and Murata, Naoki and Takida, Yuhta and Uesaka, Toshimitsu and He, Yutong and Mitsufuji, Yuki and Ermon, Stefano},
  booktitle={International Conference on Learning Representations (ICLR)},
  year={2024}
}

@inproceedings{song2020score,
  title={Score-Based Generative Modeling through Stochastic Differential Equations},
  author={Song, Yang and Sohl-Dickstein, Jascha and Kingma, Diederik P and Kumar, Abhishek and Ermon, Stefano and Poole, Ben},
  booktitle={International Conference on Learning Representations (ICLR)},
  year={2020}
}

@inproceedings{popov2021diffusion,
  title={Diffusion-Based Voice Conversion with Fast Maximum Likelihood Sampling Scheme},
  author={Popov, Vadim and Vovk, Ivan and Gogoryan, Vladimir and Sadekova, Tasnima and Kudinov, Mikhail Sergeevich and Wei, Jiansheng},
  booktitle={International Conference on Learning Representations (ICLR)}, 
  year={2022}
}

@inproceedings{bao2022analytic,
  title={{Analytic-DPM}: an Analytic Estimate of the Optimal Reverse Variance in Diffusion Probabilistic Models},
  author={Bao, Fan and Li, Chongxuan and Zhu, Jun and Zhang, Bo},
  booktitle={International Conference on Learning Representations (ICLR)},
  year={2022}
}

@article{lu2025dpm,
  title={{DPM-Solver++: Fast solver for guided sampling of diffusion probabilistic models}},
  author={Lu, Cheng and Zhou, Yuhao and Bao, Fan and Chen, Jianfei and Li, Chongxuan and Zhu, Jun},
  journal={Machine Intelligence Research},
  pages={1--22},
  year={2025},
  publisher={Springer}
}

@article{zheng2023dpm,
  title={{DPM-Solver-v3: Improved diffusion ODE solver with empirical model statistics}},
  author={Zheng, Kaiwen and Lu, Cheng and Chen, Jianfei and Zhu, Jun},
  journal={Advances in Neural Information Processing Systems (NeurIPS)},
  volume={36},
  pages={55502--55542},
  year={2023}
}

@article{sagheer2020review,
  title={A review on medical image denoising algorithms},
  author={Sagheer, Sameera V Mohd and George, Sudhish N},
  journal={Biomedical signal processing and control},
  volume={61},
  pages={102036},
  year={2020},
  publisher={Elsevier}
}

@article{wang2022comprehensive,
  title={A comprehensive review on deep learning based remote sensing image super-resolution methods},
  author={Wang, Peijuan and Bayram, Bulent and Sertel, Elif},
  journal={Earth-Science Reviews},
  volume={232},
  pages={104110},
  year={2022},
  publisher={Elsevier}
}

@article{delbracio2021mobile,
  title={Mobile computational photography: A tour},
  author={Delbracio, Mauricio and Kelly, Damien and Brown, Michael S and Milanfar, Peyman},
  journal={Annual review of vision science},
  volume={7},
  number={1},
  pages={571--604},
  year={2021},
  publisher={Annual Reviews}
}

@article{ho2020denoising,
  title={Denoising diffusion probabilistic models},
  author={Ho, Jonathan and Jain, Ajay and Abbeel, Pieter},
  journal={Advances in neural information processing systems (NeurIPS)},
  volume={33},
  pages={6840--6851},
  year={2020}
}

@article{saharia2022image,
  title={Image super-resolution via iterative refinement},
  author={Saharia, Chitwan and Ho, Jonathan and Chan, William and Salimans, Tim and Fleet, David J and Norouzi, Mohammad},
  journal={IEEE Transactions on Pattern Analysis and Machine Intelligence},
  volume={45},
  number={4},
  pages={4713--4726},
  year={2022},
  publisher={IEEE}
}

@inproceedings{lugmayr2022repaint,
  title={Repaint: Inpainting using denoising diffusion probabilistic models},
  author={Lugmayr, Andreas and Danelljan, Martin and Romero, Andres and Yu, Fisher and Timofte, Radu and Van Gool, Luc},
  booktitle={Proceedings of the IEEE/CVF conference on computer vision and pattern recognition (CVPR)},
  pages={11461--11471},
  year={2022}
}

@inproceedings{whang2022deblurring,
  title={Deblurring via stochastic refinement},
  author={Whang, Jay and Delbracio, Mauricio and Talebi, Hossein and Saharia, Chitwan and Dimakis, Alexandros G and Milanfar, Peyman},
  booktitle={Proceedings of the IEEE/CVF conference on computer vision and pattern recognition (CVPR)},
  pages={16293--16303},
  year={2022}
}

@inproceedings{chung2022diffusion,
  title={Diffusion posterior sampling for general noisy inverse problems},
  author={Chung, Hyungjin and Kim, Jeongsol and Mccann, Michael T and Klasky, Marc L and Ye, Jong Chul},
  booktitle={International Conference on Learning Representations (ICLR)},
  year={2023}
}

@inproceedings{chung2022come,
  title={{Come-Closer-Diffuse-Faster: Accelerating conditional diffusion models for inverse problems through stochastic contraction}},
  author={Chung, Hyungjin and Sim, Byeongsu and Ye, Jong Chul},
  booktitle={Proceedings of the IEEE/CVF conference on computer vision and pattern recognition (CVPR)},
  pages={12413--12422},
  year={2022}
}

@article{shih2023parallel,
  title={Parallel sampling of diffusion models},
  author={Shih, Andy and Belkhale, Suneel and Ermon, Stefano and Sadigh, Dorsa and Anari, Nima},
  journal={Advances in Neural Information Processing Systems (NeurIPS)},
  volume={36},
  pages={4263--4276},
  year={2023}
}

@article{yue2023resshift,
  title={{ResShift: Efficient diffusion model for image super-resolution by residual shifting}},
  author={Yue, Zongsheng and Wang, Jianyi and Loy, Chen Change},
  journal={Advances in Neural Information Processing Systems (NeurIPS)},
  volume={36},
  pages={13294--13307},
  year={2023}
}

@inproceedings{song2020denoising,
  title={Denoising Diffusion Implicit Models},
  author={Song, Jiaming and Meng, Chenlin and Ermon, Stefano},
  booktitle={International Conference on Learning Representations (ICLR)},
  year={2021}
}

@article{kawar2022denoising,
  title={Denoising diffusion restoration models},
  author={Kawar, Bahjat and Elad, Michael and Ermon, Stefano and Song, Jiaming},
  journal={Advances in Neural Information Processing Systems (NeurIPS)},
  volume={35},
  pages={23593--23606},
  year={2022}
}

@article{dabov2007image,
  title={Image denoising by sparse 3-D transform-domain collaborative filtering},
  author={Dabov, Kostadin and Foi, Alessandro and Katkovnik, Vladimir and Egiazarian, Karen},
  journal={IEEE Transactions on Image Processing},
  volume={16},
  number={8},
  pages={2080--2095},
  year={2007},
  publisher={IEEE}
}

@article{zhang2017beyond,
  title={{Beyond a Gaussian denoiser: Residual learning of deep CNN for image denoising}},
  author={Zhang, Kai and Zuo, Wangmeng and Chen, Yunjin and Meng, Deyu and Zhang, Lei},
  journal={IEEE Transactions on Image Processing},
  volume={26},
  number={7},
  pages={3142--3155},
  year={2017},
  publisher={IEEE}
}

@inproceedings{wang2018esrgan,
  title={{ESRGAN: Enhanced super-resolution generative adversarial networks}},
  author={Wang, Xintao and Yu, Ke and Wu, Shixiang and Gu, Jinjin and Liu, Yihao and Dong, Chao and Qiao, Yu and Change Loy, Chen},
  booktitle={Proceedings of the European conference on computer vision (ECCV) workshops},
  year={2018}
}

@article{dhariwal2021diffusion,
  title={{Diffusion models beat GANs on image synthesis}},
  author={Dhariwal, Prafulla and Nichol, Alexander},
  journal={Advances in neural information processing systems (NeurIPS)},
  volume={34},
  pages={8780--8794},
  year={2021}
}

@article{song2019generative,
  title={Generative modeling by estimating gradients of the data distribution},
  author={Song, Yang and Ermon, Stefano},
  journal={Advances in neural information processing systems (NeurIPS)},
  volume={32},
  year={2019}
}

@article{li2022srdiff,
  title={{SRDiff: Single image super-resolution with diffusion probabilistic models}},
  author={Li, Haoying and Yang, Yifan and Chang, Meng and Chen, Shiqi and Feng, Huajun and Xu, Zhihai and Li, Qi and Chen, Yueting},
  journal={Neurocomputing},
  volume={479},
  pages={47--59},
  year={2022},
  publisher={Elsevier}
}

@inproceedings{zhang2019poisson,
  title={A poisson-gaussian denoising dataset with real fluorescence microscopy images},
  author={Zhang, Yide and Zhu, Yinhao and Nichols, Evan and Wang, Qingfei and Zhang, Siyuan and Smith, Cody and Howard, Scott},
  booktitle={Proceedings of the IEEE/CVF Conference on Computer Vision and Pattern Recognition (CVPR)},
  pages={11710--11718},
  year={2019}
}

@inproceedings{agustsson2017ntire,
  title={{NTIRE 2017 challenge on single image super-resolution: Dataset and study}},
  author={Agustsson, Eirikur and Timofte, Radu},
  booktitle={Proceedings of the IEEE conference on computer vision and pattern recognition workshops (CVPRW)},
  pages={126--135},
  year={2017}
}

@inproceedings{timofte2017ntire,
  title={{NTIRE 2017 challenge on single image super-resolution: Methods and results}},
  author={Timofte, Radu and Agustsson, Eirikur and Van Gool, Luc and Yang, Ming-Hsuan and Zhang, Lei},
  booktitle={Proceedings of the IEEE conference on computer vision and pattern recognition workshops (CVPRW)},
  pages={114--125},
  year={2017}
}

@inproceedings{abdelhamed2018high,
  title={A high-quality denoising dataset for smartphone cameras},
  author={Abdelhamed, Abdelrahman and Lin, Stephen and Brown, Michael S},
  booktitle={Proceedings of the IEEE conference on computer vision and pattern recognition (CVPR)},
  pages={1692--1700},
  year={2018}
}

@inproceedings{abdelhamed2019ntire,
  title={{NTIRE 2019 challenge on real image denoising: Methods and results}},
  author={Abdelhamed, Abdelrahman and Timofte, Radu and Brown, Michael S},
  booktitle={Proceedings of the IEEE/CVF Conference on Computer Vision and Pattern Recognition Workshops (CVPRW)},
  year={2019}
}

@inproceedings{karras2019style,
  title={A style-based generator architecture for generative adversarial networks},
  author={Karras, Tero and Laine, Samuli and Aila, Timo},
  booktitle={Proceedings of the IEEE/CVF conference on computer vision and pattern recognition (CVPR)},
  pages={4401--4410},
  year={2019}
}

@book{bishop2006pattern,
  title={Pattern recognition and machine learning},
  author={Bishop, Christopher M and Nasrabadi, Nasser M},
  volume={4},
  OPTnumber={4},
  year={2006},
  publisher={Springer}
}

@article{paszke2019pytorch,
  title={{PyTorch: An imperative style, high-performance deep learning library}},
  author={Paszke, Adam and Gross, Sam and Massa, Francisco and Lerer, Adam and Bradbury, James and Chanan, Gregory and Killeen, Trevor and Lin, Zeming and Gimelshein, Natalia and Antiga, Luca and others},
  journal={Advances in neural information processing systems (NeurIPS)},
  volume={32},
  year={2019}
}

@article{kingma2014adam,
  title={Adam: A method for stochastic optimization},
  author={Kingma, Diederik P and Ba, Jimmy},
  journal={arXiv preprint arXiv:1412.6980},
  year={2014}
}

@inproceedings{liu2024residual,
  title={Residual denoising diffusion models},
  author={Liu, Jiawei and Wang, Qiang and Fan, Huijie and Wang, Yinong and Tang, Yandong and Qu, Liangqiong},
  booktitle={Proceedings of the IEEE/CVF Conference on Computer Vision and Pattern Recognition (CVPR)},
  pages={2773--2783},
  year={2024}
}

@article{wu2024one,
  title={One-step effective diffusion network for real-world image super-resolution},
  author={Wu, Rongyuan and Sun, Lingchen and Ma, Zhiyuan and Zhang, Lei},
  journal={Advances in Neural Information Processing Systems (NeurIPS)},
  volume={37},
  pages={92529--92553},
  year={2024}
}

@inproceedings{luo2023image,
  title={Image restoration with mean-reverting stochastic differential equations},
  author={Luo, Ziwei and Gustafsson, Fredrik K and Zhao, Zheng and Sj{\"o}lund, Jens and Sch{\"o}n, Thomas B},
  booktitle={International Conference on Machine Learning (ICML)},
  pages={23045--23066},
  year={2023}
}

@inproceedings{zhou2024denoising,
  title={Denoising Diffusion Bridge Models},
  author={Zhou, Linqi and Lou, Aaron and Khanna, Samar and Ermon, Stefano},
  booktitle={International Conference on Learning Representations (ICLR)},
  year={2024}
}

@inproceedings{yue2024image,
  title={Image Restoration Through Generalized {Ornstein-Uhlenbeck} Bridge},
  author={Yue, Conghan and Peng, Zhengwei and Ma, Junlong and Du, Shiyan and Wei, Pengxu and Zhang, Dongyu},
  booktitle={International Conference on Machine Learning (ICML)},
  pages={58068--58089},
  year={2024},
}

@inproceedings{zhu2025unidb,
  title={{UniDB}: A Unified Diffusion Bridge Framework via Stochastic Optimal Control},
  author={Zhu, Kaizhen and Pan, Mokai and Ma, Yuexin and Fu, Yanwei and Yu, Jingyi and Wang, Jingya and Shi, Ye},
  booktitle={International Conference on Machine Learning (ICML)},
  year={2025}
}

@article{pan2025unidb++,
  title={{UniDB++}: Fast Sampling of Unified Diffusion Bridge},
  author={Pan, Mokai and Zhu, Kaizhen and Ma, Yuexin and Fu, Yanwei and Yu, Jingyi and Wang, Jingya and Shi, Ye},
  journal={arXiv preprint arXiv:2505.21528},
  year={2025}
}

@InProceedings{you2025consistency,
  title= {Consistency Trajectory Matching for One-Step Generative Super-Resolution},
  author= {You, Weiyi and Zhang, Mingyang and Zhang, Leheng and Zhou, Xingyu and Shi, Kexuan and Gu, Shuhang},
  booktitle={Proceedings of the IEEE/CVF International Conference on Computer Vision (ICCV)},
  month={October},
  year={2025},
  pages={12747-12756}
}

@inproceedings{li2025mars,
  title={{MaRS}: A Fast Sampler for Mean Reverting Diffusion based on {ODE} and {SDE} Solvers},
  author={Li, Ao and Fang, Wei and Zhao, Hongbo and Lu, Le and Yang, Ge and Xu, Minfeng},
  booktitle={International Conference on Learning Representations (ICLR)},
  year={2025}
}

@inproceedings{zheng2024diffusion,
  title={Diffusion Bridge Implicit Models},
  author={Zheng, Kaiwen and He, Guande and Chen, Jianfei and Bao, Fan and Zhu, Jun},
  booktitle={International Conference on Learning Representations (ICLR)},
  year={2024}
}

@inproceedings{blau2018perception,
  title={The Perception-Distortion Tradeoff},
  author={Blau, Yochai and Michaeli, Tomer},
  booktitle={IEEE Conference on Computer Vision and Pattern Recognition (CVPR)},
  pages={6228--6237},
  year={2018}
}

@article{delbracioinversion,
  title={Inversion by Direct Iteration: An Alternative to Denoising Diffusion for Image Restoration},
  author={Delbracio, Mauricio and Milanfar, Peyman},
  journal={Transactions on Machine Learning Research},
  year={2023}
}

@article{fabian2024diracdiffusion,
  title={Diracdiffusion: Denoising and incremental reconstruction with assured data-consistency},
  author={Fabian, Zalan and Tinaz, Berk and Soltanolkotabi, Mahdi},
  journal={Proceedings of machine learning research},
  volume={235},
  pages={12754},
  year={2024}
}

@inproceedings{heitz2023iterative,
  title={Iterative $\alpha$-(de) blending: A minimalist deterministic diffusion model},
  author={Heitz, Eric and Belcour, Laurent and Chambon, Thomas},
  booktitle={ACM SIGGRAPH 2023 Conference Proceedings},
  pages={1--8},
  year={2023}
}
\bibliographystyle{iclr2026_conference}

\appendix
\section{Derivations}
\label{app:derivations}
This section presents detailed mathematical derivations to support this work. All intermediate steps and calculations omitted for brevity in the main text are included here for completeness and reference.

\subsection{\texorpdfstring{Forward process cumulative transition distribution $q(\boldsymbol{x_t}|\boldsymbol{x_0}, \boldsymbol{\Delta})$}{Forward process cumulative transition distribution}}
\label{app:forward_process_cumulative_transition_distribution}
The \ac{RDIM} forward process is designed to align with a forward model that converts the data, $\boldsymbol{x_0}$, into the corresponding corrupted version, $\boldsymbol{y_0}$. To achieve this, the Gaussian transition distribution in Equation (\ref{eq:forward_process_transition_markovian}) is derived for a Markovian version of the \ac{RDIM} forward process. However, when generating a latent variable $\boldsymbol{x_t}$ starting from $\boldsymbol{x_0}$, the sequential formulation of the diffusion process can become computationally expensive, particularly as the timestep $t$ increases. To address this problem, the reparameterization trick can be leveraged, allowing the cumulative Gaussian transitions of the forward process to be expressed in closed form. As a result, $\boldsymbol{x_t}$ can be computed at an arbitrary timestep $t$ as a function of $\boldsymbol{x_0}$, the fraction of residual between $\boldsymbol{x_0}$ and $\boldsymbol{y_0}$, $\lambda_t \boldsymbol{\Delta}$ (with $\lambda_t$ determining the amount of residual to be removed between each diffusion step), and optional forward variance parameter $\gamma$:
\begin{equation}
\begin{split}
\boldsymbol{x_t} &= \boldsymbol{x_{t-1}} - \lambda_t\boldsymbol{\Delta} + \sqrt{\gamma^2\lambda_t}\boldsymbol{\epsilon_t} \\
    &= \boldsymbol{x_{t-2}} - \lambda_{t-1}\boldsymbol{\Delta} + \sqrt{\gamma^2\lambda_{t-1}}\boldsymbol{\epsilon_{t-1}} - \lambda_t\boldsymbol{\Delta} + \sqrt{\gamma^2\lambda_t}\boldsymbol{\epsilon_t} \\
    &= \cdots \\
    &= \boldsymbol{x_{0}} - \boldsymbol{\Delta}\underbrace{\left(\lambda_1 + \lambda_2 + \cdots + \lambda_t\right)}_{\bar{\lambda}_{t}} + \sqrt{\gamma^2\lambda_1}\boldsymbol{\epsilon_1} + \sqrt{\gamma^2\lambda_2}\boldsymbol{\epsilon_2} + \cdots + \sqrt{\gamma^2\lambda_t}\boldsymbol{\epsilon_t} \\
\end{split}
\end{equation}
where $\boldsymbol{\epsilon_1}, \boldsymbol{\epsilon_2}, \dots, \boldsymbol{\epsilon_t} \sim \mathcal{N}(0, \boldsymbol{I})$. Hence:
\begin{equation}
\label{eq:forward_process_markovian_cumulative_transition_derivation}
\begin{split}
   \boldsymbol{x_t} 
    &\sim \mathcal{N}\left(\boldsymbol{x_{0}} - \bar{\lambda}_t\boldsymbol{\Delta}, \gamma^2\left(\lambda_1 + \lambda_2 + \cdots + \lambda_t\right)\boldsymbol{I}\right) \\
    &\sim \mathcal{N}\left(\boldsymbol{x_{0}} - \bar{\lambda}_t\boldsymbol{\Delta}, \gamma^2\bar{\lambda}_t \boldsymbol{I}\right),
\end{split}
\end{equation}
Therefore, the cumulative Gaussian transition in the forward process can be defined as in Equation (\ref{eq:forward_process_cumulative_transition}) and, when $\gamma = 0$, it collapses into a Dirac delta function.

\paragraph{\texorpdfstring{Cumulative sum of weights $\lambda_{t}$}{Cumulative sum of weights}}
Each weight $\lambda_t$, used to control the variance and amount of residual to be removed in each diffusion step, is computed as $\lambda_t = \beta_t - \beta_{t-1}$, with $\beta_t$ representing the transition at forward step $t$ between original and corrupted data in the Markov chain. Consequently, the cumulative sum of weights $\lambda_t$ from the initial timestep $t = 1$ up to timestep $t = \tau$ is given as follows:
\begin{equation}
\label{eq:cumulative_sum_lambdas}
\begin{split}
\bar{\lambda}_{\tau} &= \sum_{t=1}^{\tau} \lambda_t= \sum_{t=1}^{\tau} \left(\beta_t - \beta_{t-1}\right) = \beta_{\tau} - \beta_0 
\end{split}
\end{equation}


\paragraph{\texorpdfstring{Distribution of the last latent variable $q(\boldsymbol{x_T}|\boldsymbol{x_0}, \boldsymbol{\Delta}) = q(\boldsymbol{x_T}|\boldsymbol{y_0})$.}{Distribution of the last latent variable.}}
Given that the \ac{RDIM} forward process is designed to align with a forward model that converts the data, $\boldsymbol{x_0}$, into the corresponding corrupted version, $\boldsymbol{y_0}$, the residual, $\boldsymbol{\Delta}$, should be fully removed from $\boldsymbol{x_{0}}$ at the end of the forward process, i.e., after exactly $T$ timesteps. This ensures that the last latent variable, $\boldsymbol{x_T}$, will coincide exactly with the corrupted data, $\boldsymbol{y_0}$, when the forward process is deterministic, and will converge to a noisy sample centered at $\boldsymbol{y_0}$ when the forward process is stochastic. Hence, considering Equation (\ref{eq:forward_process_markovian_cumulative_transition_derivation}), the last latent variable, $\boldsymbol{x_{T}}$, of the forward process can be sampled as:
\begin{equation}
\label{eq:last_latent_variable_distribution_derivation_part_1}
\begin{split}
\boldsymbol{x_T} &\sim \mathcal{N}\left(\boldsymbol{x_{0}} - \bar{\lambda}_T\boldsymbol{\Delta}, \gamma^2\bar{\lambda}_T \boldsymbol{I}\right) \\
    &\sim \mathcal{N}\left(\boldsymbol{x_{0}} - \bar{\lambda}_T\left(\boldsymbol{x_0} - \boldsymbol{y_0}\right), \gamma^2\bar{\lambda}_T \boldsymbol{I}\right) \\
    &\sim \mathcal{N}\left(\boldsymbol{x_{0}}\left(1 - \bar{\lambda}_T\right) + \bar{\lambda}_T\boldsymbol{y_{0}}, \gamma^2\bar{\lambda}_T \boldsymbol{I}\right).
\end{split}
\end{equation}
Logically, to ensure the aforementioned condition of centering the distribution $q(\boldsymbol{x_T}|\boldsymbol{x_0}, \boldsymbol{\Delta})$ on the corrupted data, $\boldsymbol{y_0}$, the cumulative sum of weights $\lambda_t$ over the $T$ timesteps must satisfy $\bar{\lambda}_T = 1$. This imposes that $\beta_0 = 0$ and $\beta_T = 1$, since $\bar{\lambda}_T = \beta_T - \beta_0$, as mentioned above. Accordingly:
\begin{equation}
\label{eq:last_latent_variable_distribution_derivation_part_2}
\begin{split}
\boldsymbol{x_T} &\sim \mathcal{N}\left(\boldsymbol{y_0}, \gamma^2\boldsymbol{I}\right).
\end{split}
\end{equation}

This formulation assures that the residual $\boldsymbol{\Delta}$ is fully removed after exactly $T$ timesteps ($\bar{\lambda}_t = 1$ only when $t=T$) and that the distribution $q(\boldsymbol{x_T}|\boldsymbol{x_0}, \boldsymbol{\Delta})$ is centered at the corrupted data, $\boldsymbol{y_0}$. As a result of this deliberate design choice, $q(\boldsymbol{x_T}|\boldsymbol{x_0}, \boldsymbol{\Delta}) = q(\boldsymbol{x_T}|\boldsymbol{y_0})$ holds exactly at $t = T$. In addition, when $\gamma = 0$, the Gaussian collapses into a Dirac delta function centered at $\boldsymbol{y_0}$, thereby the final latent variable, $\boldsymbol{x_T}$, coincides exactly with the corrupted data, i.e., $\boldsymbol{x_T} = \boldsymbol{y_0}$. 

Additionally, the $\beta$-schedule defined in Equation (\ref{eq:beta_schedule}) is designed to impose $\beta_{0} = 0$ and $\beta_{T} = 1$, thus satisfying the aforementioned requirements. In particular, the cumulative sum of weights $\lambda_t$ is $\bar{\lambda}_t = \beta_t$ when $\beta_0 = 0$ (see Equation (\ref{eq:cumulative_sum_lambdas})). Figure \ref{fig:beta_schedule} showcases the progression of the weights $\beta_t$ and $\lambda_t$ across timesteps. If $p=1.0$, the $\beta$-schedule is linear and $\lambda_t$ is constant, resulting in uniform fractions of $\Delta$ removed along the forward process. 

Accordingly, under this condition of $\beta_{0} = 0$, the cumulative forward transition distribution, $q(\boldsymbol{x_t}|\boldsymbol{x_0}, \boldsymbol{\Delta})$, expressed in Equation (\ref{eq:forward_process_markovian_cumulative_transition_derivation}) can be further simplified to:
\begin{equation}
\label{eq:forward_process_markovian_cumulative_transition_derivation_simplified}
\begin{split}
\boldsymbol{x_t} 
&\sim \mathcal{N}\left(\boldsymbol{x_{0}} - \beta_t\boldsymbol{\Delta}, \gamma^2 \beta_t \boldsymbol{I}\right).
\end{split}
\end{equation}

\subsection{\texorpdfstring{Reverse process transition distribution $q(\boldsymbol{x_{t-1}}|\boldsymbol{x_{t}}, \boldsymbol{x_0}, \boldsymbol{\Delta})$}{Reverse process transition distribution}}
\label{app:reverse_process_transition_distribution}
The reverse process involves computing the reverse transition, which is defined as the Gaussian distribution in Equation (\ref{eq:reverse_process_transition}) and is designed to preserve the marginal $q(\boldsymbol{x_t}|\boldsymbol{x_0}, \boldsymbol{\Delta})$ in Equation (\ref{eq:forward_process_cumulative_transition}). Considering that Gaussian distributions exhibit the property that their conditional means are linear combinations of the conditioning variables (see Lemma \ref{lem:conditional_gaussian_linear_mean}), then the mean $\boldsymbol{\tilde{\mu}_t}$ of $q(\boldsymbol{x_{t-1}}|\boldsymbol{x_{t}}, \boldsymbol{x_0}, \boldsymbol{\Delta})$ can be expressed as a linear interpolation between $\boldsymbol{x_{t}}$, $\boldsymbol{x_0}$, and $\boldsymbol{\Delta}$. Particularly, to match the form of the forward process cumulative transition, $q(\boldsymbol{x_t}|\boldsymbol{x_0}, \boldsymbol{\Delta})$, the mean $\boldsymbol{\tilde{\mu}_t}$ is assumed to be a linear combination between $(\boldsymbol{x_0} - \beta_t\boldsymbol{\Delta})$ and $\boldsymbol{x_t}$:
\begin{equation}
\label{eq:reverse_process_transition_mean_linear_combination}
\boldsymbol{\tilde{\mu}_t} = a\left(\boldsymbol{x_0} - \beta_t\boldsymbol{\Delta}\right) + b\boldsymbol{x_t},
\end{equation}
where $a$ and $b$ are constants.

Following, given $q(\boldsymbol{x_t}|\boldsymbol{x_0}, \boldsymbol{\Delta})$ and the formulation assumed for $q(\boldsymbol{x_{t-1}}|\boldsymbol{x_t}, \boldsymbol{x_0}, \boldsymbol{\Delta})$, then $q(\boldsymbol{x_{t-1}}|\boldsymbol{x_0}, \boldsymbol{\Delta})$ can be defined by leveraging a property of marginal and conditional Gaussians (see Lemma \ref{lem:marginal_conditional_gaussians}):
\begin{equation}
\label{eq:previous_latent_variable_forward_process_cumulative_transition}
\begin{split}
q(\boldsymbol{x_{t-1}}|\boldsymbol{x_0}, \boldsymbol{\Delta}) &= \mathcal{N}\left(\boldsymbol{x_{t-1}} \big| b\left(\boldsymbol{x_0} - \beta_t\boldsymbol{\Delta}\right) + a\left(\boldsymbol{x_0} - \beta_t\boldsymbol{\Delta}\right), \tilde{\sigma}^2_t\boldsymbol{I} + b\gamma^2\beta_t \boldsymbol{I}b\right) \\
& = \mathcal{N}\left(\boldsymbol{x_{t-1}} \big| \left(\boldsymbol{x_0} - \beta_t\boldsymbol{\Delta}\right)\left(a + b\right), \left(\tilde{\sigma}^2_t + \gamma^2\beta_t b^2\right)\boldsymbol{I}\right).
\end{split}
\end{equation}

Recalling that $q(\boldsymbol{x_t}|\boldsymbol{x_0}, \boldsymbol{\Delta}) = \mathcal{N}\left(\boldsymbol{x_t} | \boldsymbol{x_0} - \beta_t\boldsymbol{\Delta}, \gamma^2\beta_t \boldsymbol{I}\right)$ is being enforced, the cumulative Gaussian transition to obtain $\boldsymbol{x_{t-1}}$ given $\boldsymbol{x_0}$ and $\boldsymbol{\Delta}$ is also defined as: 
\begin{equation}
\label{eq:previous_latent_variable_forward_process_cumulative_transition_default}
q(\boldsymbol{x_{t-1}}|\boldsymbol{x_0}, \boldsymbol{\Delta}) = \mathcal{N}\left(\boldsymbol{x_{t-1}} | \boldsymbol{x_0} - \beta_{t-1}\boldsymbol{\Delta}, \gamma^2\beta_{t-1} \boldsymbol{I}\right). 
\end{equation}
Accordingly, to ensure that the designed reverse transition preserves the marginal $q(\boldsymbol{x_t}|\boldsymbol{x_0}, \boldsymbol{\Delta})$, \hl{as guaranteed by Lemma \mbox{\ref{lem:marginal_conditional_gaussians}}}, the following equality must be satisfied:
\begin{equation}
\label{eq:previous_latent_variable_forward_process_cumulative_transition_equality}
\mathcal{N}\left(\boldsymbol{x_{t-1}} \big| \left(\boldsymbol{x_0} - \beta_t\boldsymbol{\Delta}\right)\left(a + b\right), \left(\tilde{\sigma}^2_t + \gamma^2\beta_t b^2\right)\boldsymbol{I}\right) = \mathcal{N}\left(\boldsymbol{x_{t-1}} \big| \boldsymbol{x_0} - \beta_{t-1}\boldsymbol{\Delta}, \gamma^2\beta_{t-1} \boldsymbol{I}\right),
\end{equation}
and thus $a$ and $b$ can be computed by solving the following system of equations:
\begin{equation}
\label{eq:previous_latent_variable_forward_process_cumulative_transition_system}
\begin{split}
&\begin{cases}
    \left(\boldsymbol{x_0} - \beta_t\boldsymbol{\Delta}\right)\left(a + b\right) = \boldsymbol{x_0} - \beta_{t-1}\boldsymbol{\Delta} \\
    \tilde{\sigma}^2_t + \gamma^2\beta_t b^2 = \gamma^2\beta_{t-1}
\end{cases} \Leftrightarrow
\begin{cases}
    a = 1 + \frac{\lambda_t\boldsymbol{\Delta}}{\boldsymbol{x_0} - \beta_t\boldsymbol{\Delta}} - \sqrt{\frac{\gamma^2\beta_{t-1} - \tilde{\sigma}^2_t}{\gamma^2\beta_t}} \\
    b = \sqrt{\frac{\gamma^2\beta_{t-1} - \tilde{\sigma}^2_t}{\gamma^2\beta_t}}
\end{cases}.
\end{split}
\end{equation}

Consequently, the mean of each reverse transition, $\boldsymbol{\tilde{\mu}_t}$, is given as:
\begin{equation}
\label{eq:reverse_process_transition_mean}
\begin{split}
\boldsymbol{\tilde{\mu}_t} &= a\left(\boldsymbol{x_0} - \beta_t\boldsymbol{\Delta}\right) + b\boldsymbol{x_t} \\
&= \left(1 + \frac{\lambda_t\boldsymbol{\Delta}}{\boldsymbol{x_0} - \beta_t\boldsymbol{\Delta}} - \sqrt{\frac{\gamma^2\beta_{t-1} - \tilde{\sigma}^2_t}{\gamma^2\beta_t}}\right)\left(\boldsymbol{x_0} - \beta_t\boldsymbol{\Delta}\right) + \sqrt{\frac{\gamma^2\beta_{t-1} - \tilde{\sigma}^2_t}{\gamma^2\beta_t}}\boldsymbol{x_t} \\
&= \boldsymbol{x_0} - \beta_t\boldsymbol{\Delta} + \lambda_t\boldsymbol{\Delta} -\sqrt{\frac{\gamma^2\beta_{t-1} - \tilde{\sigma}^2_t}{\gamma^2\beta_t}}\left(\boldsymbol{x_0} - \beta_t\boldsymbol{\Delta}\right) + \sqrt{\frac{\gamma^2\beta_{t-1} - \tilde{\sigma}^2_t}{\gamma^2\beta_t}}\boldsymbol{x_t} \\
&= \boldsymbol{x_0} - \beta_t\boldsymbol{\Delta} + \left(\beta_t -\beta_{t-1}\right)\boldsymbol{\Delta} + \sqrt{\frac{\gamma^2\beta_{t-1} - \tilde{\sigma}^2_t}{\gamma^2\beta_t}}\left(\boldsymbol{x_t} - \boldsymbol{x_0} + \beta_t\boldsymbol{\Delta}\right) \\
&= \boldsymbol{x_0} - \beta_{t-1}\boldsymbol{\Delta} + \sqrt{\gamma^2\beta_{t-1} - \tilde{\sigma}^2_t}\left(\frac{\boldsymbol{x_t} - \boldsymbol{x_0} + \beta_t\boldsymbol{\Delta}}{{\sqrt{\gamma^2\beta_t}}}\right),
\end{split}
\end{equation}
where, in particular, singularities can occur for $\gamma = 0$. Therefore, for $\gamma \ne 0$, the mean of the reverse process transition distribution that preserves the marginal $q(\boldsymbol{x_t}|\boldsymbol{x_0}, \boldsymbol{\Delta})$ is given as:
\begin{equation}
\label{eq:reverse_process_transition_closed_form_mean_non_zero_gamma}
\boldsymbol{\tilde{\mu}}_{\boldsymbol{t}|\gamma\ne 0} = 
\boldsymbol{x_0} - \beta_{t-1}\boldsymbol{\Delta} + \sqrt{\gamma^2\beta_{t-1} - \tilde{\sigma}^2_t}\left(\frac{\boldsymbol{x_t} - \boldsymbol{x_0} + \beta_t\boldsymbol{\Delta}}{{\sqrt{\gamma^2\beta_t}}}\right).
\end{equation}

Essentially, the mean, $\boldsymbol{\tilde{\mu}_t}$, is chosen to ensure that $q(\boldsymbol{x_t}|\boldsymbol{x_0}, \boldsymbol{\Delta}) = \mathcal{N}\left(\boldsymbol{x_t} | \boldsymbol{x_0} - \beta_t\boldsymbol{\Delta}, \gamma^2\beta_t \boldsymbol{I}\right)$ is satisfied for all $t \in \{1, 2, \dots, T\}$. Meanwhile, the variance $\tilde{\sigma}^2_t$ is set equal to the variance of the ResShift reverse transition (see Appendix \ref{app:reverse_transition_with_resshift_variance}), thus $\tilde{\sigma}^2_t = \gamma^2\frac{\beta_{t-1}}{\beta_t}\lambda_t = \tilde{\lambda}_t$.

\paragraph{Relationship between $\boldsymbol{x_t}$, $\boldsymbol{x_0}$, $\boldsymbol{\Delta}$, and $\boldsymbol{\epsilon}$.}
Considering the marginal $q(\boldsymbol{x_t}|\boldsymbol{x_0}, \boldsymbol{\Delta})$ and $\gamma \ne 0$, a relationship between $\boldsymbol{x_t}$, $\boldsymbol{x_0}$, $\boldsymbol{\Delta}$, and $\boldsymbol{\epsilon}\sim \mathcal{N}\left(0, \boldsymbol{I}\right)$ can be derived from the reparameterization trick:
\begin{equation}
\label{eq:forward_process_cumulative_transition_reparameterization}
\begin{split}
q(\boldsymbol{x_t}|\boldsymbol{x_0}, \boldsymbol{\Delta}) &= \mathcal{N}\left(\boldsymbol{x_t} | \boldsymbol{x_0} - \beta_t\boldsymbol{\Delta}, \gamma^2\beta_t \boldsymbol{I}\right) \\
\Rightarrow \boldsymbol{x_t} &= \boldsymbol{x_0} - \beta_t\boldsymbol{\Delta} + \sqrt{\gamma^2\beta_t}\boldsymbol{\epsilon} \\
\Leftrightarrow \boldsymbol{\epsilon} &= \frac{\boldsymbol{x_t} - \boldsymbol{x_0} + \beta_t\boldsymbol{\Delta}}{\sqrt{\gamma^2\beta_t}},
\end{split}
\end{equation}
This expression exactly matches the term between parentheses in the mean of the reverse process transition distribution for $\gamma \ne 0$, 
in Equation (\ref{eq:reverse_process_transition_closed_form_mean_non_zero_gamma}). Accordingly, the mean can be rewritten as:
\begin{equation}
\label{eq:reverse_process_transition_closed_form_mean_epsilon}
\boldsymbol{\tilde{\mu}}_{\boldsymbol{t}|\gamma{\ne}0} = \boldsymbol{x_0} - \beta_{t-1}\boldsymbol{\Delta} + \sqrt{\gamma^2\beta_{t-1} - \tilde{\sigma}^2_t}\boldsymbol{\epsilon},
\end{equation}
which structurally matches the reparameterized form of the marginal $q(\boldsymbol{x_{t-1}}|\boldsymbol{x_0}, \boldsymbol{\Delta})$, exhibiting the same functional form and differing only in the variance term. This highlights that, when $\gamma \ne 0$, the reverse transition is aligned with cumulative transitions and can be leveraged to efficiently sample any state at an arbitrary timestep.

\paragraph{Reverse transition with $\gamma = 0$.}
Particularly, for $\gamma = 0$, the forward process cumulative transition, defined as a Gaussian distribution, degenerates into a Dirac delta function (see also Appendix \ref{app:forward_process_cumulative_transition_distribution}). Consequently, for $\gamma = 0$, Lemma \ref{lem:marginal_conditional_gaussians} is not applicable. 
In fact, in this case, the forward process  
effectively becomes a linear interpolation between $\boldsymbol{x_0}$ and $\boldsymbol{y_0}$. 
Logically, when $\gamma = 0$, it follows that the reverse process simply needs to invert this deterministic process. However, the continuity of the mean, $\boldsymbol{\tilde{\mu}_t}$, should be assured at $\gamma = 0$, i.e., $\boldsymbol{\tilde{\mu}}_{\boldsymbol{t}|\gamma{=}0} = \lim_{\gamma \to 0} \boldsymbol{\tilde{\mu}}_{\boldsymbol{t}|\gamma{\ne}0}$.

Considering Equation (\ref{eq:forward_process_markovian_cumulative_transition_derivation_simplified}) in Appendix \ref{app:forward_process_cumulative_transition_distribution}, it follows $\lim_{\gamma \to 0} \boldsymbol{x_t} = \boldsymbol{x_0} - \beta_t\boldsymbol{\Delta}$, which implies that $\boldsymbol{x_t} - \boldsymbol{x_0} + \beta_t\boldsymbol{\Delta} \to \boldsymbol{0}$ as $\gamma \to 0$. Accordingly, given $\tilde{\sigma}^2_t = \gamma^2\frac{\beta_{t-1}}{\beta_t}\lambda_t$, then $\lim_{\gamma \to 0} \boldsymbol{\tilde{\mu}}_{\boldsymbol{t}|\gamma{\ne}0} = \boldsymbol{x_0} - \beta_{t-1}\boldsymbol{\Delta}$. As a result, to ensure the continuity of the mean $\boldsymbol{\tilde{\mu}_t}$ at $\gamma = 0$, the Gaussian transition $q(\boldsymbol{x_{t-1}}|\boldsymbol{x_{t}}, \boldsymbol{x_0}, \boldsymbol{\Delta})$ is assumed to collapse into a Dirac delta function centered at $\boldsymbol{x_0} - \beta_{t-1}\boldsymbol{\Delta}$. Hence, for $\gamma = 0$, the mean of the reverse process transition distribution is defined as:
\begin{equation}
\label{eq:reverse_process_transition_closed_form_mean_zero_gamma}
\boldsymbol{\tilde{\mu}}_{\boldsymbol{t}|\gamma{=}0} = \boldsymbol{x_0} - \beta_{t-1}\boldsymbol{\Delta}.
\end{equation}

Notably, this formulation of $\boldsymbol{\tilde{\mu}}_{\boldsymbol{t}|\gamma{=}0}$ matches the mean of the cumulative forward transition, $q(\boldsymbol{x_{t-1}}|\boldsymbol{x_0}, \boldsymbol{\Delta})$ (see Appendix \ref{app:forward_process_cumulative_transition_distribution}), showing that the reverse process, when $\gamma = 0$, reduces to a linear interpolation between $\boldsymbol{y_0}$ and $\boldsymbol{x_0}$ (inverse of the deterministic forward process). Additionally, it aligns with the concept of cumulative transitions, which is paramount for long-range transitions (see Section \ref{sec:long_range_reverse_transition}). In essence, the mean, $\boldsymbol{\tilde{\mu}_t}$, is expressed as in Equation (\ref{eq:reverse_process_transition_closed_form_mean}) and is continuous at $\gamma = 0$. Nonetheless, the $\gamma$ constant hyperparameter is immutable in practice, i.e., set only once for each model instance, thereby no discontinuity issues would ever arise due to $\gamma$ (see Appendix \ref{app:training objective}).

\subsection{\texorpdfstring{Reverse transition with $\tilde{\sigma}^2_t = \gamma^2\frac{\beta_{t-1}}{\beta_t}\lambda_t$ (ResShift variance, $\tilde{\lambda}_t$)}{Reverse transition with ResShift variance}}
\label{app:reverse_transition_with_resshift_variance}
In particular, if the reverse process transition variance, $\tilde{\sigma}^2_t$, is set to be the same as in ResShift, $\tilde{\lambda}_t = \gamma^2\frac{\beta_{t-1}}{\beta_t}\lambda_t$, then the mean, $\boldsymbol{\tilde{\mu}}_{\boldsymbol{t}|\gamma{\ne}0}$, reduces to:
\begingroup
\allowdisplaybreaks
\stepcounter{equation}
\begin{align*}
\label{eq:reverse_process_transition_mean_non_zero_gamma_with_resshift_variance}
\boldsymbol{\tilde{\mu}}_{\boldsymbol{t}|\gamma{\ne}0} &= \boldsymbol{x_0} - \beta_{t-1}\boldsymbol{\Delta} + \sqrt{\gamma^2\beta_{t-1} - \tilde{\sigma}^2_t}\left(\frac{\boldsymbol{x_t} - \boldsymbol{x_0} + \beta_t\boldsymbol{\Delta}}{{\sqrt{\gamma^2\beta_t}}}\right) \\
&= \boldsymbol{x_0} - \beta_{t-1}\boldsymbol{\Delta} + \sqrt{\gamma^2\beta_{t-1} - \gamma^2\frac{\beta_{t-1}}{\beta_t}\lambda_t}\left(\frac{\boldsymbol{x_t} - \boldsymbol{x_0} + \beta_t\boldsymbol{\Delta}}{{\sqrt{\gamma^2\beta_t}}}\right) \\
&= \boldsymbol{x_0} - \beta_{t-1}\boldsymbol{\Delta} + \sqrt{\frac{\gamma^4\beta_t\beta_{t-1} - \gamma^4\beta_{t-1}\left(\beta_t - \beta_{t-1}\right)}{\gamma^2\beta_t}}\left(\frac{\boldsymbol{x_t} - \boldsymbol{x_0} + \beta_t\boldsymbol{\Delta}}{{\sqrt{\gamma^2\beta_t}}}\right) \\
&= \boldsymbol{x_0} - \beta_{t-1}\boldsymbol{\Delta} + \frac{\sqrt{\gamma^4\beta_{t-1}^2}\left(\boldsymbol{x_t} - \boldsymbol{x_0} + \beta_t\boldsymbol{\Delta}\right)}{\gamma^2\beta_t} \tag{\theequation} \\
&= \boldsymbol{x_0} - \beta_{t-1}\boldsymbol{\Delta} + \frac{\beta_{t-1}\boldsymbol{x_t} - \beta_{t-1}\boldsymbol{x_0} + \beta_t\beta_{t-1}\boldsymbol{\Delta}}{\beta_t} \\
&= \frac{\beta_{t-1}}{\beta_t}\boldsymbol{x_t} + \boldsymbol{x_0}\left(1 - \frac{\beta_{t-1}}{\beta_t}\right) \\
&= \frac{\beta_{t-1}}{\beta_t}\boldsymbol{x_t} + \frac{\lambda_t}{\beta_t}\boldsymbol{x_0},
\end{align*}
\endgroup
and thus the distribution $q(\boldsymbol{x_{t-1}}|\boldsymbol{x_t}, \boldsymbol{x_0}, \boldsymbol{\Delta})_{\gamma \ne 0}$ becomes:
\begin{equation}
\label{eq:reverse_process_transition_non_zero_gamma_with_resshift_variance}
q(\boldsymbol{x_{t-1}}|\boldsymbol{x_t}, \boldsymbol{x_0}, \boldsymbol{\Delta})_{\gamma \ne 0} = \mathcal{N}\left(\boldsymbol{x_{t-1}}\bigg|\frac{\beta_{t-1}}{\beta_t}\boldsymbol{x_t} + \frac{\lambda_t}{\beta_t}\boldsymbol{x_0}, \tilde{\lambda}_t\boldsymbol{I}\right),
\end{equation}
which is exactly the ResShift reverse transition distribution. In essence, if the \ac{RDIM} reverse transition variance, $\tilde{\sigma}^2_t$, is set to be the same as in ResShift, then $\boldsymbol{\tilde{\mu}_t}$ will match the mean of the ResShift reverse transition. Accordingly, \ac{RDIM} reduces to ResShift for this specific variance, revealing that ResShift is a particular case of \ac{RDIM}.

Alternatively, for $\gamma \neq 0$, if the variance is set to $\tilde{\sigma}^2_t = 0$, then there are no stochastic terms involved when traversing the reverse trajectory, as $q(\boldsymbol{x_{t-1}}|\boldsymbol{x_t}, \boldsymbol{x_0}, \boldsymbol{\Delta})_{\gamma \ne 0}$ degenerates into a $\delta$-distribution and avoids sampling random noise (given Equations (\ref{eq:reverse_process_transition}) and (\ref{eq:reverse_process_transition_closed_form_mean_non_zero_gamma})). Consequently, the reverse process becomes deterministic. Therefore, a constant hyperparameter, $\eta \in [0, 1]$, can be introduced to interpolate between a deterministic and stochastic reverse process when $\gamma \neq 0$, thus allowing control over the variability in the reverse trajectory (see Equation (\ref{eq:reverse_process_transition_closed_form_mean_non_zero_gamma_eta})). Specifically, when $\eta = 0$, the Gaussian collapses into a Dirac delta function.

\paragraph{Absence of non-real square roots.} From Equation (\ref{eq:reverse_process_transition_closed_form_mean_non_zero_gamma_eta}), it follows that to avoid a non-real square root, when $\gamma \ne 0$, the condition $\gamma^2\beta_{t-1} \ge \eta^2\tilde{\lambda}_t$ must be satisfied. Considering $\tilde{\lambda}_t = \gamma^2\frac{\beta_{t-1}}{\beta_t}\lambda_t$, then:
\begin{equation}
\label{eq:gamma_eta_inequality}
\begin{split}
&\gamma^2\beta_{t-1} \ge \eta^2\tilde{\lambda}_t \Leftrightarrow 
1 \ge \eta^2\frac{\lambda_t}{\beta_t} 
\Leftrightarrow 1 \ge \eta^2\left(1 - \frac{\beta_{t-1}}{\beta_{t}}\right). 
\end{split}
\end{equation}
Recalling that $\eta \in [0, 1]$, then $0 \le \eta^2 \le 1$. Moreover, since $0 \le \beta_{t-1} \le \beta_t$, it follows that $0 \le \frac{\beta_{t-1}}{\beta_{t}} \le 1$, which in turn ensures that the term inside parentheses meets the condition $1 - \frac{\beta_{t-1}}{\beta_{t}} \le 1$. Consequently, the product of these two terms is always less than or equal to $1$, and thus the Inequality (\ref{eq:gamma_eta_inequality}) is satisfied for all $\eta \in [0, 1]$ and $t \in \{1, 2, \dots, T\}$.

\subsection{Training objective}
\label{app:training objective}
During inference, $\boldsymbol{x_0}$ and $\boldsymbol{\Delta}$ are unknown, thus sampling from the true reverse transition distribution, $q(\boldsymbol{x_{t-1}}|\boldsymbol{x_{t}}, \boldsymbol{x_0}, \boldsymbol{\Delta})$, is not possible. Therefore, a learnable parametric model, $p_{\theta}(\boldsymbol{x_{t-1}}|\boldsymbol{x_t}, \boldsymbol{y_0})$, defined as a Gaussian distribution, is introduced to approximate $q(\boldsymbol{x_{t-1}}|\boldsymbol{x_{t}}, \boldsymbol{x_0}, \boldsymbol{\Delta})$. Particularly, an accurate estimation is required to ensure precise reconstruction of the data, $\boldsymbol{x_{0}}$, at inference. This approximation is achieved by minimizing the KL divergence between both distributions, while accounting for all timesteps:
\begin{equation}
\label{eq:training_objective}
\theta^{*} = \argmin_{\theta} D_{\mathrm{KL}}(q(\boldsymbol{x_{1:T}}|\boldsymbol{x_0}, \boldsymbol{\Delta}) \| p_{\theta}(\boldsymbol{x_{1:T}}|\boldsymbol{y_0})),
\end{equation}
where $\theta^{*}$ denotes the optimal parameters. In fact, this objective of minimizing the KL divergence in Equation (\ref{eq:training_objective}) is equivalent to minimizing the negative variational lower bound (VLB) on the conditional log-likelihood. This is the \ac{RDIM} objective function and it can expanded further:
\begingroup
\allowdisplaybreaks
\stepcounter{equation}
\begin{align*}
\label{eq:training_objective_expanded}
\mathcal{L}(\theta) &= \mathbb{E}_{q(\boldsymbol{x_{1:T}}|\boldsymbol{x_0}, \boldsymbol{\Delta})}\left[\log\left(\frac{q(\boldsymbol{x_{1:T}}|\boldsymbol{x_0}, \boldsymbol{\Delta})}{p_{\theta}(\boldsymbol{x_{0:T}}|\boldsymbol{y_0})}\right)\right] \\
&= \mathbb{E}_{q(\boldsymbol{x_{1:T}}|\boldsymbol{x_0}, \boldsymbol{\Delta})}\left[\log\left(\frac{q(\boldsymbol{x_{T}}|\boldsymbol{x_0}, \boldsymbol{\Delta})\prod_{t=2}^{T} q(\boldsymbol{x_{t-1}}|\boldsymbol{x_{t}}, \boldsymbol{x_0}, \boldsymbol{\Delta})}{p(\boldsymbol{x_{T}} | \boldsymbol{y_0})\prod_{t=1}^{T} p_{\theta}(\boldsymbol{x_{t-1}}|\boldsymbol{x_t}, \boldsymbol{y_0})}\right)\right] \\
&= \mathbb{E}_{q(\boldsymbol{x_{1:T}}|\boldsymbol{x_0}, \boldsymbol{\Delta})}\Bigg[\log\left(\frac{q(\boldsymbol{x_{T}}|\boldsymbol{x_0}, \boldsymbol{\Delta})}{p(\boldsymbol{x_{T}} | \boldsymbol{y_0})}\right) \\
&\phantom{\ =\ } + \log\left(\prod_{t=2}^{T} \frac{q(\boldsymbol{x_{t-1}}|\boldsymbol{x_{t}}, \boldsymbol{x_0}, \boldsymbol{\Delta})}{p_{\theta}(\boldsymbol{x_{t-1}}|\boldsymbol{x_t}, \boldsymbol{y_0})}\right) - \log\left(p_{\theta}(\boldsymbol{x_{0}}|\boldsymbol{x_{1}}, \boldsymbol{y_0})\right)\Bigg] \\
&= \mathbb{E}_{q(\boldsymbol{x_{T}}|\boldsymbol{x_0}, \boldsymbol{\Delta})}\left[\log\left(\frac{q(\boldsymbol{x_{T}}|\boldsymbol{x_0}, \boldsymbol{\Delta})}{p(\boldsymbol{x_{T}} | \boldsymbol{y_0})}\right)\right] \\
&\phantom{\ =\ } + \mathbb{E}_{q(\boldsymbol{x_{1:T}}|\boldsymbol{x_0}, \boldsymbol{\Delta})}\left[\sum_{t=2}^{T} \log\left(\frac{q(\boldsymbol{x_{t-1}}|\boldsymbol{x_{t}}, \boldsymbol{x_0}, \boldsymbol{\Delta})}{p_{\theta}(\boldsymbol{x_{t-1}}|\boldsymbol{x_t}, \boldsymbol{y_0})}\right)\right] \\
&\phantom{\ =\ } - \mathbb{E}_{q(\boldsymbol{x_{1}}|\boldsymbol{x_0}, \boldsymbol{\Delta})}\left[\log\left(p_{\theta}(\boldsymbol{x_{0}}|\boldsymbol{x_{1}}, \boldsymbol{y_0})\right)\right] \tag{\theequation} \\
&= D_{\mathrm{KL}}(q(\boldsymbol{x_{T}}|\boldsymbol{x_0}, \boldsymbol{\Delta}) \| p(\boldsymbol{x_{T}} | \boldsymbol{y_0})) \\
&\phantom{\ =\ } + \sum_{t=2}^{T} \mathbb{E}_{q(\boldsymbol{x_{t-1}}, \boldsymbol{x_{t}}|\boldsymbol{x_0}, \boldsymbol{\Delta})}\left[\log\left(\frac{q(\boldsymbol{x_{t-1}}|\boldsymbol{x_{t}}, \boldsymbol{x_0}, \boldsymbol{\Delta})}{p_{\theta}(\boldsymbol{x_{t-1}}|\boldsymbol{x_t}, \boldsymbol{y_0})}\right)\right] \\
&\phantom{\ =\ } - \mathbb{E}_{q(\boldsymbol{x_{1}}|\boldsymbol{x_0}, \boldsymbol{\Delta})}\left[\log\left(p_{\theta}(\boldsymbol{x_{0}}|\boldsymbol{x_{1}}, \boldsymbol{y_0})\right)\right] \\
&= D_{\mathrm{KL}}(q(\boldsymbol{x_{T}}|\boldsymbol{x_0}, \boldsymbol{\Delta}) \| p(\boldsymbol{x_{T}} | \boldsymbol{y_0})) \\
&\phantom{\ =\ } + \sum_{t=2}^{T} \mathbb{E}_{q(\boldsymbol{x_{t}}|\boldsymbol{x_0}, \boldsymbol{\Delta})}\left[\mathbb{E}_{q(\boldsymbol{x_{t-1}}|\boldsymbol{x_{t}}, \boldsymbol{x_0}, \boldsymbol{\Delta})}\left[\log\left(\frac{q(\boldsymbol{x_{t-1}}|\boldsymbol{x_{t}}, \boldsymbol{x_0}, \boldsymbol{\Delta})}{p_{\theta}(\boldsymbol{x_{t-1}}|\boldsymbol{x_t}, \boldsymbol{y_0})}\right)\right]\right] \\
&\phantom{\ =\ } - \mathbb{E}_{q(\boldsymbol{x_{1}}|\boldsymbol{x_0}, \boldsymbol{\Delta})}\left[\log\left(p_{\theta}(\boldsymbol{x_{0}}|\boldsymbol{x_{1}}, \boldsymbol{y_0})\right)\right] \\
&=  \underbrace{D_{\mathrm{KL}}(q(\boldsymbol{x_{T}}|\boldsymbol{x_0}, \boldsymbol{\Delta}) \| p(\boldsymbol{x_{T}} | \boldsymbol{y_0}))}_{\mathcal{L}_{T}} \\
&\phantom{\ =\ } + \sum_{t=2}^{T} \underbrace{\mathbb{E}_{q(\boldsymbol{x_{t}}|\boldsymbol{x_0}, \boldsymbol{\Delta})}\left[D_{\mathrm{KL}}(q(\boldsymbol{x_{t-1}}|\boldsymbol{x_{t}}, \boldsymbol{x_0}, \boldsymbol{\Delta}) \| p_{\theta}(\boldsymbol{x_{t-1}}|\boldsymbol{x_t}, \boldsymbol{y_0}))\right]}_{\mathcal{L}_{t-1}} \\
&\phantom{\ =\ } \underbrace{- \mathbb{E}_{q(\boldsymbol{x_{1}}|\boldsymbol{x_0}, \boldsymbol{\Delta})}\left[\log\left(p_{\theta}(\boldsymbol{x_{0}}|\boldsymbol{x_{1}}, \boldsymbol{y_0})\right)\right]}_{\mathcal{L}_{0}} = \mathcal{L}_{T} + \mathcal{L}_{1:T-1} + \mathcal{L}_{0}.
\end{align*}
\endgroup

Hence, analogous to \acp{DDPM}, the \ac{RDIM} objective function, $\mathcal{L}(\theta)$, decomposes into $\mathcal{L}_{T}$ (prior matching term), $\mathcal{L}_{1:T-1}$ (consistency terms), and $\mathcal{L}_{0}$ (reconstruction term).

\paragraph{Prior matching term $\mathcal{L}_{T}$.}
The term $\mathcal{L}_{T}$ is minimized when the prior, $p(\boldsymbol{x_{T}} | \boldsymbol{y_0})$, matches the true distribution of the last latent variable, $q(\boldsymbol{x_T}|\boldsymbol{x_0}, \boldsymbol{\Delta}) = q(\boldsymbol{x_T}|\boldsymbol{y_0})=\mathcal{N}(\boldsymbol{y_0}, \gamma^2I)$. Accordingly, $p(\boldsymbol{x_{T}} | \boldsymbol{y_0})$ is fixed to such a Gaussian distribution, which is parameterized by constants and involves no learnable parameters. Therefore, $\mathcal{L}_{T}$ is constant with respect to the model parameters, $\theta$, and is minimized, i.e., $\mathcal{L}_{T} = 0$. Consequently, this term can be excluded from the optimization objective, unlike the terms $\mathcal{L}_{0:T-1}$, which explicitly depend on $\theta$ through the parameterized distribution $p_\theta$.


\paragraph{Consistency terms $\mathcal{L}_{1:T-1}$.}
The terms $\mathcal{L}_{1:T-1}$ enforce that the learnable parametric model, $p_{\theta}(\boldsymbol{x_{t-1}}|\boldsymbol{x_t}, \boldsymbol{y_0})$, accurately approximates the true reverse transition, $q(\boldsymbol{x_{t-1}}|\boldsymbol{x_t}, \boldsymbol{x_0}, \boldsymbol{\Delta})$. This fundamentally ensures that the model learns to refine the data at intermediate timesteps, leading to consistency in the reconstruction.

The true reverse transition distribution is known in closed form (see Section \ref{sec:reverse_process} along with Appendices \ref{app:reverse_process_transition_distribution} and \ref{app:reverse_transition_with_resshift_variance}), having mean and variance parameterized as:
\begin{equation}
\label{eq:reverse_process_transition_closed_form_mean_eta}
\boldsymbol{\tilde{\mu}_t} =
\begin{cases}
    \boldsymbol{x_0} - \beta_{t-1}\boldsymbol{\Delta}, & \text{if } \gamma = 0, \\
    \boldsymbol{x_0} - \beta_{t-1}\boldsymbol{\Delta} + \sqrt{\gamma^2\beta_{t-1} - \eta^2\tilde{\lambda}_t}\left(\frac{\boldsymbol{x_t} - \boldsymbol{x_0} + \beta_t\boldsymbol{\Delta}}{{\sqrt{\gamma^2\beta_t}}}\right), & \text{if } \gamma \ne 0,
\end{cases}
\end{equation}
and
\begin{equation}
\label{eq:reverse_process_transition_closed_form_std_eta}
\tilde{\sigma}^2_t =
\begin{cases}
    \tilde{\lambda}_t, & \text{if } \gamma = 0, \\[6pt]
    \eta^2\tilde{\lambda}_t, & \text{if } \gamma \ne 0,
\end{cases}
\end{equation}
where $\tilde{\lambda}_t = \gamma^2\frac{\beta_{t-1}}{\beta_t}\lambda_t$.

Given that $p_{\theta}(\boldsymbol{x_{t-1}}|\boldsymbol{x_t}, \boldsymbol{y_0})$ is defined as a Gaussian distribution with mean $\boldsymbol{\mu_{\theta}}\left(\boldsymbol{x_t}, \boldsymbol{y_0}, t\right)$ and variance $\sigma^2_{\theta}\left(\boldsymbol{x_t}, \boldsymbol{y_0}, t\right)$, to minimize the KL divergence of each term $\mathcal{L}_{1:T-1}$, the mean and variance of the parametric model should approximate $\boldsymbol{\tilde{\mu}_t}$ and $\tilde{\sigma}^2_t$, respectively. Particularly, the variance of $q(\boldsymbol{x_{t-1}}|\boldsymbol{x_t}, \boldsymbol{x_0}, \boldsymbol{\Delta})$ does not have learnable parameters because it is defined in terms of constant hyperparameters, which are known. Therefore, $\sigma^2_{\theta}\left(\boldsymbol{x_t}, \boldsymbol{y_0}, t\right)$ can be fixed to equal exactly $\tilde{\sigma}^2_t$, as expressed in Equation (\ref{eq:reverse_process_transition_approximation_variance}). Following, each term $\mathcal{L}_{1:T-1}$ is computed by applying the closed-form expression for the KL divergence between two $d$-dimensional multivariate Gaussian distributions, yielding:
\begingroup
\allowdisplaybreaks
\stepcounter{equation}
\begin{align*}
\label{eq:training_objective_kl_divergence_multivariate_gaussians}
& D_{\mathrm{KL}}(q(\boldsymbol{x_{t-1}}|\boldsymbol{x_{t}}, \boldsymbol{x_0}, \boldsymbol{\Delta}) \| p_{\theta}(\boldsymbol{x_{t-1}}|\boldsymbol{x_t}, \boldsymbol{y_0})) \\
&= \frac{1}{2}\bigg(\log\left(\frac{|\sigma^2_{\theta}\left(\boldsymbol{x_t}, \boldsymbol{y_0}, t\right)\boldsymbol{I}|}{|\tilde{\sigma}^2_t\boldsymbol{I}|}\right) - d + \text{tr}\left(\left(\sigma^2_{\theta}\left(\boldsymbol{x_t}, \boldsymbol{y_0}, t\right)\boldsymbol{I}\right)^{-1}\tilde{\sigma}^2_t\boldsymbol{I}\right) \\
&\phantom{\ =\ } + \left(\boldsymbol{\mu_{\theta}}\left(\boldsymbol{x_t}, \boldsymbol{y_0}, t\right) - \boldsymbol{\tilde{\mu}_t}\right)^{\top}\left(\sigma^2_{\theta}\left(\boldsymbol{x_t}, \boldsymbol{y_0}, t\right)\boldsymbol{I}\right)^{-1}\left(\boldsymbol{\mu_{\theta}}\left(\boldsymbol{x_t}, \boldsymbol{y_0}, t\right) - \boldsymbol{\tilde{\mu}_t}\right)\bigg) \\
&= \frac{1}{2}\bigg(\log\left(\frac{|\tilde{\sigma}^2_t\boldsymbol{I}|}{|\tilde{\sigma}^2_t\boldsymbol{I}|}\right) - d + \text{tr}\left(\left(\tilde{\sigma}^2_t\boldsymbol{I}\right)^{-1}\tilde{\sigma}^2_t\boldsymbol{I}\right) \\
&\phantom{\ =\ } + \left(\boldsymbol{\mu_{\theta}}\left(\boldsymbol{x_t}, \boldsymbol{y_0}, t\right) - \boldsymbol{\tilde{\mu}_t}\right)^{\top}\left(\tilde{\sigma}^2_t\boldsymbol{I}\right)^{-1}\left(\boldsymbol{\mu_{\theta}}\left(\boldsymbol{x_t}, \boldsymbol{y_0}, t\right) - \boldsymbol{\tilde{\mu}_t}\right)\bigg) \tag{\theequation} \\
&= \frac{1}{2}\bigg(
\left(\frac{1}{\tilde{\sigma}^2_t}\right)\left(\boldsymbol{\mu_{\theta}}\left(\boldsymbol{x_t}, \boldsymbol{y_0}, t\right) - \boldsymbol{\tilde{\mu}_t}\right)^{\top}\left(\boldsymbol{\mu_{\theta}}\left(\boldsymbol{x_t}, \boldsymbol{y_0}, t\right) - \boldsymbol{\tilde{\mu}_t}\right)\bigg) \\
&= \frac{1}{2\tilde{\sigma}^2_t}\|\boldsymbol{\mu_{\theta}}\left(\boldsymbol{x_t}, \boldsymbol{y_0}, t\right) - \boldsymbol{\tilde{\mu}_t}\|^2,
\end{align*}
where 
$|\cdot|$ denotes the determinant of a matrix, and $\text{tr}(\cdot)$ is the trace of a matrix. Notably, minimizing the KL divergence effectively reduces to decreasing the difference between the means $\boldsymbol{\mu_{\theta}}\left(\boldsymbol{x_t}, \boldsymbol{y_0}, t\right)$ and $\boldsymbol{\tilde{\mu}_t}$.

However, this is only valid with $\gamma \ne 0$ and $\eta \ne 0$. In contrast, when either $\gamma = 0$ or $\eta = 0$, the true reverse transition Gaussian collapses into a Dirac delta function:
\begin{equation}
\label{eq:reverse_process_transition_closed_form_distributions_eta}
q(\boldsymbol{x_{t-1}}|\boldsymbol{x_t}, \boldsymbol{x_0}, \boldsymbol{\Delta}) =
\left\{
\begin{array}{@{} l l r @{}}
    \delta\left(\boldsymbol{x_{t-1}} - \boldsymbol{\tilde{\mu}}_{\boldsymbol{t}|\gamma{=}0}\right), & \text{if } \gamma = 0, & {\color{NavyBlue} \text{ (A)}} \\
    \delta\left(\boldsymbol{x_{t-1}} - \boldsymbol{\tilde{\mu}}_{\boldsymbol{t}|\gamma{\ne}0}\right), & \text{if } \gamma \ne 0 \text{ and } \eta = 0, & {\color{BurntOrange} \text{ (B)}} \\
    \mathcal{N}(\boldsymbol{x_{t-1}}|\boldsymbol{\tilde{\mu}}_{\boldsymbol{t}|\gamma{\ne}0}, {}\tilde{\sigma}^2_{\boldsymbol{t}|\gamma{\ne}0}\boldsymbol{I}), & \text{if } \gamma \ne 0 \text{ and } \eta \ne 0, & {\color{ForestGreen} \text{ (C)}}
\end{array}
\right.
\end{equation}
where $\color{NavyBlue} \text{A}$, $\color{BurntOrange} \text{B}$, and $\color{ForestGreen} \text{C}$ correspond to the cases of $\gamma = 0$, $(\gamma \ne 0$ and $\eta = 0)$, and $(\gamma \ne 0$ and $\eta \ne 0)$, respectively.

The KL divergence between two Dirac delta functions is not defined in the conventional sense due to their singular nature, but it can be analyzed through limiting behavior. Two delta functions centered at different points have infinite divergence, thereby the KL divergence of each term $\mathcal{L}_{1:T-1}$ tends to infinity, when $\gamma = 0$ or $\eta = 0$, unless $\boldsymbol{\mu_{\theta}}\left(\boldsymbol{x_t}, \boldsymbol{y_0}, t\right) = \boldsymbol{\tilde{\mu}_t}$:
\begin{equation}
\label{eq:training_objective_mean_difference}
\begin{split}
& D_{\mathrm{KL}}(q(\boldsymbol{x_{t-1}}|\boldsymbol{x_{t}}, \boldsymbol{x_0}, \boldsymbol{\Delta}) \| p_{\theta}(\boldsymbol{x_{t-1}}|\boldsymbol{x_t}, \boldsymbol{y_0})) \\
&= \begin{cases}
    0, & \begin{split}
            &\text{if } ({\color{NavyBlue} \text{A}} \text{ and } 
            \boldsymbol{\mu}_{\boldsymbol{\theta}|\gamma{=}0}\left(\boldsymbol{x_t}, \boldsymbol{y_0}, t\right) = \boldsymbol{\tilde{\mu}}_{\boldsymbol{t}|\gamma{=}0}) \\ &\text{or } ({\color{BurntOrange} \text{B}} \text{ and } \boldsymbol{\mu}_{\boldsymbol{\theta}|\gamma{\ne}0}\left(\boldsymbol{x_t}, \boldsymbol{y_0}, t\right) = \boldsymbol{\tilde{\mu}}_{\boldsymbol{t}|\gamma{\ne}0}),
        \end{split} \\[1.5em]
    \infty, & \begin{split}
                &\text{if } ({\color{NavyBlue} \text{A}} \text{ and } \boldsymbol{\mu}_{\boldsymbol{\theta}|\gamma{=}0}\left(\boldsymbol{x_t}, \boldsymbol{y_0}, t\right) \ne \boldsymbol{\tilde{\mu}}_{\boldsymbol{t}|\gamma{=}0}) \\ &\text{or } ({\color{BurntOrange} \text{B}} \text{ and } \boldsymbol{\mu}_{\boldsymbol{\theta}|\gamma{\ne}0}\left(\boldsymbol{x_t}, \boldsymbol{y_0}, t\right) \ne \boldsymbol{\tilde{\mu}}_{\boldsymbol{t}|\gamma{\ne}0}),
            \end{split} \\[1.5em]
    \frac{1}{2{{}\tilde{\sigma}^2_{t|\gamma{\ne}0}}}\|\boldsymbol{\mu}_{\boldsymbol{\theta}|\gamma{\ne}0}\left(\boldsymbol{x_t}, \boldsymbol{y_0}, t\right) - \boldsymbol{\tilde{\mu}}_{\boldsymbol{t}|\gamma{\ne}0}\|^2, & \text{if } {\color{ForestGreen} \text{C}}.
\end{cases}
\end{split}
\end{equation}

For the Dirac delta cases, where either $\gamma = 0$ or $\eta = 0$, to avoid an infinite loss, the only choice is to force $\boldsymbol{\mu_{\theta}}\left(\boldsymbol{x_t}, \boldsymbol{y_0}, t\right) = \boldsymbol{\tilde{\mu}_t}$. However, directly optimizing under such a hard constraint is infeasible in practice, as it provides no gradient information unless the condition is already satisfied. To circumvent this, a relaxed proxy objective is adopted, mirroring the approach used in the Gaussian case. Specifically, it minimizes half of the squared Euclidean distance between $\boldsymbol{\mu_{\theta}}\left(\boldsymbol{x_t}, \boldsymbol{y_0}, t\right)$ and $\boldsymbol{\tilde{\mu}_t}$. This mean-matching proxy loss serves as a differentiable surrogate that naturally encourages the model to align the means and can be interpreted as the limiting case of the KL divergence when the variance tends to zero. Consequently, the reduction of the KL divergence to mean matching holds for all scenarios of $\gamma$ and $\eta$.

Moreover, considering the formulation of $\boldsymbol{\tilde{\mu}_t}$ given in Equation (\ref{eq:reverse_process_transition_closed_form_mean_eta}) and since at every timestep, $t$, in the reverse process, only the exact values of $\boldsymbol{x_0}$ and $\boldsymbol{\Delta}$ are unknown, then $\boldsymbol{\mu_{\theta}}\left(\boldsymbol{x_t}, \boldsymbol{y_0}, t\right)$ can be defined as in Equation (\ref{eq:reverse_process_transition_approximation_mean}). In this definition of $\boldsymbol{\mu_{\theta}}\left(\boldsymbol{x_t}, \boldsymbol{y_0}, t\right)$, the only components dependent on the parameters $\theta$ are $\boldsymbol{\hat{x}_{0}}$ and $\boldsymbol{\hat{\Delta}}$. Since, $\boldsymbol{\Delta}$ can be estimated from $\boldsymbol{x_0}$ and $\boldsymbol{y_0}$, then the model solely needs to predict $\boldsymbol{x_0}$. Hence, $\boldsymbol{\hat{x}_{0}} = f_{\theta}(\boldsymbol{x_t}, \boldsymbol{y_0}, t)$ denotes the $\boldsymbol{x_0}$ prediction from a neural network given $\boldsymbol{x_t}$, $\boldsymbol{y_0}$, and timestep $t$. Meanwhile, $\boldsymbol{\hat{\Delta}} = \boldsymbol{\hat{x}_0} - \boldsymbol{y_0}$ represents the $\boldsymbol{\Delta}$ estimation, computed from the $\boldsymbol{x_0}$ prediction and the known $\boldsymbol{y_0}$. The remaining components are fixed hyperparameters and $\boldsymbol{x_t}$, which are known for every reverse transition from $\boldsymbol{x_t}$ at any timestep, $t$. In essence, the approximate reverse transition, $p_{\theta}(\boldsymbol{x_{t-1}}|\boldsymbol{x_t}, \boldsymbol{y_0})$, is modeled as a Gaussian whose mean is computed using a neural network that predicts $\boldsymbol{x_0}$. Accordingly, the KL divergence of each term $\mathcal{L}_{1:T-1}$ can be further expanded as:



\begingroup
\allowdisplaybreaks
\stepcounter{equation}
\begin{align*}
\label{eq:training_objective_mean_difference_surrogate}
& D_{\mathrm{KL}}(q(\boldsymbol{x_{t-1}}|\boldsymbol{x_{t}}, \boldsymbol{x_0}, \boldsymbol{\Delta}) \| p_{\theta}(\boldsymbol{x_{t-1}}|\boldsymbol{x_t}, \boldsymbol{y_0})) \\
&= \begin{cases}
    \frac{1}{2}\|{\boldsymbol{\mu}}_{\boldsymbol{\theta}|\gamma{=}0}\left(\boldsymbol{x_t}, \boldsymbol{y_0}, t\right) - {\boldsymbol{\tilde{\mu}}}_{\boldsymbol{t}|\gamma{=}0}\|^2, & \text{if } {\color{NavyBlue} \text{A}}, \\
    \frac{1}{2}\|{\boldsymbol{\mu}}_{\boldsymbol{\theta}|\gamma{\ne}0}\left(\boldsymbol{x_t}, \boldsymbol{y_0}, t\right) - {\boldsymbol{\tilde{\mu}}}_{\boldsymbol{t}|\gamma{\ne}0}\|^2, & \text{if } {\color{BurntOrange} \text{B}}, \\
    \frac{1}{2{}\tilde{\sigma}^2_{t|\gamma{\ne}0}}\|{\boldsymbol{\mu}}_{\boldsymbol{\theta}|\gamma{\ne}0}\left(\boldsymbol{x_t}, \boldsymbol{y_0}, t\right) - {\boldsymbol{\tilde{\mu}}}_{\boldsymbol{t}|\gamma{\ne}0}\|^2, & \text{if } {\color{ForestGreen} \text{C}}, \\
\end{cases} \\
&= \begin{cases}
    \frac{1}{2}\|\boldsymbol{x_0} - \beta_{t-1}\boldsymbol{\Delta} - (\boldsymbol{\hat{x}_{0}} - \beta_{t-1}\boldsymbol{\hat{\Delta}})\|^2, & \text{if } {\color{NavyBlue} \text{A}}, \\
    \begin{aligned}
        \textstyle\frac{1}{2}\bigg\|&\boldsymbol{x_0} - \beta_{t-1}\boldsymbol{\Delta} + \sqrt{\gamma^2\beta_{t-1} - \eta^2\tilde{\lambda}_t}\left(\textstyle\frac{\boldsymbol{x_t} - \boldsymbol{x_0} + \beta_t\boldsymbol{\Delta}}{{\sqrt{\gamma^2\beta_t}}}\right) \\
        &- \left(\boldsymbol{\hat{x}_{0}} - \beta_{t-1}\boldsymbol{\hat{\Delta}} + \sqrt{\gamma^2\beta_{t-1} - \eta^2\tilde{\lambda}_t}\left(\textstyle\frac{\boldsymbol{x_t} - \boldsymbol{\hat{x}_{0}} + \beta_t\boldsymbol{\hat{\Delta}}}{{\sqrt{\gamma^2\beta_t}}}\right)\right)\bigg\|^2
    \end{aligned}, & \text{if } {\color{BurntOrange} \text{B}}, \\
    \begin{aligned}
        \textstyle\frac{1}{2\eta^2\tilde{\lambda}_t}\bigg\|&\boldsymbol{x_0} - \beta_{t-1}\boldsymbol{\Delta} + \sqrt{\gamma^2\beta_{t-1} - \eta^2\tilde{\lambda}_t}\left(\textstyle\frac{\boldsymbol{x_t} - \boldsymbol{x_0} + \beta_t\boldsymbol{\Delta}}{{\sqrt{\gamma^2\beta_t}}}\right) \\
        &- \left(\boldsymbol{\hat{x}_{0}} - \beta_{t-1}\boldsymbol{\hat{\Delta}} + \sqrt{\gamma^2\beta_{t-1} - \eta^2\tilde{\lambda}_t}\left(\textstyle\frac{\boldsymbol{x_t} - \boldsymbol{\hat{x}_{0}} + \beta_t\boldsymbol{\hat{\Delta}}}{{\sqrt{\gamma^2\beta_t}}}\right)\right)\bigg\|^2
    \end{aligned}, & \text{if } {\color{ForestGreen} \text{C}}, \\
\end{cases} \\
&= \begin{cases}
    \frac{1}{2}\|\boldsymbol{x_0} - \boldsymbol{\hat{x}_{0}} - \beta_{t-1}(\boldsymbol{x_0} - \boldsymbol{\hat{x}_0})\|^2, & \text{if } {\color{NavyBlue} \text{A}}, \\
    \begin{aligned}
        \textstyle\frac{1}{2}\bigg\|&\boldsymbol{x_0} - \boldsymbol{\hat{x}_{0}} - \beta_{t-1}(\boldsymbol{x_0} - \boldsymbol{\hat{x}_0}) \\
        &+ \sqrt{\textstyle\frac{\beta_{t-1}}{{\beta_t}}}(\boldsymbol{\hat{x}_{0}} - \boldsymbol{x_0} + \beta_t(\boldsymbol{x_0} - \boldsymbol{\hat{x}_0} ))\bigg\|^2
    \end{aligned}, & \text{if } {\color{BurntOrange} \text{B}}, \\
    \begin{aligned}
        \textstyle\frac{1}{2\eta^2\tilde{\lambda}_t}\bigg\|&\boldsymbol{x_0} - \boldsymbol{\hat{x}_{0}} - \beta_{t-1}(\boldsymbol{x_0} - \boldsymbol{\hat{x}_0}) \\
        &+ \sqrt{\textstyle\frac{\gamma^2\beta_{t-1} - \eta^2\tilde{\lambda}_t}{{\gamma^2\beta_t}}}(\boldsymbol{\hat{x}_{0}} - \boldsymbol{x_0} + \beta_t(\boldsymbol{x_0} - \boldsymbol{\hat{x}_0} ))\bigg\|^2
    \end{aligned}, & \text{if } {\color{ForestGreen} \text{C}}, \\
\end{cases} \tag{\theequation} \\
&= \begin{cases}
    \frac{1}{2}\|(\boldsymbol{x_0} - \boldsymbol{\hat{x}_{0}})(1 - \beta_{t-1})\|^2, & \text{if } {\color{NavyBlue} \text{A}}, \\
    \begin{aligned}
        \textstyle\frac{1}{2}\bigg\|&(\boldsymbol{x_0} - \boldsymbol{\hat{x}_{0}})(1 - \beta_{t-1}) \\
        &+ \sqrt{\textstyle\frac{\beta_{t-1}}{{\beta_t}}}(\boldsymbol{\hat{x}_{0}} - \boldsymbol{x_0} + \beta_t(\boldsymbol{x_0} - \boldsymbol{\hat{x}_0}))\bigg\|^2
    \end{aligned}, & \text{if } {\color{BurntOrange} \text{B}}, \\
    \begin{aligned}
        \textstyle\frac{1}{2\eta^2\tilde{\lambda}_t}\bigg\|&(\boldsymbol{x_0} - \boldsymbol{\hat{x}_{0}})(1 - \beta_{t-1}) \\
        &+ \sqrt{\textstyle\frac{\gamma^2\beta_{t-1} - \eta^2\tilde{\lambda}_t}{{\gamma^2\beta_t}}}(\boldsymbol{\hat{x}_{0}} - \boldsymbol{x_0} + \beta_t(\boldsymbol{x_0} - \boldsymbol{\hat{x}_0}))\bigg\|^2
    \end{aligned}, & \text{if } {\color{ForestGreen} \text{C}}, \\
\end{cases} \\
&= \begin{cases}
    \frac{1 - \beta_{t-1}}{2}\|\boldsymbol{x_0} - \boldsymbol{\hat{x}_{0}}\|^2, & \text{if } {\color{NavyBlue} \text{A}}, \\
    \frac{1}{2}\bigg\|(\boldsymbol{x_0} - \boldsymbol{\hat{x}_{0}})(1 - \beta_{t-1}) - \sqrt{\frac{\beta_{t-1}}{{\beta_t}}}(\boldsymbol{x_0} - \boldsymbol{\hat{x}_{0}})(1 - \beta_t)\bigg\|^2, & \text{if } {\color{BurntOrange} \text{B}}, \\
    \frac{1}{2\eta^2\tilde{\lambda}_t}\bigg\|(\boldsymbol{x_0} - \boldsymbol{\hat{x}_{0}})(1 - \beta_{t-1}) - \sqrt{\frac{\gamma^2\beta_{t-1} - \eta^2\tilde{\lambda}_t}{{\gamma^2\beta_t}}}(\boldsymbol{x_0} - \boldsymbol{\hat{x}_{0}})(1 - \beta_t)\bigg\|^2, & \text{if } {\color{ForestGreen} \text{C}}, \\
\end{cases} \\
&= \begin{cases}
    \frac{1 - \beta_{t-1}}{2}\|\boldsymbol{x_0} - \boldsymbol{\hat{x}_{0}}\|^2, & \text{if } {\color{NavyBlue} \text{A}}, \\
    \frac{1}{2}\bigg\|(\boldsymbol{x_0} - \boldsymbol{\hat{x}_{0}})\left(1 - \beta_{t-1} - \sqrt{\frac{\beta_{t-1}}{{\beta_t}}}(1 - \beta_t)\right)\bigg\|^2, & \text{if } {\color{BurntOrange} \text{B}}, \\
    \frac{1}{2\eta^2\tilde{\lambda}_t}\bigg\|(\boldsymbol{x_0} - \boldsymbol{\hat{x}_{0}})\left(1 - \beta_{t-1} - \sqrt{\frac{\gamma^2\beta_{t-1} - \eta^2\tilde{\lambda}_t}{{\gamma^2\beta_t}}}(1 - \beta_t)\right)\bigg\|^2, & \text{if } {\color{ForestGreen} \text{C}}, \\
\end{cases} \\
&= \begin{cases}
    \frac{1 - \beta_{t-1}}{2}\|\boldsymbol{x_0} - \boldsymbol{\hat{x}_{0}}\|^2, & \text{if } \gamma = 0, \\
    \frac{1 - \beta_{t-1} - \sqrt{\frac{\beta_{t-1}}{{\beta_t}}}(1 - \beta_t)}{2}\|\boldsymbol{x_0} - \boldsymbol{\hat{x}_{0}}\|^2, & \text{if } \gamma \ne 0 \text{ and } \eta = 0, \\
    \frac{1 - \beta_{t-1} - \sqrt{\frac{\gamma^2\beta_{t-1} - \eta^2\tilde{\lambda}_t}{{\gamma^2\beta_t}}}(1 - \beta_t)}{2\eta^2\tilde{\lambda}_t}\|\boldsymbol{x_0} - \boldsymbol{\hat{x}_{0}}\|^2, & \text{if } \gamma \ne 0 \text{ and } \eta \ne 0,
\end{cases} \\
&= \omega_t(\gamma, \eta, t)\|\boldsymbol{x_0} - \boldsymbol{\hat{x}_{0}}\|^2.
\end{align*}
\endgroup

Therefore, irrespective of the specific values of $\gamma$ and $\eta$, each consistency term $\mathcal{L}_{1:T-1}$ ultimately reduces to the expectation of a weighted squared Euclidean distance between the original data $\boldsymbol{x_0}$ and its prediction, where the expectation is taken over $q(\boldsymbol{x_{t}}|\boldsymbol{x_0}, \boldsymbol{\Delta})$:
\begin{equation}
\label{eq:training_objective_consistency_term} 
\mathcal{L}_{t-1} = \mathbb{E}_{q(\boldsymbol{x_{t}}|\boldsymbol{x_0}, \boldsymbol{\Delta})}\left[\omega_t(\gamma, \eta, t)\|\boldsymbol{x_0} - \boldsymbol{\hat{x}_{0}}\|^2\right],
\end{equation}
with weights $\omega_t(\cdot)$ defined as a function of $\gamma$, $\eta$, and $t$. Essentially, approximating $\boldsymbol{\hat{x}_{0}}$ to the original data effectively ensures that $\boldsymbol{\mu_{\theta}}\left(\boldsymbol{x_t}, \boldsymbol{y_0}, t\right)$ converges to $\boldsymbol{\tilde{\mu}_t}$. As a result, $p_{\theta}(\boldsymbol{x_{t-1}}|\boldsymbol{x_t}, \boldsymbol{y_0})$ accurately models $q(\boldsymbol{x_{t-1}}|\boldsymbol{x_t}, \boldsymbol{x_0}, \boldsymbol{\Delta})$, which is the primary purpose of the consistency terms $\mathcal{L}_{1:T-1}$.

In particular, due to the relationship between $\boldsymbol{x_0}$ and $\boldsymbol{\epsilon}$ given in Equation (\ref{eq:forward_process_cumulative_transition_reparameterization}), the objective derived in Equation (\ref{eq:training_objective_mean_difference_surrogate}) could be converted to predicting noise $\boldsymbol{\epsilon}$ similar to \acp{DDPM} \citep{ho2020denoising}. However, this reformulation of the objective would not be possible with a deterministic forward process ($\gamma=0$), as it works only for cases where noise was added during the forward process ($\gamma \ne 0$). Hence, having the neural network predict $\boldsymbol{x_0}$ directly is preferred for broader applicability and improved generalizability. 

Notably, the mean is continuous at $\gamma = 0$ (see Appendix \ref{app:reverse_process_transition_distribution}), thus there are no problems during gradient computation, such as taking gradients where a function is not differentiable. Nonetheless, for each specific value of $\gamma$, the mean is continuous and the $\gamma$ constant hyperparameter is immutable, i.e., set only once for each model instance, thereby no discontinuity issues would ever arise due to $\gamma$.

\paragraph{Reconstruction term $\mathcal{L}_{0}$.}
The $\mathcal{L}_{0}$ term is essentially the expectation of the negative log-likelihood (NLL) of the original data, $\boldsymbol{x_0}$, conditioned on the first latent variable, $\boldsymbol{x_1}$, and the corrupted version, $\boldsymbol{y_0}$, where the expectation is taken over $\boldsymbol{x_1} \sim q(\boldsymbol{x_{1}}|\boldsymbol{x_0}, \boldsymbol{\Delta})$. In essence, it quantifies how well the model can reconstruct $\boldsymbol{x_0}$ given $\boldsymbol{x_1}$ and $\boldsymbol{y_0}$. Since minimizing the NLL encourages the model to output high-probability (accurate) reconstructions, it can be interpreted as a reconstruction loss. Conceptually, this term acts as a final quality check, ensuring that after practically all the diffusion degradation is removed\footnote{The forward process progressively incorporates degradation and removes $\boldsymbol{\Delta}$. The reverse process removes degradation and reintroduces $\boldsymbol{\Delta}$.} iteratively, the model can accurately reconstruct the original clean data, $\boldsymbol{x_0}$, from the almost degradation-free input, $\boldsymbol{x_1}$. It assures that the model not only learns to refine the data at intermediate timesteps, but also produces outputs consistent with the underlying real data distribution conditioned on $\boldsymbol{y_0}$. As a result, it contributes to aligning the model marginal $p_{\theta}(\boldsymbol{x_{0}}|\boldsymbol{y_0})$ with the true posterior distribution $q(\boldsymbol{x_0}|\boldsymbol{y_0})$ as given in Equation (\ref{eq:target_data_distribution}). Nonetheless, similar to \acp{DDPM}, this term is omitted in practice, since it is implicitly included in a simplified training objective. 

\paragraph{Simplified objective function.}
Since the term $\mathcal{L}_{T}$ can be excluded from the optimization objective, the loss function in Equation (\ref{eq:training_objective_expanded}) becomes:
\begin{equation}
\label{eq:training_objective_no_prior_matching_term}
\begin{split}
\mathcal{L}(\theta) &= \cancel{\mathcal{L}_{T}} + \mathcal{L}_{1:T-1} + \mathcal{L}_{0} = \sum_{t=2}^{T} \mathcal{L}_{t-1} + \mathcal{L}_{0} \\
&= \sum_{t=2}^{T} \mathbb{E}_{q(\boldsymbol{x_{t}}|\boldsymbol{x_0}, \boldsymbol{\Delta})}\left[D_{\mathrm{KL}}(q(\boldsymbol{x_{t-1}}|\boldsymbol{x_{t}}, \boldsymbol{x_0}, \boldsymbol{\Delta}) \| p_{\theta}(\boldsymbol{x_{t-1}}|\boldsymbol{x_t}, \boldsymbol{y_0}))\right] + \mathcal{L}_{0}.
\end{split}
\end{equation}

Following, the term $\mathcal{L}_{0}$ can be omitted, as it is implicitly included by extending the sum to encompass all timesteps, $t \in \{1, 2, \dots, T\}$, thereby accounting for the transition from $\boldsymbol{x_1}$ to $\boldsymbol{x_0}$:
\begin{equation}
\label{eq:training_objective_no_reconstruction_term}
\mathcal{L}(\theta) = \sum_{t=1}^{T} \mathbb{E}_{q(\boldsymbol{x_{t}}|\boldsymbol{x_0}, \boldsymbol{\Delta})}\left[D_{\mathrm{KL}}(q(\boldsymbol{x_{t-1}}|\boldsymbol{x_{t}}, \boldsymbol{x_0}, \boldsymbol{\Delta}) \| p_{\theta}(\boldsymbol{x_{t-1}}|\boldsymbol{x_t}, \boldsymbol{y_0}))\right],
\end{equation}
and given Equation (\ref{eq:training_objective_mean_difference_surrogate}), then:
\begin{equation}
\label{eq:training_objective_consistency_terms} 
\begin{split}
\mathcal{L}(\theta) &= \sum_{t=1}^{T} \mathbb{E}_{q(\boldsymbol{x_{t}}|\boldsymbol{x_0}, \boldsymbol{\Delta})}\left[D_{\mathrm{KL}}(q(\boldsymbol{x_{t-1}}|\boldsymbol{x_{t}}, \boldsymbol{x_0}, \boldsymbol{\Delta}) \| p_{\theta}(\boldsymbol{x_{t-1}}|\boldsymbol{x_t}, \boldsymbol{y_0}))\right] \\
&= \sum_{t=1}^{T} \mathbb{E}_{q(\boldsymbol{x_{t}}|\boldsymbol{x_0}, \boldsymbol{\Delta})}\left[\omega_t(\gamma, \eta, t)\|\boldsymbol{x_0} - \boldsymbol{\hat{x}_{0}}\|^2\right].
\end{split}
\end{equation}

Considering $\tilde{\lambda}_t = \gamma^2\frac{\beta_{t-1}}{\beta_t}\lambda_t$, the weights $\omega_t$ only depend on the predefined $\gamma$, $\eta$, and $\beta$-schedule constant hyperparameters. In many practical implementations, such as \acp{DDPM}, this weighting is often omitted for all timesteps, finding that this still produces excellent results \citep{ho2020denoising, yue2023resshift}. Therefore, the loss function can be further simplified by excluding the scaling:
\begin{equation}
\label{eq:training_objective_consistency_terms_no_scaling}
\mathcal{L}(\theta) = \sum_{t=1}^{T} \mathbb{E}_{q(\boldsymbol{x_{t}}|\boldsymbol{x_0}, \boldsymbol{\Delta})}\left[\|\boldsymbol{x_0} - \boldsymbol{\hat{x}_{0}}\|^2\right],
\end{equation}
and since evaluating the full sum over all time steps is computationally expensive, a single time step can be sampled per training example. This yields an unbiased estimator of the full objective and significantly improves training efficiency:
\begin{equation}
\label{eq:training_objective_function_simplified_final}
\mathcal{L}_{\text{simple}}(\theta) = \mathbb{E}_{\boldsymbol{x_0}, \boldsymbol{\Delta}, t}\left[\|\boldsymbol{x_0} - \boldsymbol{\hat{x}_{0}}\|^2\right],
\end{equation}
where $\boldsymbol{x_t} \sim q(\boldsymbol{x_t}|\boldsymbol{x_0}, \boldsymbol{\Delta})$, $t \sim \mathcal{U}(1, T)$, and the case $t = 1$ corresponds to $\mathcal{L}_{0}$. Consequently, the objective function of \acp{RDIM} simplifies to a squared Euclidean distance between the original data and its prediction. Notably, \ac{RDIM} and ResShift lead to the same training objective, further highlighting that ResShift is a particular case of \ac{RDIM}. This follows from the objective depending only on the marginal distribution $q(\boldsymbol{x_t}|\boldsymbol{x_0}, \boldsymbol{\Delta})$, which both models share. It does not strictly require the forward process to be a Markov chain. 

\section{Lemmas}
\label{app:lemmas}
This section presents lemmas that support this work. These lemmas provide foundational results and properties that support the main arguments and proofs.

\begin{lemma}[\citet{bishop2006pattern}]
\label{lem:marginal_conditional_gaussians}
Given a marginal Gaussian distribution for random variable $\boldsymbol{x}$ and a conditional Gaussian distribution for random variable $\boldsymbol{y}$ given $\boldsymbol{x}$ in the form:
\begin{equation}
\label{eq:marginal_conditional_gaussians_1}
\begin{split}
p(\boldsymbol{x}) &= \mathcal{N}\left(\boldsymbol{x}| \boldsymbol{\mu_x}, \boldsymbol{\Sigma_x}\right), \\
p(\boldsymbol{y}|\boldsymbol{x}) &= \mathcal{N}\left(\boldsymbol{y}|\boldsymbol{C}\boldsymbol{x} + \boldsymbol{c}, \boldsymbol{\Sigma_{y|x}}\right),
\end{split}
\end{equation}
where $\boldsymbol{\mu_x}$, $\boldsymbol{C}$, and $\boldsymbol{c}$ are parameters governing the means, while $\boldsymbol{\Sigma_x}$ and $\boldsymbol{\Sigma_{y|x}}$ denote covariance matrices. Then the marginal distribution of $\boldsymbol{y}$ and the conditional distribution of $\boldsymbol{x}$ given $\boldsymbol{y}$ are in the form:
\begin{equation}
\label{eq:marginal_conditional_gaussians_2}
\begin{split}
p(\boldsymbol{y}) &= \mathcal{N}\left(\boldsymbol{y}| \boldsymbol{C}\boldsymbol{\mu_x} + \boldsymbol{c}, \boldsymbol{\Sigma_{y|x}} + \boldsymbol{C}\boldsymbol{\Sigma_x}\boldsymbol{C}^{\top}\right), \\
p(\boldsymbol{x}|\boldsymbol{y}) &= \mathcal{N}\left(\boldsymbol{x}\Big|\boldsymbol{\Sigma_{x|y}}\left(\boldsymbol{C}^{\top}\boldsymbol{\Sigma}^{-1}_{\boldsymbol{y|x}}\left(\boldsymbol{y} - \boldsymbol{c}\right) + \boldsymbol{\Sigma}^{-1}_{\boldsymbol{x}}\boldsymbol{\mu_x}\right), \boldsymbol{\Sigma_{x|y}}\right),
\end{split}
\end{equation}
with $\boldsymbol{\Sigma_{x|y}}$ representing the conditional covariance matrix of $\boldsymbol{x}$ given $\boldsymbol{y}$, defined as:
\begin{equation}
\label{eq:marginal_conditional_gaussians_3}
\boldsymbol{\Sigma_{x|y}} = \left(\boldsymbol{\Sigma}^{-1}_{\boldsymbol{x}} + \boldsymbol{C}^{\top}\boldsymbol{\Sigma}^{-1}_{\boldsymbol{y|x}}\boldsymbol{C}\right)^{-1}.
\end{equation}
\end{lemma}

\begin{lemma}[\citet{bishop2006pattern}]
\label{lem:conditional_gaussian_linear_mean}
Given a joint Gaussian distribution over random variables $\boldsymbol{x}$ and $\boldsymbol{y}$ of the form:
\begin{equation}
\label{eq:joint_gaussian_partitioned}
p\left( \begin{bmatrix} \boldsymbol{x} \\ \boldsymbol{y} \end{bmatrix} \right) = 
\mathcal{N}\left(
\begin{bmatrix} \boldsymbol{\mu_x} \\ \boldsymbol{\mu_y} \end{bmatrix},
\begin{bmatrix}
\boldsymbol{\Sigma_{xx}} & \boldsymbol{\Sigma_{xy}} \\
\boldsymbol{\Sigma_{yx}} & \boldsymbol{\Sigma_{yy}}
\end{bmatrix}
\right),
\end{equation}
where $\boldsymbol{\mu_x}$ and $\boldsymbol{\mu_y}$ are the mean vectors of $\boldsymbol{x}$ and $\boldsymbol{y}$, respectively, while $\boldsymbol{\Sigma_{xx}}$, $\boldsymbol{\Sigma_{xy}}$, $\boldsymbol{\Sigma_{yx}}$, and $\boldsymbol{\Sigma_{yy}}$ denote covariance matrices. Then the conditional distribution of $\boldsymbol{x}$ given $\boldsymbol{y}$ is Gaussian:
\begin{equation}
\label{eq:conditional_gaussian_partitioned}
p(\boldsymbol{x}|\boldsymbol{y}) = \mathcal{N}\left(
\boldsymbol{x} \Big| \boldsymbol{\mu_{x|y}}, \boldsymbol{\Sigma_{x|y}}
\right),
\end{equation}
with the conditional mean and covariance given by:
\begin{equation}
\label{eq:conditional_gaussian_partitioned_mean_covariance}
\begin{split}
\boldsymbol{\mu_{x|y}} &= \boldsymbol{\mu_x} + \boldsymbol{\Sigma_{xy}} \boldsymbol{\Sigma_{yy}}^{-1} (\boldsymbol{y} - \boldsymbol{\mu_y}), \\
\boldsymbol{\Sigma_{x|y}} &= \boldsymbol{\Sigma_{xx}} - \boldsymbol{\Sigma_{xy}} \boldsymbol{\Sigma_{yy}}^{-1} \boldsymbol{\Sigma_{yx}},
\end{split}
\end{equation}
where the expressions follow from the Schur complement. This result shows that the conditional mean of $\boldsymbol{x}$ given $\boldsymbol{y}$ is a linear function of $\boldsymbol{y}$.
\end{lemma}

\section{Experimental Details and Additional Results}
\label{app:experimental_details_and_additional_results}
This section presents experimental details and additional results that complement those discussed in the main text. 

\subsection{Datasets}
\label{app:datasets}
Experiments were performed across eight subsets, derived from four public data collections, namely \begin{enumerate*}[label=(\roman*)]
\item Fluorescence Microscopy Denoising (FMD) dataset \citep{zhang2019poisson},\label{item:fmd}
\item DIVerse 2K Resolution High Quality Images (DIV2K) dataset \citep{agustsson2017ntire, timofte2017ntire},\label{item:div2k}
\item Smartphone Image Denoising Dataset (SIDD) \citep{abdelhamed2018high, abdelhamed2019ntire}, and\label{item:sidd}
\item Flickr-Faces-HQ (FFHQ) dataset \citep{karras2019style}.\label{item:ffhq}
\end{enumerate*}

The FMD dataset is specifically designed for Poisson-Gaussian denoising tasks and consists of $12{,}000$ real images acquired from representative biological samples, including \ac{BPAE} cells, zebrafish embryos, and mouse brain tissues, using confocal, two-photon, and wide-field modalities. The dataset contains images with multiple noise levels, resulting in several subsets, but only the strongest noise level (labeled raw in \citet{zhang2019poisson}) subsets are considered, thus prioritizing the most challenging conditions. Solely confocal images were used and mouse images are excluded. Accordingly, the two FMD dataset partitions used are Confocal-BPAE-Raw (herein named FMD-BPAE) with $1{,}000$ noisy-clean image pairs and Confocal-Zebrafish-Raw (herein named FMD-Zebrafish) with $1{,}000$ pairs. Moreover, each subset was randomly partitioned into training, test and validation splits, corresponding to $80\%$, $10\%$, and $10\%$ of the data, respectively.

DIV2K is a publicly available benchmark dataset originally introduced for the NTIRE 2017 Challenge on Single Image Super-Resolution. It is specifically designed for \ac{SR} tasks and comprises a collection of \ac{HR} images along with their corresponding \ac{LR} counterparts. Each \ac{HR} image in the dataset is paired with several downscaled versions, generated through different degradation operations and scaling factors of $2$, $3$, and $4$. Particularly, \hl{three} subsets of DIV2K with unknown \hl{and bicubic} degradation operators are used, namely DIV2K-Unknown-$\times2$\hl{,} DIV2K-Unknown-$\times4$\hl{, and DIV2K-Bicubic-$\times4$}. Each subset \hl{includes $1{,}000$ LR-HR image pairs and} is divided into $800$ images used for training, $100$ for validation, and $100$ for testing. The validation split will be employed to evaluate the performance of the models as the testing split is not available.

The SIDD dataset is specifically designed for image denoising tasks, particularly focusing on real-world noisy images captured with smartphone cameras. The dataset consists of $\approx30{,}000$ noisy images with their corresponding clean ground truth, from $10$ scenes under different lighting conditions and using five representative smartphone cameras, hence spanning a wide range of image types and noise levels. Only images from the SIDD-Medium subset are used, comprising $320$ noisy-clean image pairs. Ultimately, SIDD-Medium was randomly partitioned into training, test and validation splits, corresponding to $80\%$, $10\%$, and $10\%$ of the data, respectively.

The FFHQ dataset consists of $70{,}000$ \ac{HQ} human face images, originally created as a benchmark for \acp{GAN}. It contains faces with considerable variation in terms of age, ethnicity, and image background. In this work, it is used for image inpainting, colorization, and deblurring. For computational efficiency, images were downsampled to a quarter of the original resolution using bicubic interpolation. Subsequently, corrupted-original image pairs were generated, resulting in three task-specific subsets, namely FFHQ-Inpainting, FFHQ-Colorization, and FFHQ-Deblurring. For image inpainting, pixels in the original images are randomly masked and set to zero with probability $p_{\text{mask}}=0.5$. For colorization, grayscale inputs are obtained by converting the original RGB images to luminance. For deblurring, synthetic blurred images are generated from ground truth images by applying a Gaussian blur with kernel size $15 \times 15$ and standard deviation $\sigma = 3.0$. Each subset was randomly partitioned into training, validation, and test splits corresponding to $80\%$, $10\%$, and $10\%$ of the data, respectively.

\subsection{Network Architecture}
\label{app:network_architecture}
\ac{RDIM} employs a U-Net-based architecture to predict $\boldsymbol{\hat{x}_{0}}$ at each iteration of the reverse process. As illustrated in Figure \ref{fig:u_net_based_network}, the network is composed of encoder, bottleneck, and decoder blocks, with skip connections linking encoder and decoder blocks at matching spatial resolutions. For \ac{SR} tasks, an upsample block transforms $\boldsymbol{y_0}$ to match the dimensionality (number of channels and resolution) expected by the network. For other tasks, this layer simplifies to a projection layer. At each iteration, the network is conditioned on a timestep embedding, which is computed with sinusoidal positional encoding and transformed through a small multilayer perceptron (MLP) consisting of a fully connected layer, a Swish activation, and a second fully connected layer. This embedding encodes the current diffusion step, providing information about the position within the reverse process.

\begin{figure*}[h]
    \centering
    \includegraphics[width=\textwidth]{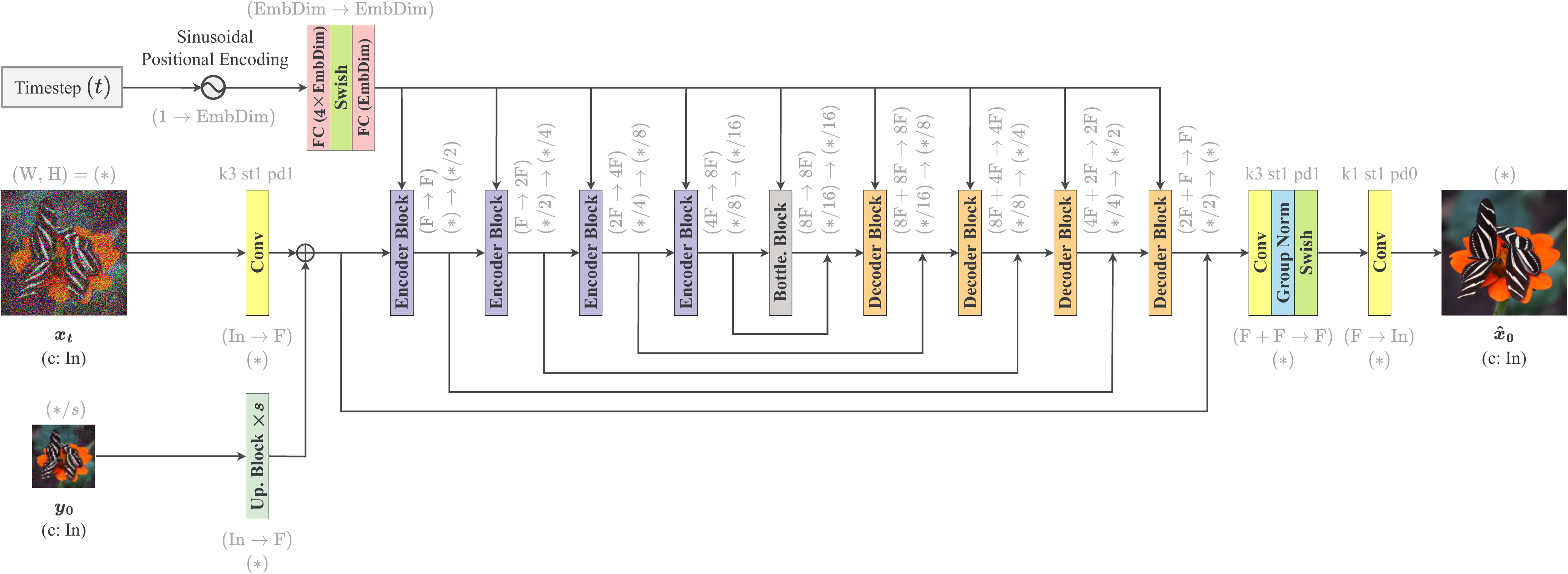}
    \caption{U-Net-based network. In the convolutional layers, the parameters $k$, $st$, and $pd$ represent the kernel size, stride, and padding, respectively. Additionally, $(*/s)$ denotes $(W/s, H/s)$, where $s$ is a scale factor ($s > 1$ for \ac{SR} tasks and $s=1$ otherwise).}
    \label{fig:u_net_based_network}
\end{figure*}

Figure \ref{fig:u_net_building_blocks} shows the core blocks of the network. Each encoder block consists of multiple residual blocks, each optionally followed by a self-attention block, and concludes with a downsample block to reduce spatial resolution. Bottleneck blocks operate at the lowest spatial resolution and consist of multiple residual blocks interleaved with self-attention blocks. Decoder blocks consist of multiple residual blocks, each optionally followed by a self-attention block, and conclude with an upsample block to increase spatial resolution. Notably, all residual blocks incorporate the timestep embedding. Self-attention blocks are included only at the two lowest spatial resolution levels of the encoder and decoder blocks due to computational constraints at higher resolutions.

\begin{figure*}[h]
    \centering
    \includegraphics[width=0.885\textwidth]{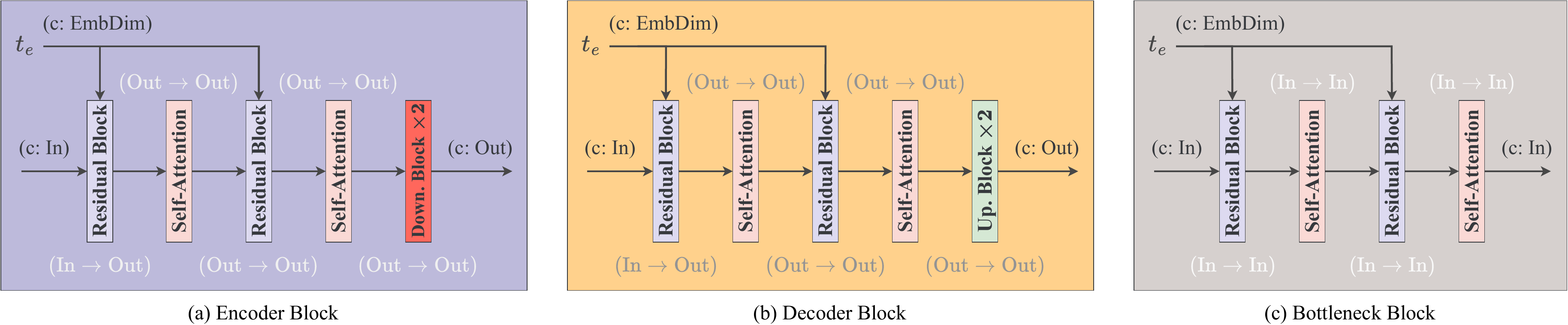}
    \caption{Core blocks of the U-Net-based network. (a) Encoder Block, (b) Decoder Block, and (c) Bottleneck Block.}
    \label{fig:u_net_building_blocks}
\end{figure*}

The building blocks of the network are illustrated in Figure \ref{fig:residual_block}. Each residual block applies two convolutional layers with group normalization and Swish activation. They also contain a projection layer for the timestep embedding, composed of a Swish activation followed by a fully connected layer. Moreover, if the number of input channels (In) does not match the number of output channels (Out), an additional convolutional layer is included in the skip connection to project the input to the expected number of channels (Out), ensuring that the element-wise addition is well-defined. Self-attention blocks model long-range dependencies and incorporate group normalization both before and after the attention mechanism, operating over flattened spatial dimensions. Upsample and downsample blocks perform spatial resizing. Upsample blocks first perform bilinear interpolation (trilinear in case of 3D settings) to increase spatial resolution, followed by a convolutional layer, while downsample blocks perform convolution with stride greater than $1$ ($\text{st} > 1$) to reduce spatial resolution. In the current implementation, activations are omitted, although the generalized block design can optionally include them.

\begin{figure*}[h]
    \centering
    \includegraphics[width=0.885\textwidth]{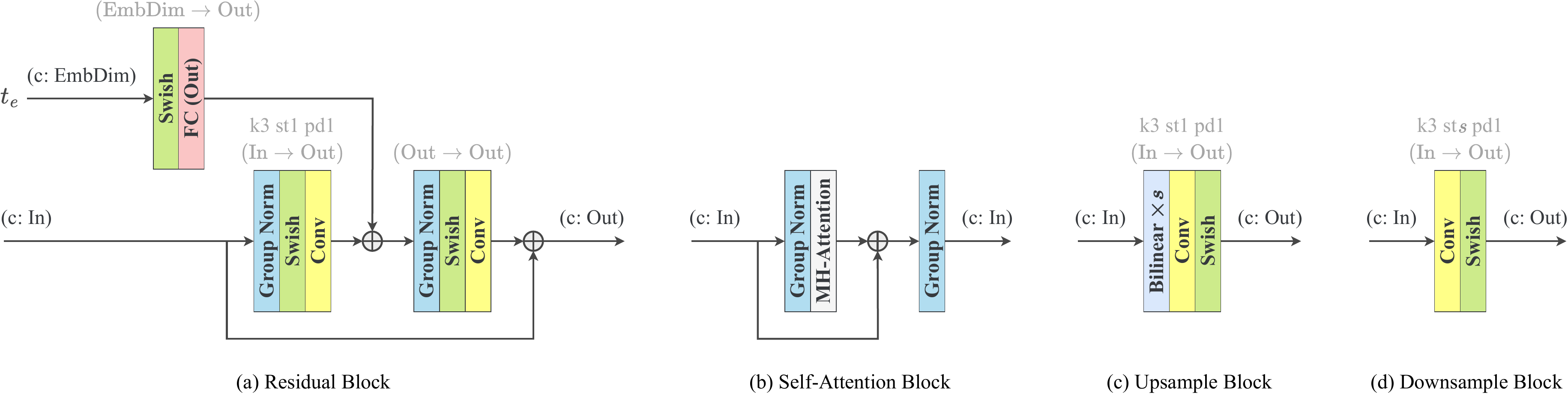}
    \caption{Building blocks. (a) Residual Block, (b) Self-Attention Block, (c) Upsample Block, and (d) Downsample Block.}
    \label{fig:residual_block}
\end{figure*}

\subsection{Implementation Details}
\label{app:implementation_details}
\ac{RDIM} is implemented in PyTorch 2.5.1 \citep{paszke2019pytorch} and trained using the Adam optimizer \citep{kingma2014adam} with $\beta_1=0.9$ and $\beta_2=0.999$. The learning rate was initialized at $1.0\times10^{-4}$ and decayed following a cosine annealing schedule with minimum value \hl{$\eta_{\mathrm{min}}=1.0\times10^{-9}$}. Additionally, RDIM-PQ was trained using a combination of \ac{MSE} and \ac{LPIPS} losses, with the total loss defined as $\mathcal{L} = \mathcal{L}_{\mathrm{MSE}} + \alpha \mathcal{L}_{\mathrm{LPIPS}}$, where $\alpha = 5.0\times10^{-2}$. All experiments were conducted with a batch size of $64$ and an effective patch resolution of $64 \times 64$ \hl{(except for DIV2K-Bicubic-$\times4$, where a larger resolution of $128\times128$ was used)}. For \ac{SR}, this corresponds to \ac{LR} patch sizes of $32 \times 32$ and $16 \times 16$ for $\times 2$ and $\times 4$ scale factors, respectively \hl{(scaled proportionally for DIV2K-Bicubic-$\times4$)}.

\hl{DDPM, DDIM, ResShift, and RDIM} were trained with the same number of diffusion timesteps ($T=50$ for FFHQ and $T=100$ for experiments on FMD, SIDD, and DIV2K) and network architecture with $128$ base channels (detailed in Appendix \ref{app:network_architecture}). The only difference lies in the diffusion framework employed. ResShift is a specific case of \ac{RDIM}, thus a single network was trained for both. For DDPM, following SR3 \citep{saharia2022image}, the model learns to approximate a reverse process, starting from pure Gaussian noise and iteratively denoising $\boldsymbol{x_t}$ toward the \ac{HQ} image, $\boldsymbol{x_0}$, by predicting noise at each step, while conditioned on the \ac{LQ} input, $\boldsymbol{y_0}$. DDIM employed the network trained in the DDPM framework. Training was conducted for $4{,}000{,}000$ iterations on FMD-Confocal datasets and DIV2K-Unknown subsets, $280{,}000$ iterations on the DIV2K-Bicubic-$\times4$ subset, $640{,}000$ iterations on SIDD, and $4{,}375{,}000$ iterations on FFHQ. For \ac{SR} tasks in \ac{RDIM} and ResShift, the \ac{LR} input, $\boldsymbol{y_0}$, is upsampled to the target \ac{HR} resolution using bilinear interpolation, ensuring compatibility with the resolution employed in the diffusion framework (i.e., the size of $\boldsymbol{x_0}, \boldsymbol{x_1}, \dots, \boldsymbol{x_T}$).

All other techniques used in the comparative analysis of Section \ref{sec:experiments} strictly followed the reference papers and the official source codes. BM3D was applied with noise standard deviations of $10$ for FMD-BPAE, $30$ for FMD-Zebrafish, and $50$ for SIDD-Medium. DnCNN was trained for $2{,}500{,}000$ iterations on FMD-Confocal and SIDD datasets. ESRGAN was trained for a total of $1{,}400{,}000$ iterations, with $1{,}000{,}000$ iterations used to train a \ac{PSNR}-oriented model that serves as initialization for the adversarial model, which was optimized for the remaining $400{,}000$ iterations. \hl{Ultimately, GOUB and UniDB were trained for $900{,}000$ iterations on DIV2K-Bicubic-$\times4$, while CTMSR was trained for $500{,}000$ iterations.}

\subsection{Uniform Sampling Timestep Schedule}
\label{app:uniform_sampling_timestep_scheduler}
At inference, \ac{RDIM} intends to reconstruct the original data, $\boldsymbol{x_0}$, starting from the degraded final latent variable, $\boldsymbol{x_{T}}$. Unlike \acp{DDPM} and ResShift, where the sampling process requires iterating over all diffusion timesteps, $T$, the \ac{RDIM} reverse process can be simulated with fewer timesteps. This results from the formulation of the \ac{RDIM} reverse transition, which allows skipping intermediate timesteps during sampling (see Section \ref{sec:long_range_reverse_transition}). Accordingly, this flexibility motivates the selection of a subset, $\Upsilon$, of $S < T$ sampling timesteps to traverse the reverse trajectory.

A simple yet effective approach is to adopt a linear sampling schedule, where the selected timesteps are uniformly spaced. Geometric schedules with denser allocation toward earlier or later stages of the reverse process were empirically evaluated, but they underperformed against a uniform alternative or yielded marginal improvements. As a result, the following uniform scheduler is devised:
\begin{equation}
\label{eq:uniform_sampling_timestep_scheduler}
\Upsilon = \left\{\tau_{k} = \left\lfloor\frac{k}{S} \cdot T\right\rfloor \;\middle|\; k \in \{0, 1, \dots, S\}\right\},
\end{equation}
where, during sampling, $\Upsilon$ is iterated from $\tau_{S} = T$ to $\tau_{1}$, resulting in the order of sampling points $\tau_{S} \to \tau_{S - 1} \to \cdots \to \tau_{1}$. Reverse transitions occur exclusively at these selected timesteps, from each $\tau_t$ to $\tau_{t-1}$, with all intermediate timesteps being skipped. The exception is the target timestep $\tau_{0} = 0$, which marks the end of the reverse trajectory and does not produce a further transition. Moreover, all adjacent sampling timestep pairs, $(\tau_{k-1}, \tau_k)$, satisfy the following condition:
\begin{equation}
\label{eq:uniform_sampling_timestep_scheduler_pairs_condition}
(\tau_{k-1}, \tau_k) \in \left\{(t', t) \in \mathbb{N}_{0}^{2} \mid t' + 1 \leq t \leq T \right\}.
\end{equation}

In essence, only the latent variables associated with these timesteps are sampled, enabling a more efficient inference process. Figure \ref{fig:uniform_sampling_timestep_scheduler} illustrates the sampling points (where reverse transitions occur) along the reverse trajectory, contextualized with the corresponding values of the $\beta$-schedule.

\begin{figure*}[h]
    \centering
    \includegraphics[width=0.216\textwidth]{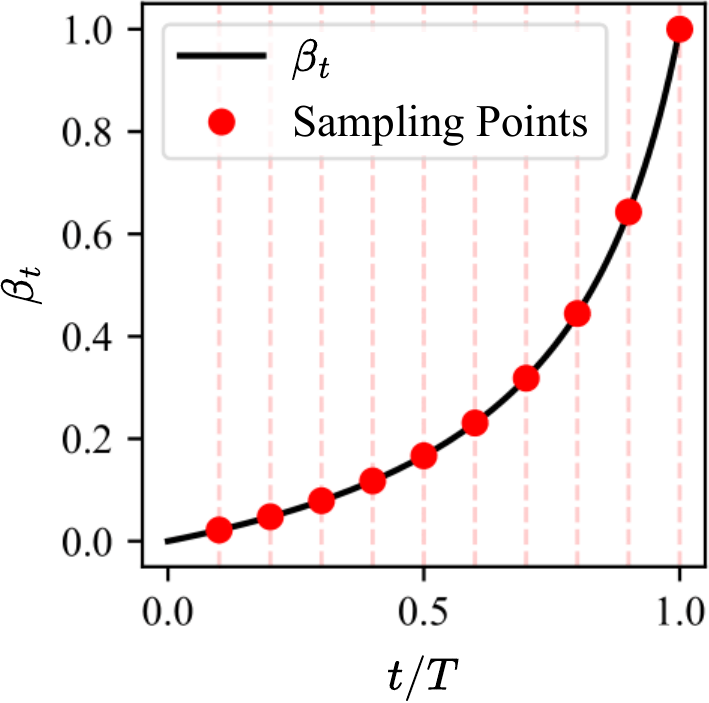}
    \caption{Residual $\beta$-schedule ($p = 5.0$) overlaid with red markers indicating the $\beta_t$ value at each sampling point. The schedule adopted selects timesteps uniformly spaced. In this illustration, the forward process involves $T=100$ timesteps and the number of sampling steps, where reverse transitions occur, is $S=10$.}
    \label{fig:uniform_sampling_timestep_scheduler}
\end{figure*}

\subsection{\texorpdfstring{Impact of $\beta$-schedule parameter $p$}{Impact of Beta-schedule parameter p}}
\label{app:impact_of_beta_schedule_parameter_p}
Experiments were conducted to determine an appropriate value for the $\beta$-schedule parameter $p$ (see Section \ref{sec:residual_beta_schedule}). In these experiments, \ac{RDIM} employs $T=10$ diffusion timesteps and a forward variance hyperparameter $\gamma = 9.0$. During inference, the reverse variance hyperparameter is set to $\eta = 1.0$ and multiple reverse trajectory lengths were evaluated. All other implementation details follow those described in Appendix \ref{app:implementation_details}.

\begin{table}[h!]
  \caption{Impact of parameter $p$, which controls the steepness of the curve in the $\beta$-schedule. All \ac{RDIM} configurations were trained using a forward process with $T=10$ timesteps and variance hyperparameter $\gamma=9.0$. During inference, the reverse process variance hyperparameter is fixed to $\eta=1.0$. \textcolor[rgb]{0.85, 0.65, 0.45}{Orange color} rows highlight ResShift scenarios, corresponding to particular cases where \ac{RDIM} reduces to ResShift under the conditions $\eta=1.0$ and $S=T$.}
  \label{tab:results_rdim_2}
  \centering
  \scriptsize 
  \begin{tabular}{ccccc}
    \toprule
    \multirow{2.6}{*}{$p$} & \multirow{2.6}{*}{$S$} & \multicolumn{3}{c}{FMD-BPAE} \\
    \cmidrule(lr){3-5}
    & & PSNR$\uparrow$ & SSIM$\uparrow$ & LPIPS$\downarrow$ \\
    \midrule
    \multirow{4.2}{*}{1.0} & 1 & 40.0836 & 0.9678 & 0.0205 \\
    \cmidrule(l){2-5}
    & 5 & 40.0772 & 0.9678 & 0.0205 \\
    \cmidrule(l){2-5}
    \rowcolor[rgb]{1.0, 0.91, 0.8}\cellcolor{white} &  10 & 40.0565 & 0.9677 & 0.0205 \\
    \midrule
    \multirow{4.2}{*}{5.0} & 1 & \textbf{40.1100} & \textbf{0.9681} & \textbf{0.0202} \\
    \cmidrule(l){2-5}
    & 5 & 40.0436 & 0.9679 & \textbf{0.0202} \\
    \cmidrule(l){2-5}
    \rowcolor[rgb]{1.0, 0.91, 0.8}\cellcolor{white} &  10 & 39.9524 & 0.9674 & \textbf{0.0202} \\
    \midrule
    \multirow{4.2}{*}{15.0} & 1 & 39.4014 & 0.9639 & 0.0238 \\
    \cmidrule(l){2-5}
    & 5 & 39.1177 & 0.9624 & 0.0236 \\
    \cmidrule(l){2-5}
    \rowcolor[rgb]{1.0, 0.91, 0.8}\cellcolor{white} &  10 & 38.8951 & 0.9609 & 0.0236 \\
    \bottomrule
  \end{tabular}
\end{table}

Table \ref{tab:results_rdim_2} presents the results for $p=1$, $p=5$, and $p=15$. It follows that on denoising images from the FMD-BPAE dataset, \ac{RDIM} with $T=10$ and $\gamma=9.0$ achieves the best overall performance when $p=5.0$. Moreover, irrespective of the steepness of the $\beta$-schedule, skipping timesteps and using fewer reverse timesteps ($S < T$) consistently yields superior results in terms of \ac{PSNR} and \ac{SSIM} compared to iterating through all diffusion steps ($S = T$). Particularly, Figure \ref{fig:beta_schedule} in Section \ref{sec:residual_beta_schedule} illustrates the $\beta$-schedule curves and the effect on the diffusion process corresponding to these parameter values.




\subsection{\texorpdfstring{Impact of variance parameter $\gamma$}{Impact of parameter gamma}}
\label{app:impact_of_parameter_gamma}
The diffusion process variance is controlled with a constant hyperparameter $\gamma \in [0, \infty)$, which allows interpolation between a deterministic $(\gamma = 0 \Rightarrow \text{ Gaussian collapses into a } \delta\text{-distribution})$ and a stochastic $(\gamma > 0)$ forward process. Figure \ref{fig:gamma} illustrates the impact of $\gamma$ on the forward process.

\begin{figure*}[h]
    \centering
    \includegraphics[width=\textwidth]{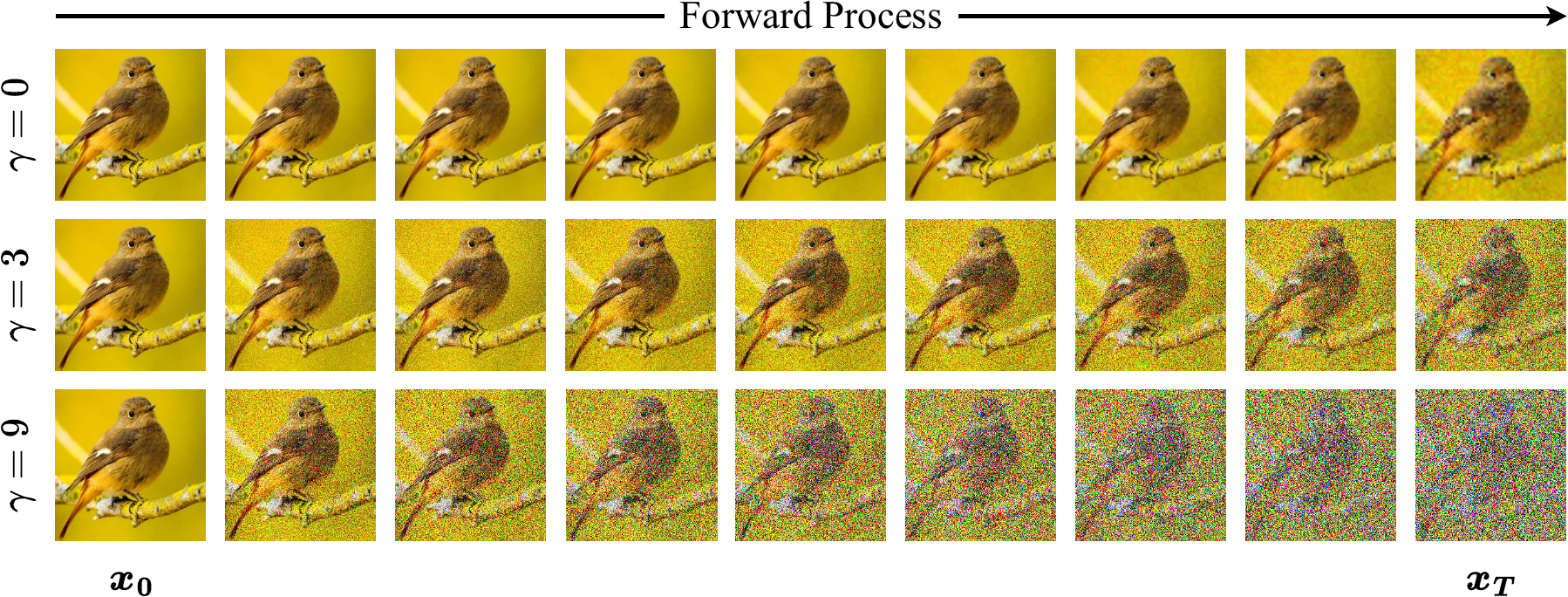}
    \caption{Impact of $\gamma$ on the diffusion process with $\beta$-schedule parameter fixed to $p=5.0$.}
    \label{fig:gamma}
\end{figure*}

\subsection{\texorpdfstring{Comparative analysis of multiple \ac{RDIM} configurations in image denoising}{Comparative analysis of multiple RDIM configurations in image denoising}}
\label{app:comparative_analysis_of_multiple_rdim_configurations_in_image_denoising}
To identify the best \ac{RDIM} configuration, several setups were compared on denoising of BPAE confocal images from the benchmark dataset FMD. The experiments explore the impact of the diffusion chain length ($T$), the number of sampling timesteps ($S$), and the variance controlled by the constant hyperparameters $\gamma$ and $\eta$. Setups with $T = 100$ followed the implementation details described in Appendix \ref{app:implementation_details}. For configurations with a different number of diffusion timesteps, the number of iterations (and consequently the training time) was adjusted linearly in proportion to the number of diffusion steps, $T$. This ensures that each timestep undergoes a similar number of weight updates across all configurations, thereby preventing imbalanced training between timesteps in configurations with different chain lengths. All other implementation details follow those described in Appendix \ref{app:implementation_details}. 

Table \ref{tab:results_fmd_bpae_gamma} summarizes the results. The \ac{RDIM} configuration with $\gamma = 3.0$, $T = 100$, and $S = 10$ achieves the best performance in terms of \ac{PSNR} and \ac{SSIM}. Overall, the results suggest that increasing the number of diffusion timesteps improves denoising performance. Meanwhile, reducing the number of sampling timesteps ($S < T$) often yields better results. In contrast, \ac{LPIPS} scores show that fewer sampling timesteps lead to worse perceptual quality. This highlights a trade-off between content fidelity (measured by \ac{PSNR} and \ac{SSIM}) and perceptual realism (measured by \ac{LPIPS}). Iterative refinement enhances fine-grained details and promotes the recovery of natural textures. Logically, more sampling timesteps allow greater refinement, producing highly realistic outputs. However, results may diverge slightly from the ground truth in terms of pixel-wise similarity, resulting in lower \ac{PSNR} and \ac{SSIM}.

\begin{table}[h!]
  \caption{Denoising performance comparison of several \ac{RDIM} configurations on \ac{BPAE} confocal images from the benchmark dataset FMD. It exhibits the impact of the constant hyperparameter $\gamma$ that controls the variance in the forward process and the impact of the number of diffusion timesteps during training ($T$) and inference ($S$). The constant hyperparameter that controls the variance in the reverse process is fixed to $\eta=1.0$. \textcolor[rgb]{0.85, 0.65, 0.45}{Orange color} rows highlight ResShift scenarios, corresponding to cases where \ac{RDIM} reduces to ResShift ($\eta=1.0$ and $S=T$).}
  \label{tab:results_fmd_bpae_gamma}
  \centering
  \scriptsize 
  \begin{tabular}{ccccccccccc}
    \toprule
    \multirow{4.2}{*}{$T$} & \multirow{4.2}{*}{$S$} & \multicolumn{9}{c}{FMD-BPAE} \\
    \cmidrule(lr){3-11}
    & & \multicolumn{3}{c}{$\gamma = 0.0$} & \multicolumn{3}{c}{$\gamma = 3.0$} & \multicolumn{3}{c}{$\gamma = 9.0$} \\
    \cmidrule(lr){3-5}
    \cmidrule(lr){6-8}
    \cmidrule(lr){9-11}
    & & PSNR$\uparrow$ & SSIM$\uparrow$ & LPIPS$\downarrow$ & PSNR$\uparrow$ & SSIM$\uparrow$ & LPIPS$\downarrow$ & PSNR$\uparrow$ & SSIM$\uparrow$ & LPIPS$\downarrow$ \\
    \midrule
    \multirow{2.6}{*}{10} & 1 & 38.3703 & 0.9575 & 0.0296 & 40.0998 & 0.9686 & 0.0196 & 40.1100 & 0.9681 & 0.0202 \\
    \cmidrule(l){2-11}
    \rowcolor[rgb]{1.0, 0.91, 0.8}\cellcolor{white} & 10 & 38.3487 & 0.9572 & 0.0299 & 39.3632 & 0.9644 & 0.0187 & 39.9524 & 0.9674 & 0.0202 \\
    \midrule
    \multirow{4.2}{*}{50} & 1 & 38.4181 & 0.9578 & 0.0293 & 42.5354 & 0.9803 & 0.0093 & 43.1161 & 0.9821 & 0.0079 \\
    \cmidrule(l){2-11}
    & 10 & 38.4162 & 0.9578 & 0.0295 & 42.4566 & 0.9802 & 0.0079 & 43.2029 & 0.9824 & 0.0077 \\
    \cmidrule(l){2-11}
    \rowcolor[rgb]{1.0, 0.91, 0.8}\cellcolor{white} & 50 & 38.3198 & 0.9564 & 0.0299 & 42.0566 & 0.9785 & 0.0071 & 43.1970 & 0.9824 & 0.0074 \\
    \midrule
    \multirow{4.2}{*}{100} & 1 & 38.3758 & 0.9575 & 0.0295 & 43.9872 & 0.9851 & 0.0056 & 43.3040 & 0.9828 & 0.0075 \\
    \cmidrule(l){2-11}
    & 10 & 38.3752 & 0.9575 & 0.0295 & \textbf{44.1468} & \textbf{0.9855} & 0.0047 & 43.3743 & 0.9830 & 0.0073 \\
    \cmidrule(l){2-11}
    \rowcolor[rgb]{1.0, 0.91, 0.8}\cellcolor{white} &  100 & 38.1775 & 0.9548 & 0.0298 & 43.5990 & 0.9837 & \textbf{0.0042} & 43.2484 & 0.9826 & 0.0068 \\
    \bottomrule
  \end{tabular}
\end{table}

Moreover, results further indicate that controlled stochasticity in the forward process is beneficial. Setting $\gamma = 0.0$ leads to poor results, indicating that some variance is necessary. Conversely, $\gamma = 3.0$ and $\gamma = 9.0$ achieve significantly superior performance. Particularly, $\gamma = 9.0$ outperforms $\gamma = 3.0$ for configurations with few diffusion timesteps, but its relative performance gains diminish as $T$ increases, whereas $\gamma = 3.0$ continues to improve with longer diffusion chains, ultimately surpassing $\gamma = 9.0$ for larger $T$. These observations underline the importance of carefully balancing variance.

Figure \ref{fig:results_fmd_bpae_sampling_timesteps} showcases \ac{PSNR}, \ac{SSIM}, and \ac{LPIPS} scores for different numbers of sampling timesteps, using the configuration with $T = 100$ and $\gamma = 3.0$, which obtained the best results in denoising BPAE confocal images from the FMD dataset. The number of sampling timesteps evaluated includes a single-step prediction and then ranges from $10$ to $100$ in increments of $10$. It follows that \ac{PSNR} and \ac{SSIM} performances peak around $S = 10$, while \ac{LPIPS} achieves the best scores between $S = 90$ and $S = 100$, showing that more sampling timesteps result in higher perceptual quality but reduced reconstruction fidelity.

\begin{figure*}[h]
    \centering
    \includegraphics[width=0.691\textwidth]{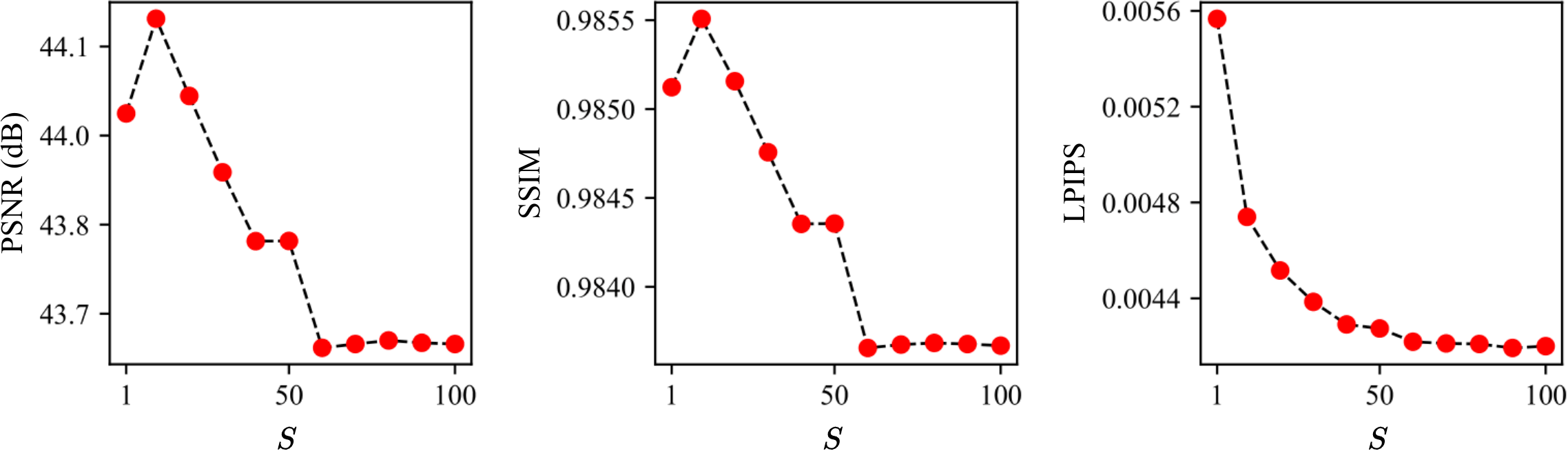}
    \caption{\ac{PSNR}, \ac{SSIM}, and \ac{LPIPS} performance as a function of $S$ (number of sampling timesteps) on denoising of BPAE confocal images from the FMD dataset. \ac{RDIM} is trained with $T = 100$ and $\gamma = 3.0$. The constant hyperparameter that controls the variance in the reverse process is fixed to $\eta=1.0$. The number of sampling timesteps evaluated includes a single-step prediction and then ranges from $10$ to $100$ in increments of $10$.}
    \label{fig:results_fmd_bpae_sampling_timesteps}
\end{figure*}

Table \ref{tab:results_fmd_bpae_eta} demonstrates the effect of varying the constant hyperparameter $\eta$, which controls the variance in the reverse process. Looking at Table \ref{tab:results_fmd_bpae_eta}, the parameter $\eta$ manifests marginal impact on denoising of BPAE confocal images from the FMD dataset. Additionally, when $\gamma = 0$, the parameter $\eta$ does not affect performance, as $\eta$ is absent in the reparameterized form of $p_{\theta}(\boldsymbol{x_{t-1}}|\boldsymbol{x_t}, \boldsymbol{y_0})_{\gamma = 0}$ (see Equations (\ref{eq:reverse_process_transition_approximation_variance}) and (\ref{eq:reverse_process_transition_approximation_mean})). 

\begin{table}[h!]
  \caption{Impact of the constant hyperparameter $\eta$, which controls the variance in the reverse process, on denoising \ac{BPAE} confocal images from the benchmark dataset FMD.}
  \label{tab:results_fmd_bpae_eta}
  \centering
  \scriptsize 
  \begin{tabular}{ccccccccccccc}
    \toprule
    \multirow{4.2}{*}{$T$} & \multirow{4.2}{*}{$S$} & \multirow{4.2}{*}{$\eta$} & \multicolumn{9}{c}{FMD-BPAE} \\
    \cmidrule(lr){4-12}
    & & & \multicolumn{3}{c}{$\gamma = 0.0$} & \multicolumn{3}{c}{$\gamma = 3.0$} & \multicolumn{3}{c}{$\gamma = 9.0$} \\
    \cmidrule(lr){4-6}
    \cmidrule(lr){7-9}
    \cmidrule(lr){10-12}
    & & & PSNR$\uparrow$ & SSIM$\uparrow$ & LPIPS$\downarrow$ & PSNR$\uparrow$ & SSIM$\uparrow$ & LPIPS$\downarrow$ & PSNR$\uparrow$ & SSIM$\uparrow$ & LPIPS$\downarrow$ \\
    \midrule
    \multirow{4.2}{*}{100} & \multirow{4.2}{*}{10} & 0.0 & 38.3752 & 0.9575 & 0.0295 & 44.1429 & 0.9855 & \textbf{0.0047} & 43.3732 & 0.9830 & 0.0073 \\
    \cmidrule(l){3-12}
    & & 0.5 & 38.3752 & 0.9575 & 0.0295 & 44.1440 & 0.9855 & \textbf{0.0047} & 43.3738 & 0.9830 & 0.0073 \\
    \cmidrule(l){3-12}
    & & 1.0 & 38.3752 & 0.9575 & 0.0295 & \textbf{44.1468} & \textbf{0.9855} & \textbf{0.0047} & 43.3743 & 0.9830 & 0.0073 \\
    \bottomrule
  \end{tabular}
\end{table}

\newpage

\subsection{Comparing the perception-distortion trade-off against recent work}
\label{appendix:perception-distortion trade-off}

To provide a more complete assessment of reconstruction quality, we follow the framework established in \citet{blau2018perception}. Traditional distortion metrics such as PSNR strongly penalize any deviation from the exact ground truth, often driving models toward overly smooth or conservative solutions. In contrast, perceptual metrics such as LPIPS capture human-aligned similarity in deep feature space and reward reconstructions that preserve realistic texture and structure, even at the cost of introducing plausible high-frequency hallucinations. While such hallucinated details can be undesirable in certain reconstruction domains (e.g., in medical imaging), they offer a useful lens for quantifying perceptual fidelity. Since different diffusion-based frameworks are optimized with varying objectives, plotting PSNR against LPIPS provides a principled way to visualize and measure their position along the perception–distortion trade-off.

\begin{figure}[h]
    \centering
    \includegraphics[width=0.723\linewidth]{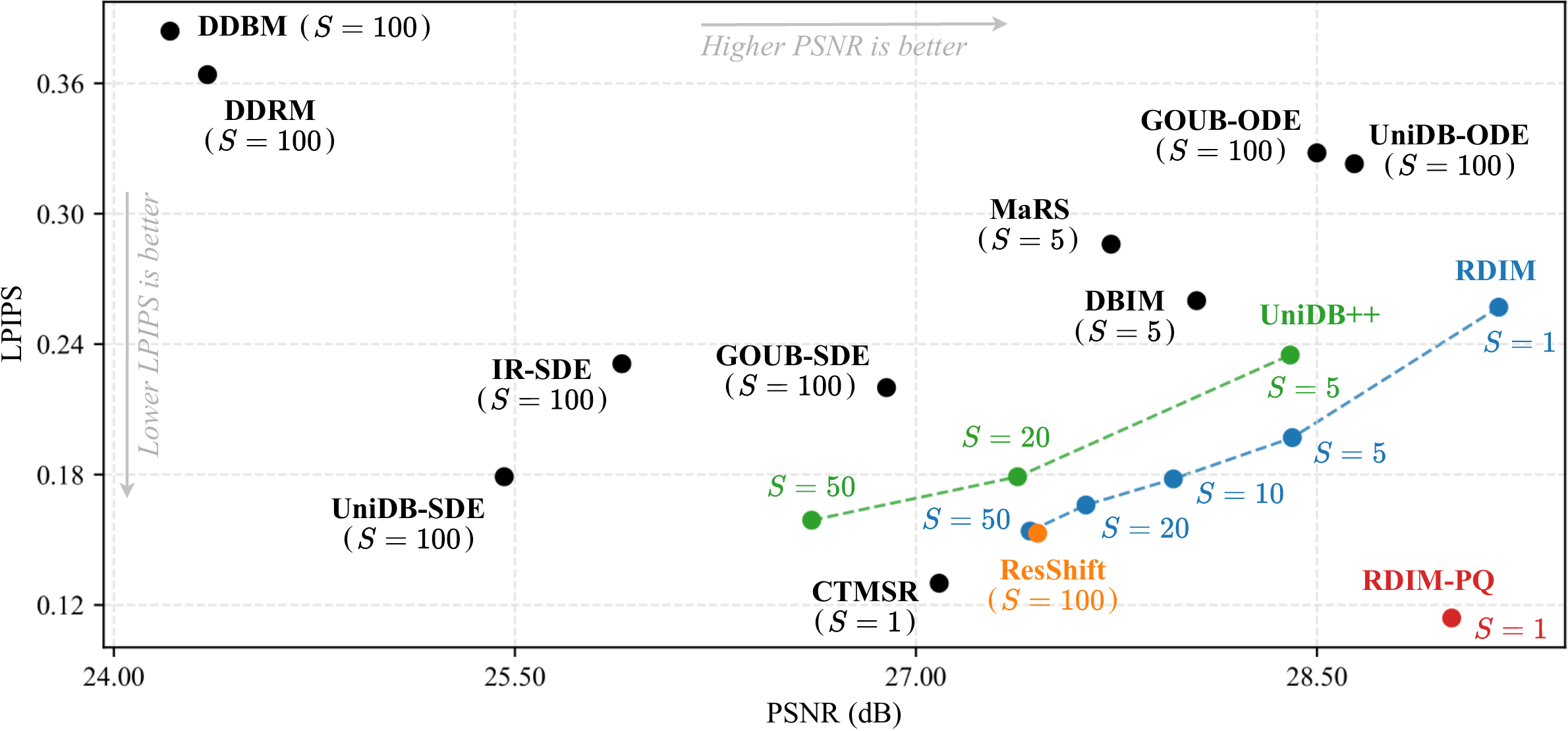}
    \caption{Comparison of the perception-distortion trade-off between the proposed RDIM and the state-of-the-art on the DIV2K dataset for 4$\times$ SR.}
    \label{fig:perception-distortion tradeoff}
\end{figure}

Figure~\ref{fig:perception-distortion tradeoff} compares the proposed method against existing techniques. For RDIM and UniDB++,
we exploit the methods native fast-sampling capabilities to extract multiple operating points along the perception–distortion curve. The resulting comparison shows that the proposed RDIM achieves a substantially improved perception–distortion profile against all other methods. In particular, when comparing with the most recent UniDB++ framework, which relies on a diffusion bridge-based on Doob’s $h$-transform~\mbox{\citep{pan2025unidb++}}, RDIM delivers significant PSNR gains for similar LPIPS values.

Moreover, CTMSR achieves a competitive LPIPS score as it explicitly optimizes this perceptual metric, effectively trading distortion for improved visual quality. Naturally, this comes at the expense of reduced PSNR. In contrast, RDIM-PQ-$1$ attains even lower perceptual scores while simultaneously yielding substantially lower distortion. This property is particularly advantageous for applications requiring high fidelity to the original signal, such as medical imaging, scientific microscopy, satellite and aerial sensing, and downstream vision tasks where hallucinated details could compromise reliability.

\subsection{Quantitative Results in Additional Image Restoration Tasks}
\label{appendix:quantitative_results_in_additional_image_restoration_tasks}
In addition to the qualitative examples in Figure \ref{fig:results_ffhq}, quantitative results for inpainting, colorization, and deblurring are reported in Table \ref{tab:results_other_tasks}. Across all three tasks, RDIM maintains the same trends observed in denoising and \ac{SR}. RDIM-$1$ consistently achieves the highest PSNR and SSIM, while RDIM-$10$ provides competitive performance with slightly better perceptual quality (LPIPS). RDIM surpasses ResShift in every metric except a few LPIPS cases, indicating that RDIM produces sharper and more faithful reconstructions even in challenging restoration settings.

\begin{table}[h!]
  \caption{Performance on the FFHQ dataset for image inpainting, colorization, and deblurring.}
  \label{tab:results_other_tasks}
  \centering
  \scriptsize 
  \begin{tabular}{cccccccccc}
    \toprule
    \multirow{2.6}{*}{\textbf{Method}} & \multicolumn{3}{c}{FFHQ-Inpainting} & \multicolumn{3}{c}{FFHQ-Colorization} & \multicolumn{3}{c}{FFHQ-Deblurring} \\
    \cmidrule(lr){2-4}
    \cmidrule(lr){5-7}
    \cmidrule(lr){8-10}
    & PSNR$\uparrow$ & SSIM$\uparrow$ & LPIPS$\downarrow$ & PSNR$\uparrow$ & SSIM$\uparrow$ & LPIPS$\downarrow$ & PSNR$\uparrow$ & SSIM$\uparrow$ & LPIPS$\downarrow$ \\
    \midrule
    Corrupted & 9.034 & 0.136 & 1.165 & 20.806$^\dagger$ & 0.926$^\dagger$ & 0.223$^\dagger$ & 24.691 & 0.686 & 0.486 \\
    ResShift & 31.721 & 0.922 & \textbf{0.022} & 25.320 & 0.948 & 0.103 & 28.331 & 0.812 & \textbf{0.089} \\
    \midrule
    RDIM-$1$ & \textbf{33.514} & \textbf{0.941} & 0.029 & \textbf{25.727} & \textbf{0.950} & \textbf{0.094} & \textbf{29.995} & \textbf{0.847} & 0.170 \\
    RDIM-$10$ & 32.330 & 0.931 & 0.025 & 25.565 & 0.949 & 0.099 & 28.870 & 0.826 & 0.111 \\
    \bottomrule \\[-0.85em]
    \multicolumn{10}{l}{$^\dagger$~To compute colorization scores on corrupted images, the existing channel is replicated.}
  \end{tabular}
\end{table}

Notably, these experiments use a forward diffusion process with $T = 50$ steps, which is half the length employed in the denoising and \ac{SR} experiments. As a result, performances reported in Table \ref{tab:results_other_tasks} still have margin for improvement when adopting longer diffusion chain lengths (see Appendix \ref{app:comparative_analysis_of_multiple_rdim_configurations_in_image_denoising}).

\subsection{Performance when the forward model is mismatched}
\label{appendix:performance_when_the_forward_model_is_mismatched}
To evaluate model robustness when the forward model is mismatched (i.e., when the testing data contain unseen degradations), signal-dependent Poisson noise is simulated and applied to the LR images of the DIV2K-Bicubic-$\times4$ validation set. A fixed peak photon count of $\lambda_{peak} = 1.0\times10^{3}$ is used. This corresponds to a moderate imaging scenario (e.g., indoor lighting or mid-ISO conditions). Notably, the models were trained on the standard DIV2K-Bicubic-$\times4$ dataset without incorporating Poisson noise.

\begin{table}[h!]
  \caption{Evaluation on Poisson-corrupted DIV2K-Bicubic-$\times4$ images after training on the original dataset without adding Poisson noise. Values in parentheses indicate the performance drop relative to the evaluation on the original noise-free DIV2K-Bicubic-$\times4$ dataset (see Table~\ref{tab:results_div2k_x4}). \textcolor{red}{Red color} indicates the worst performance drop overall and \textcolor{ForestGreen}{Green color} the best.}
  \label{tab:div2k_x4_poisson}
  \centering
  \scriptsize 
  \begin{tabular}{cccc}
    \toprule
    \multirow{2.6}{*}{\textbf{Method}} & \multicolumn{3}{c}{Poisson-corrupted DIV2K-Bicubic-$\times4$} \\
    \cmidrule(lr){2-4}
    & PSNR$\uparrow$ & SSIM$\uparrow$ & LPIPS$\downarrow$ \\
    \midrule
    ResShift & 23.610 (-3.845) & 0.521 (-0.259) & 0.428 (+0.275) \\
    GOUB-SDE & 20.165 ({\color{red}{-6.725}}) & 0.335 ({\color{red}{-0.413}}) & 0.664 (+0.444) \\
    UniDB-SDE & 19.284 (-6.176) & 0.314 (-0.372) & 0.697 ({\color{red}{+0.518}}) \\
    CTMSR-$1$ & 22.987 (-4.100) & 0.466 (-0.293) & 0.495 (+0.365) \\
    \midrule
    RDIM-$1$ & 26.260 ({\color{ForestGreen}{-2.920}}) & 0.673 ({\color{ForestGreen}{-0.151}}) & 0.519 (+0.262) \\
    RDIM-$10$ & 24.363 (-3.600) & 0.564 (-0.231) & 0.415 ({\color{ForestGreen}{+0.237}}) \\
    \bottomrule
  \end{tabular}
\end{table}

As expected, performance decreases for all methods when evaluated on unseen Poisson-corrupted images, reflecting the sensitivity of supervised reconstruction to mismatched forward degradations. Particularly, GOUB and UniDB suffer the largest drops in PSNR, SSIM, and LPIPS. This can be attributed to their bridge formulation, which relies on fixed endpoint distributions and tightly couples the reconstruction process to the forward degradation operator. When this endpoint shifts due to unseen Poisson noise, the learned bridge becomes misaligned, causing the reverse dynamics to deviate from the correct posterior and leading to substantially larger reconstruction errors. Meanwhile, CTMSR-$1$ is comparatively less affected, but its LPIPS score still deteriorates substantially, which is particularly striking given that it directly optimizes for perceptual quality. This underscores the difficulty of maintaining perceptual fidelity under unseen Poisson noise.

In contrast, \ac{RDIM} demonstrates superior robustness with \ac{RDIM}-$1$ exhibiting the smallest drop in PSNR ($-2.920$ dB) and SSIM ($-0.151$). Moreover, \ac{RDIM}-$10$ obtains the least increase in LPIPS ($+0.237$), followed by \ac{RDIM}-$1$ comparatively modest increase ($+0.262$). Unlike bridge-based methods with fixed endpoint constraints, \ac{RDIM} does not rely on a strictly specified degradation endpoint, and its controllable variance forward process allows the last latent variable in the forward process, $\boldsymbol{x_T}$, to remain near the corrupted observation, $\boldsymbol{y_0}$ (i.e., the \ac{LQ} image), without being tied to it. This flexibility helps the model remain consistent even when the degradation shifts. In addition, the few-step reconstruction enabled by the implicit sampling strategy reduces error accumulation, which is particularly beneficial under forward model mismatch. Overall, these results indicate that \ac{RDIM} not only achieves state-of-the-art performance under matched conditions but also retains faithful reconstructions under moderate deviations from the training degradation model, highlighting its practical robustness for real-world image restoration scenarios.


\newpage

\subsection{Additional Qualitative Results}
\label{app:additional_qualitative_results}

\begin{figure*}[!h]
    \centering
    \includegraphics[width=\textwidth]{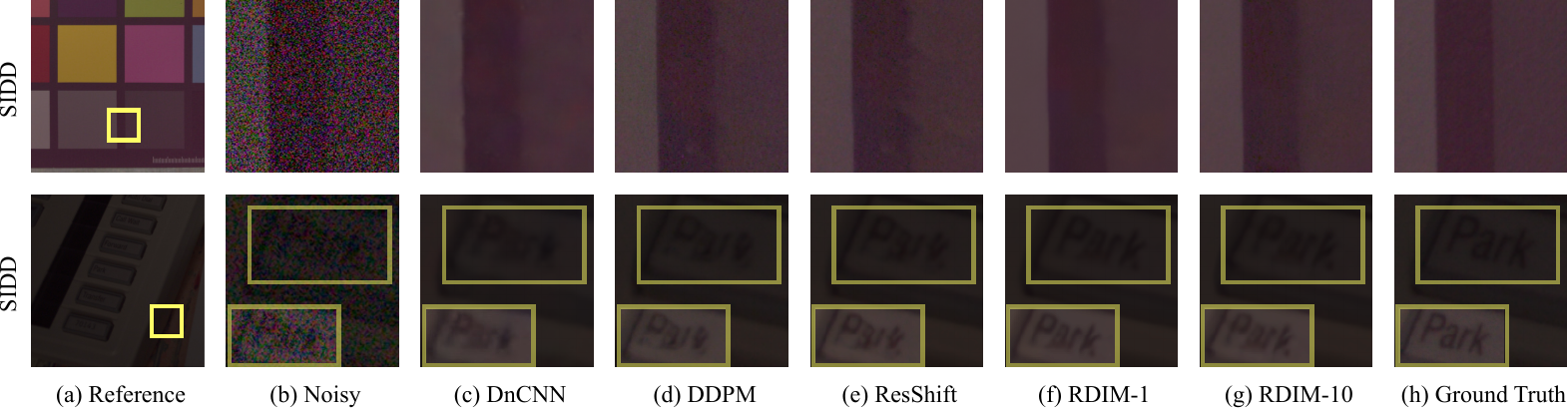}
    \caption{Qualitative denoising analysis on cropped regions from the SIDD dataset. Since SIDD contains noisy images captured under challenging lighting conditions, brightness-adjusted crops of the marked regions are shown in the bottom row for enhanced visualization.}
    \label{fig:results_sidd}
\end{figure*}

\begin{figure*}[h]
    \centering
    \includegraphics[width=\textwidth]{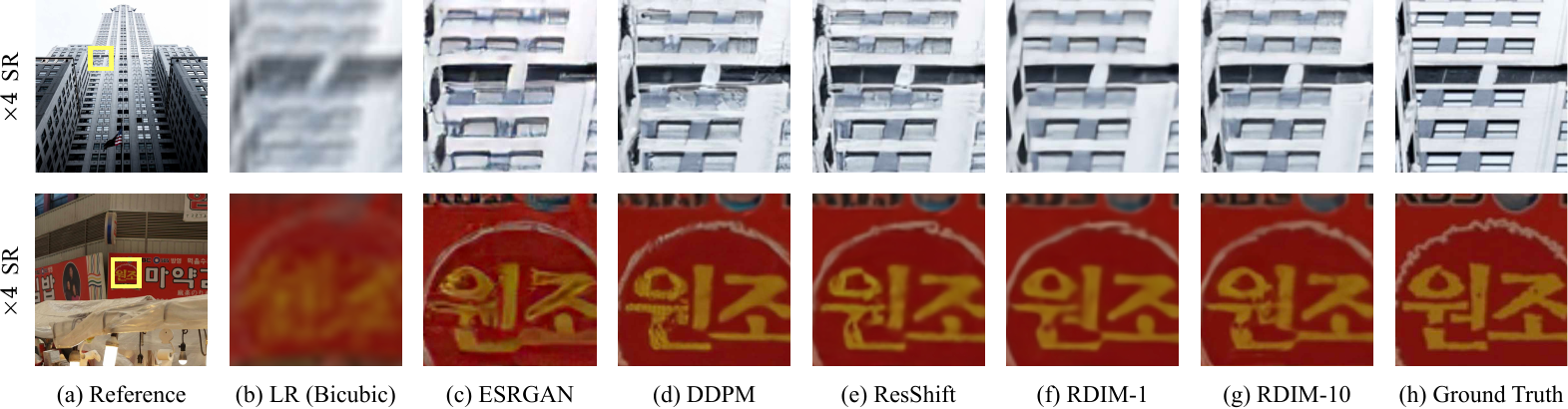}
    \caption{\ac{SR} qualitative comparison on cropped regions from the DIV2K subset with unknown degradation.}
    \label{fig:results_div2k}
\end{figure*}

\begin{figure*}[h]
    \centering
    \includegraphics[width=\textwidth]{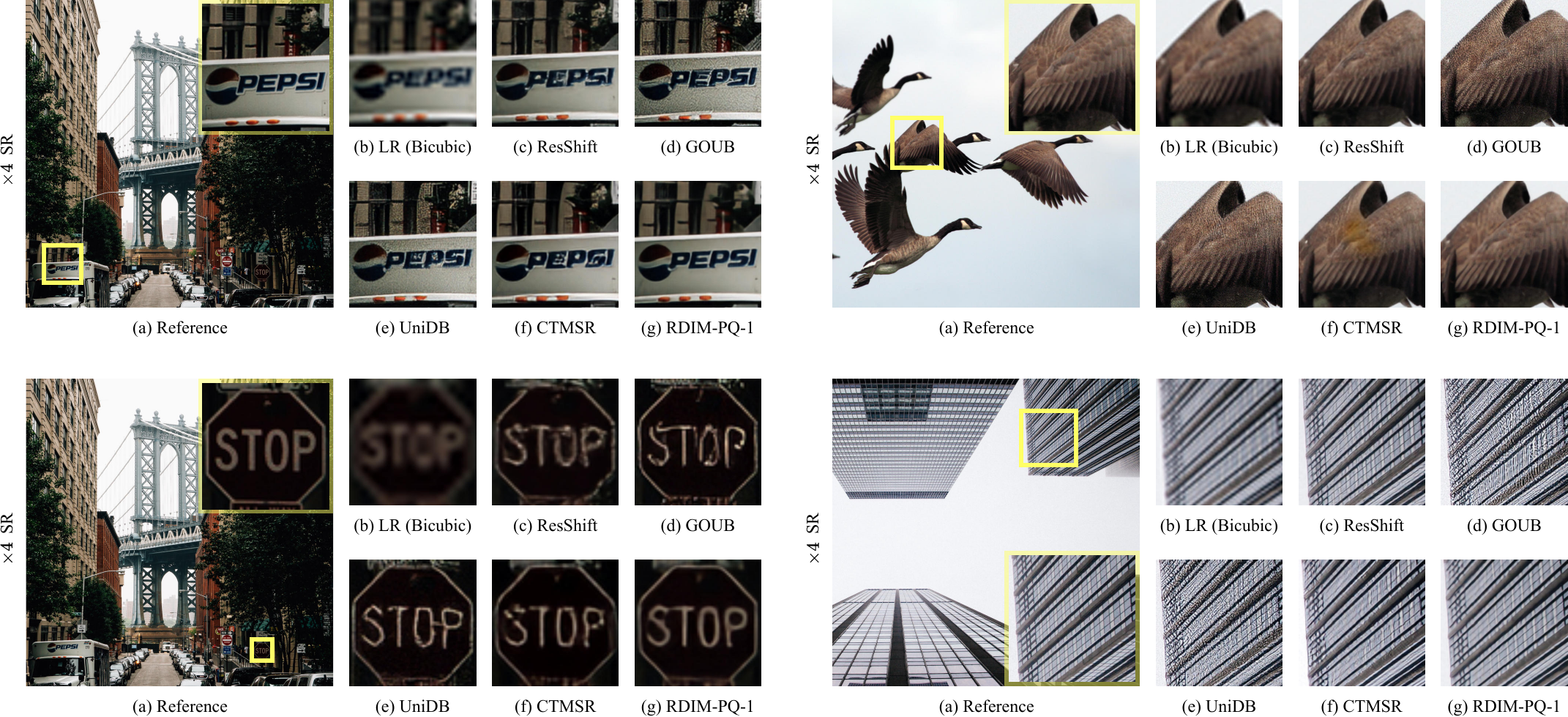}
    \caption{\ac{SR} qualitative comparison on cropped regions from the DIV2K subset with bicubic downsampled images.}
    \label{fig:results_div2k_x4_bicubic_additional_1}
\end{figure*}

\begin{figure*}[h]
    \centering
    \includegraphics[width=\textwidth]{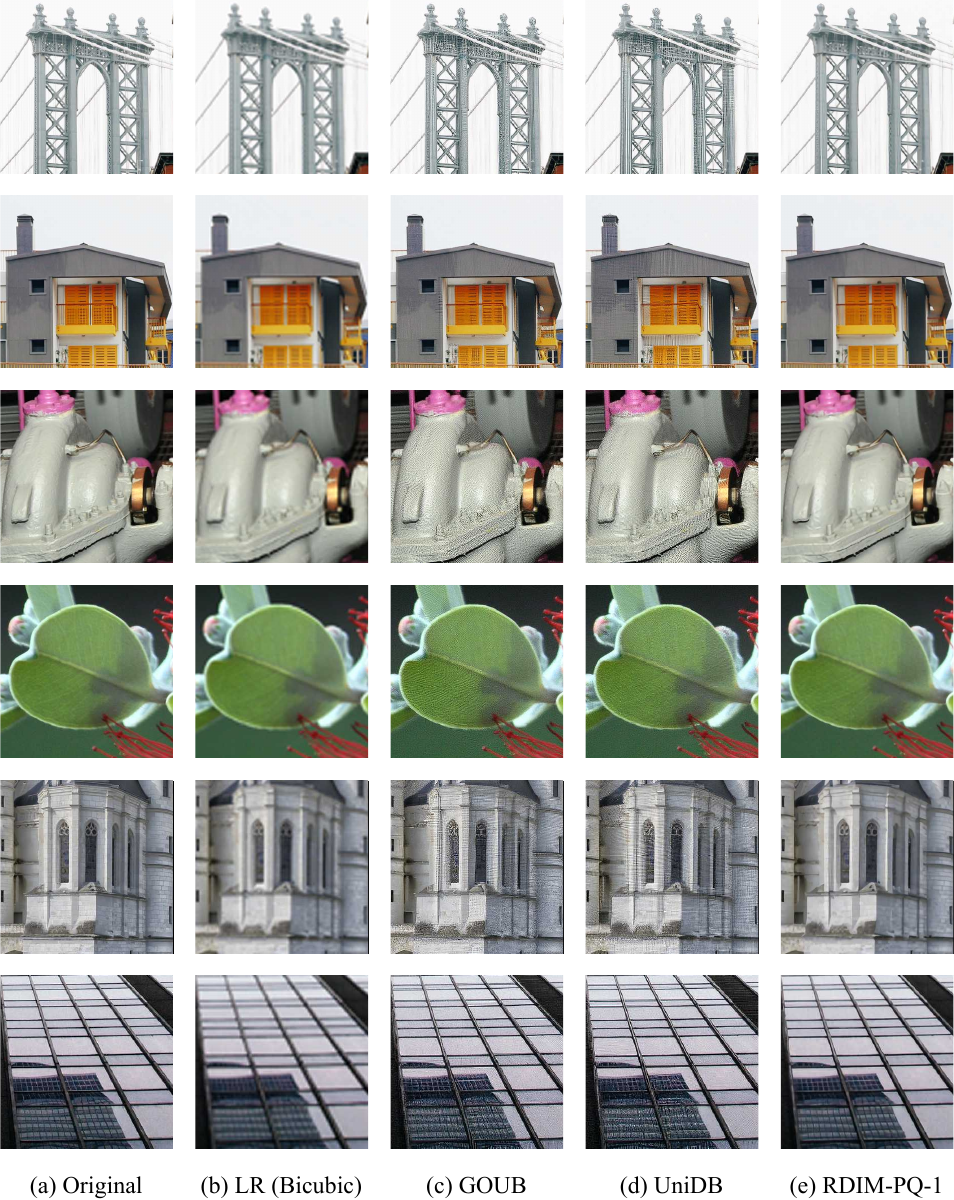}
    \caption{Qualitative comparison of RDIM-PQ-$1$ on $\times4$ \ac{SR}. Cropped regions from the DIV2K subset, with bicubic downsampled images, suggest RDIM achieves greater structural and texture fidelity than bridge-based models.}
    \label{fig:results_div2k_x4_bicubic_additional_2}
\end{figure*}

\begin{figure*}[h]
    \centering
    \includegraphics[width=\textwidth]{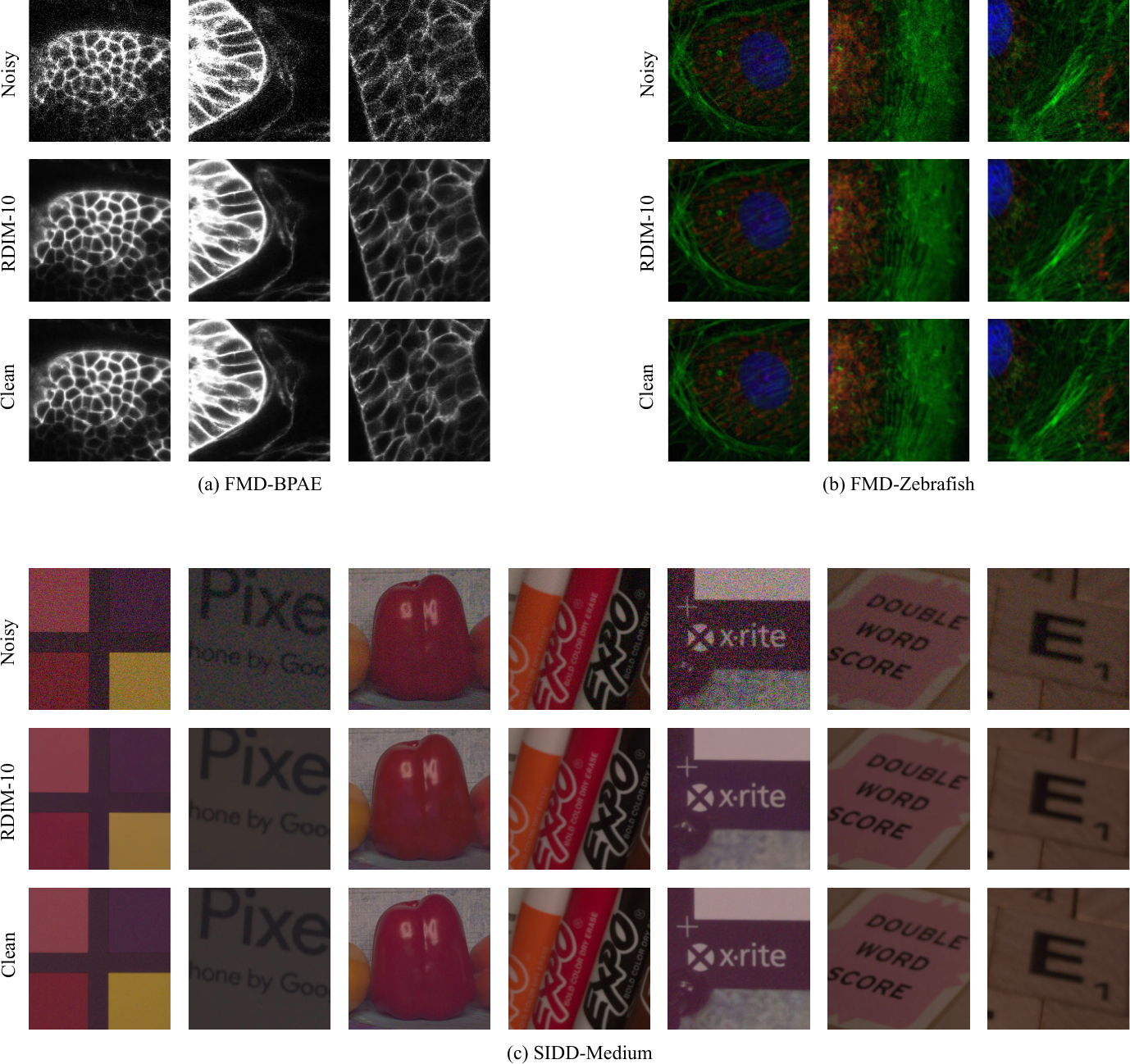}
    \caption{Denoising results of \ac{RDIM}-$10$ on images from the FMD and SIDD datasets. For improved visualization, only cropped regions are shown. \ac{RDIM} is trained with $T=100$ and $\gamma = 3.0$. Inference is conducted with $S=10$ and $\eta = 1.0$.}
    \label{fig:denoising_results_2}
\end{figure*}

\begin{figure*}[h]
    \centering
    \includegraphics[width=\textwidth]{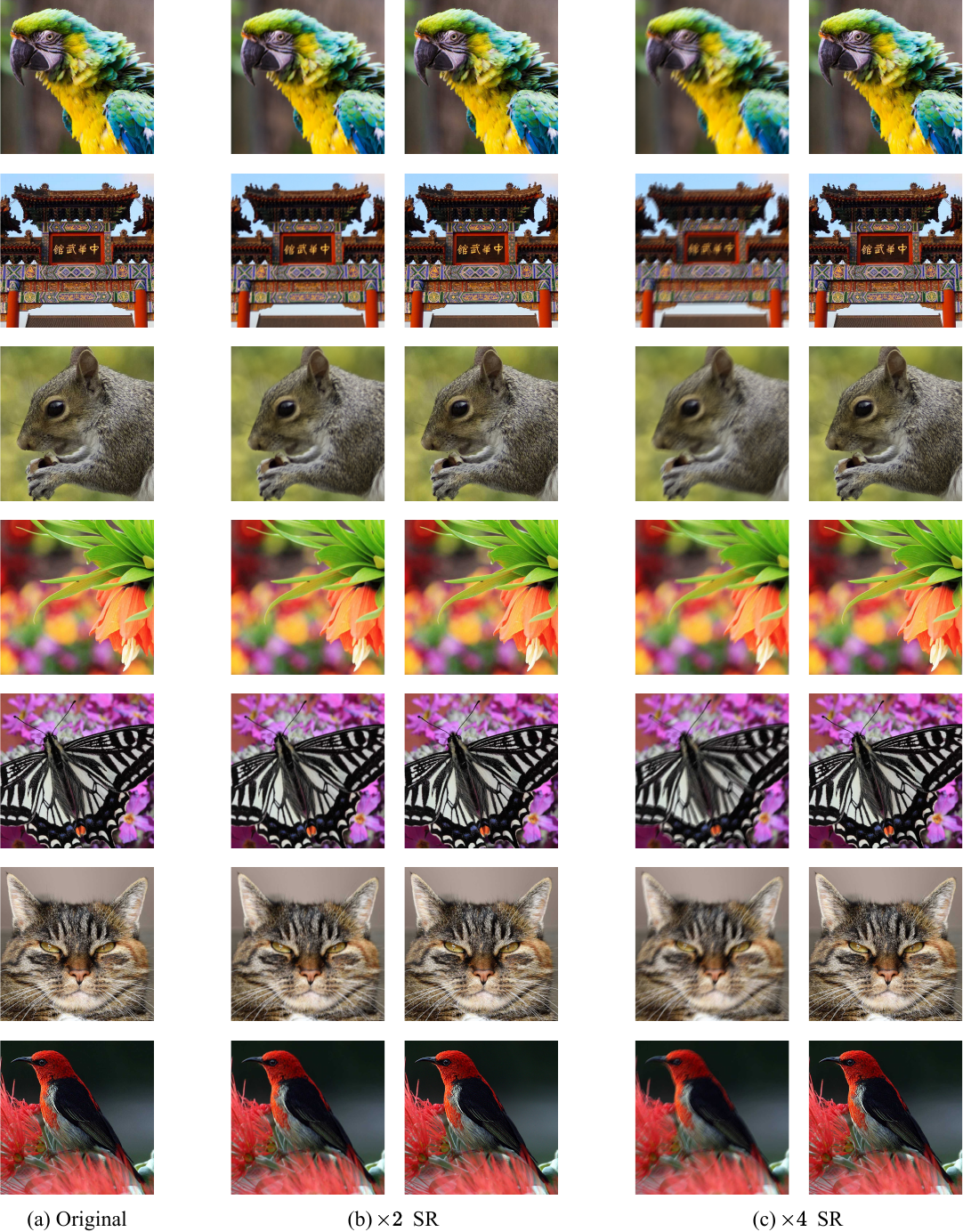}
    \caption{$\times2$ and $\times4$ \ac{SR} results of \ac{RDIM}-$10$ on images from the DIV2K dataset under unknown degradations. \ac{RDIM} is trained with $T=100$ and $\gamma = 3.0$. Inference is conducted with $S=10$ and $\eta = 1.0$. In (b) and (c), the left side represents the input image and the right side the output.}
    \label{fig:sr_results_2}
\end{figure*}

\begin{figure*}[h]
    \centering
    \includegraphics[width=\textwidth]{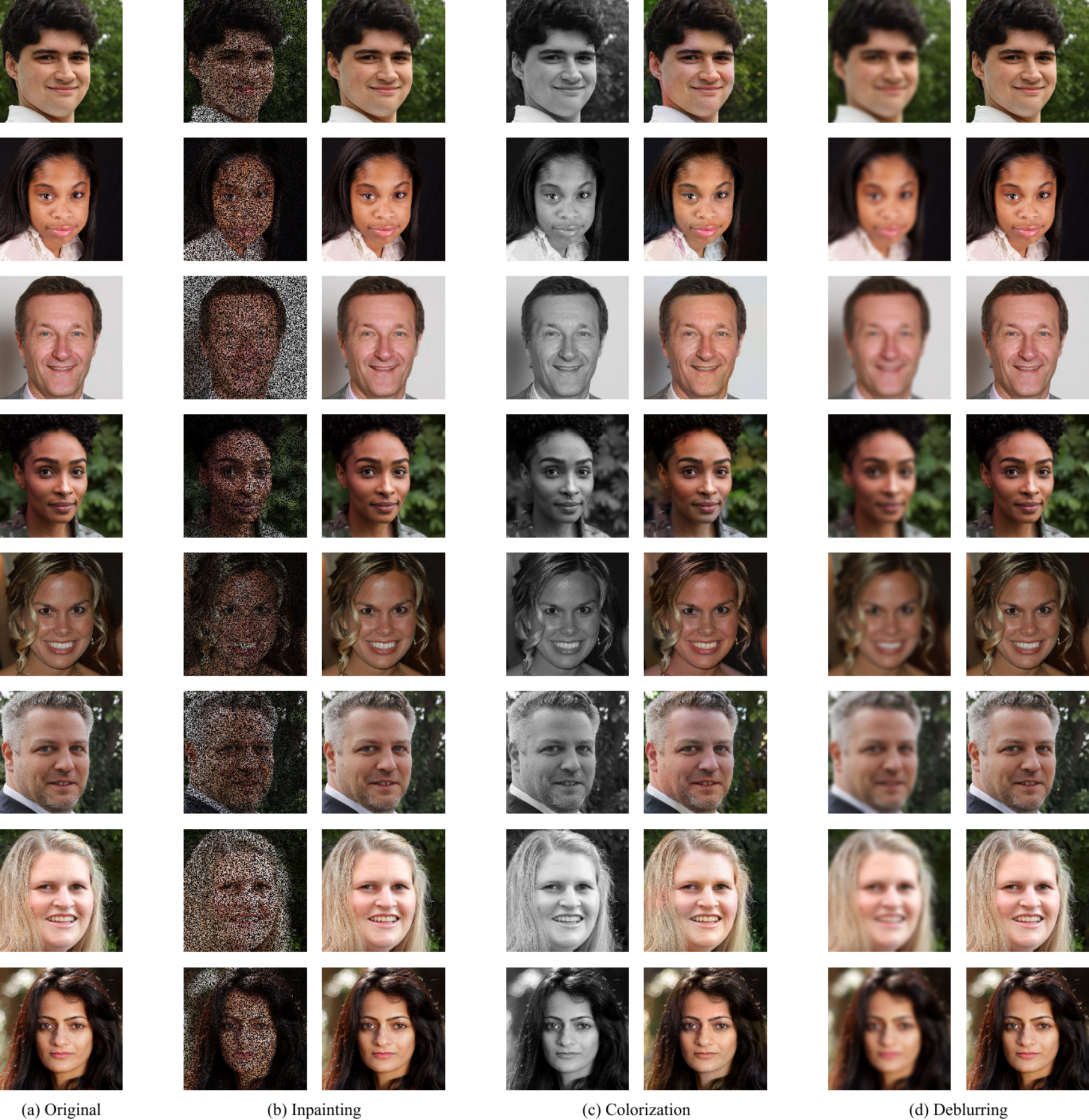}
    \caption{Image inpainting, colorization, and deblurring results of \ac{RDIM}-$10$ on images from the FFHQ dataset. For inpainting, pixels in the original images are randomly masked and set to zero with probability $p_{\text{mask}}=0.5$. For colorization, grayscale inputs are obtained by converting the original RGB images to luminance. For deblurring, synthetic blurred images are generated from ground truth images by applying a Gaussian blur with kernel size $15 \times 15$ and standard deviation $\sigma = 3.0$. \ac{RDIM} is trained with $T=50$ and $\gamma = 3.0$. Inference is conducted with $S=10$ and $\eta = 1.0$. In (b), (c) and (d), the left side represents the input image and the right side the output.}
    \label{fig:inpainting_colorization_deblurring_results}
\end{figure*}

\end{document}